\documentclass[]{interact}

\usepackage{epstopdf}                                   
\usepackage[caption=false]{subfig}                      
\usepackage[numbers,sort&compress]{natbib}              
    \bibpunct[, ]{[}{]}{,}{n}{,}{,}                     
    \makeatletter                                       
    \def\NAT@def@citea{\def\@citea{\NAT@separator}}     
    \makeatother                                        

\theoremstyle{plain}                                    

\theoremstyle{definition}

\theoremstyle{remark}

\usepackage{tabularx}                                   
\usepackage{url}                                        
\usepackage{algorithm}                                  
\usepackage{algpseudocode}                              
\usepackage{float}                                      

\newcolumntype{C}{>{\centering\arraybackslash}X}
\newcolumntype{L}{>{\raggedright\arraybackslash}X}
\newcolumntype{R}{>{\raggedleft\arraybackslash}X}

\begin{document}

\articletype{FULL PAPER}                                

\title{
    Evaluating the Impact of Adaptive Extended Reality on Human-Robot Interaction Across the Reality-Virtuality Continuum
}

\author{
    \name{
        Carl Tornberg\textsuperscript{a}\thanks{CONTACT Carl Tornberg. Email: tornbergcarl@gmail.com},
        Alicia Torck\textsuperscript{b},
        Lotfi El Hafi\textsuperscript{a}\thanks{CONTACT Lotfi El Hafi. Email: lotfi.elhafi@gmail.com},
        and Tadahiro Taniguchi\textsuperscript{a, c}
    }
    \affil{
        \textsuperscript{a}Ritsumeikan University, 1-1-1 Noji-Higashi, Kusatsu, Shiga 525-8577, Japan;\\
        \textsuperscript{b}Université catholique de Louvain (UCLouvain), 1 Place de l'Université, Louvain-la-Neuve 1348, Belgium;\\
        \textsuperscript{c}Kyoto University, Yoshida-Honmachi, Sakyo, Kyoto 606-8501, Japan
    }
}

\maketitle

\begin{abstract}
    As populations in developed countries age and labor shortages intensify, Cybernetic Avatars~(CAs) are proposed to extend human capabilities through robotic embodiments, requiring effective Human-Robot Interaction~(HRI) frameworks.
    Extended Reality~(XR), an umbrella term for Augmented Reality~(AR), Augmented Virtuality~(AV), and Virtual Reality~(VR), offers such interfaces, but prior research typically fixes the XR modality without evaluating its effect on task outcomes.
    This study examines whether the XR modality impacts HRI performance and whether an adaptive interface adjusting the level of virtuality along the Reality-Virtuality Continuum~(RVC) at runtime improves it.
    A custom XR application interfaced with a mobile manipulator supports immersive control and runtime modality switching.
    In a within-participant multi-room pick-and-place experiment comparing fixed AR, AV, and VR with dynamic RVC through task metrics, the NASA-TLX, and the System Usability Scale~(SUS), this study demonstrates that 1) the fixed reality modality affects HRI results, and 2) dynamically changing the modality along the RVC improves them.
    AR yielded significantly lower mental demand, effort, and frustration than AV and VR, while the dynamic RVC condition achieved the highest throughput and lowest workload, highlighting the value of adaptive XR interfaces for human-robot symbiosis.
    The implementation is available at \url{https://github.com/CarlTornberg/XR-HRI}.
\end{abstract}



\section{Introduction}
\label{sec:introduction}

Birth rates in developed countries have been steadily declining for decades, increasing the demand on workers in sectors such as healthcare and eldercare.
To partly combat these issues, an avatar-symbiotic society has been proposed, in which Cybernetic Avatars~(CAs) let humans act remotely through robotic embodiments, transcending limitations such as location or disability~\cite{ishiguro_realisation_2021}.
Realizing such a symbiotic future between humans, robots, and AI requires human-centered Human-Robot Interaction~(HRI) frameworks supporting intuitive control.

Traditionally, HRI has relied on 2D interfaces using a screen, mouse, and keyboard, which restrict operator mobility and lack depth cues and embodied feedback.
To address these challenges, Extended Reality~(XR) has emerged as a promising solution~\cite{suzuki_augmented_2022}, encompassing Augmented Reality~(AR), Augmented Virtuality~(AV), and Virtual Reality~(VR), which blend virtual and physical environments~\cite{milgram_augmented_1995}.
Using a head-mounted display~(HMD), XR supports stereo rendering and spatially tracked inputs~\cite{lavalle_virtual_2023}, letting operators move freely, interact naturally, and better judge distances~\cite{liu_understanding_2017}.

\begin{figure}[t]
    \centering
    \includegraphics[width=1.0\linewidth]{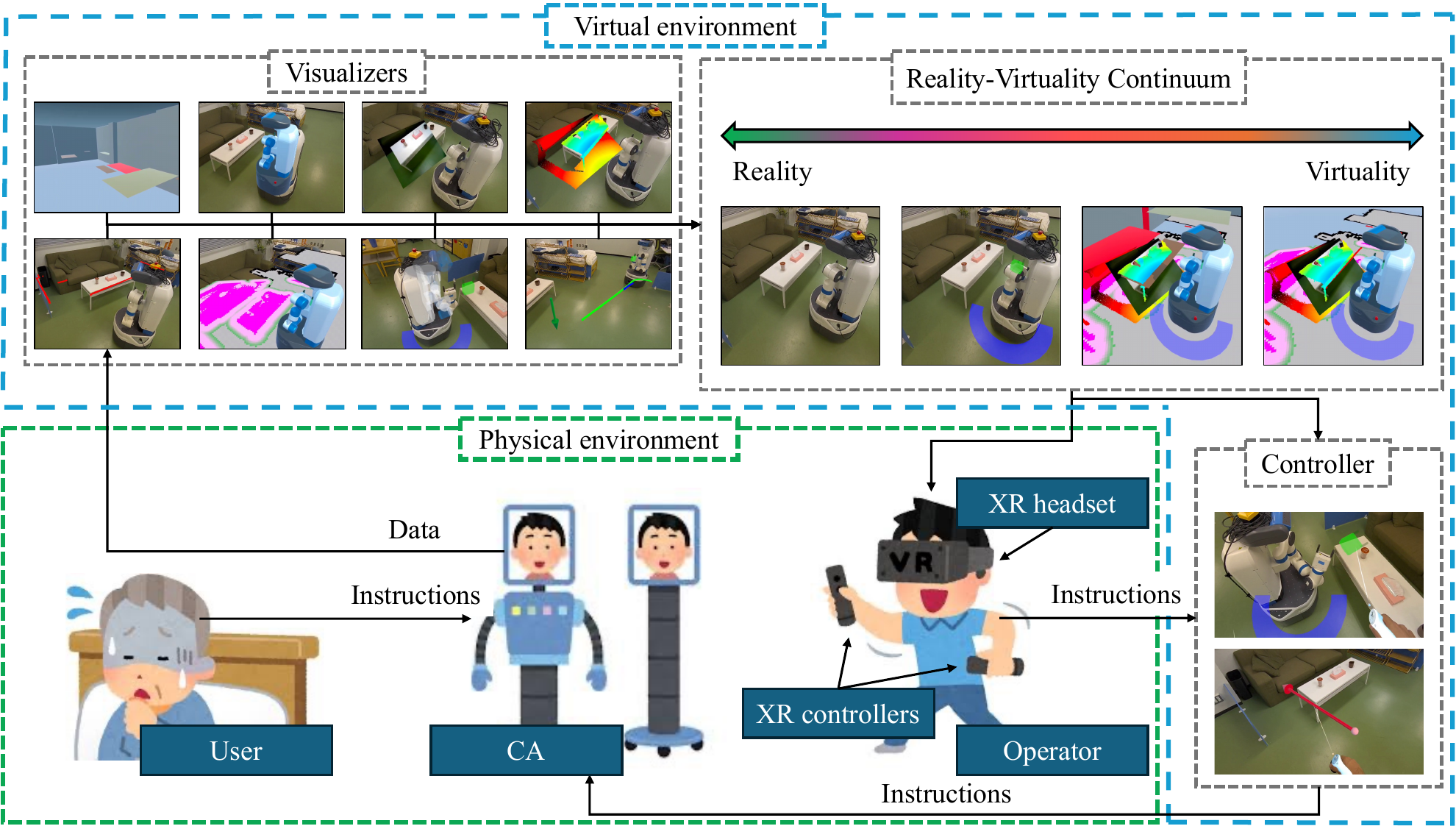}
    \caption{
        Overview of the proposed approach: an XR HMD operator's level of reality can be adjusted along the RVC to teleoperate robots to perform tasks.
    }
    \label{fig:research_overview}
\end{figure}

Modern XR HMDs, through their passthrough cameras, enable varying the degree of perceived virtual content along the Reality-Virtuality Continuum~(RVC)~\cite{milgram_augmented_1995}, illustrated in Fig.~\ref{fig:research_overview}: a linear scale from complete reality to full virtuality, with intermediate modalities AR and AV, commonly referred to as Mixed Reality~(MR), mixing real and virtual content.
Each XR modality offers different trade-offs in immersion, situational awareness, control, and task performance~\cite{suzuki_augmented_2022}: AR has been applied where real-world awareness is required, such as handovers~\cite{newbury_visualizing_2022}, collaborative assembly~\cite{maccio_mixed_2022}, and navigation in shared spaces~\cite{walker_communicating_2018}, whereas VR is frequently used for training~\cite{matsas_design_2017}, simulation~\cite{murnane_simulator_2021}, or remote operation~\cite{walker_cyber-physical_2024, szczurek_multimodal_2023}.

However, it is not clear which XR modality to use, as a task can be performed in more than one reality: teleoperation is commonly carried out in VR~\cite{walker_cyber-physical_2024, wozniak_happily_2023} yet is also feasible in AR or AV, and vice versa for AR tasks such as visualizing a robot arm's future trajectory~\cite{gruenefeld_mind_2020, maccio_mixed_2022}.
In previous research, analyzed in Tables~\ref{tab:xr_surveys_hri_summary}, \ref{tab:ar_hri_summary}, \ref{tab:mr_hri_summary}, and \ref{tab:vr_hri_summary}, the XR modality is often selected by stating the technology's capabilities, as in~\cite{tadeja_using_2024},~\cite{rauso_learning_2024}, and~\cite{wozniak_happily_2023}, or by referencing prior usage, as in~\cite{ong_augmented_2020} and~\cite{helgert_unlocking_2024}, with cross-modality comparisons insufficient or missing.

The problem is twofold.
Firstly, the selected XR modality could significantly impact productivity and user experience, as too much virtual information may overwhelm the user, while omitting real-world cues could reduce situational awareness.
Secondly, the modality is fixed at design time, preventing operators from choosing the modality that best fits their task, context, or preference at runtime.
Hence, this study hypothesizes that:
\begin{itemize}
    \item H1) The selected reality modality affects task results.
    \item H2) Enabling the user to dynamically and freely select the level of immersion improves task results.
    \item H3) Users will utilize the RVC spectrum by changing the level of reality.
\end{itemize}

To evaluate these hypotheses, an adaptive framework enabling each XR modality's visualizations and runtime adjustment of virtuality along the RVC was built with Unity and OpenXR, deployed on the Meta Quest 3 XR HMD, and interfaced with the ROS-compatible Fetch Mobile Manipulator~\cite{wise_fetch_2016, quigley_ros_2009}.
At its core, a Reality-Virtuality Manager shows or hides the robot's egocentric data, such as its camera stream, its exocentric data, such as its cost maps, and low-fidelity digital twins of real-world surfaces, according to the operator's level of virtuality.

The experiment of this study assessed fixed and dynamic reality modalities in a within-participant multi-room pick-and-place task (AR, AV, VR, and RVC conditions, 10 minutes each), combining navigation and manipulation into a 9D problem ($(x_{r}, y_{r}, \alpha_{r}) + (x_{e}, y_{e}, z_{e}, \alpha_{e}, \beta_{e}, \gamma_{e})$), where $r$ is the robot, and $e$ is the end effector.
Task performance was measured alongside subjective workload and usability through the NASA-TLX~\cite{hart_development_1988} and System Usability Scale~(SUS)~\cite{brooke_sus_1996}.
In a preliminary experiment~\cite{torck_control_2025}, the XR motion controllers scored the highest and were therefore selected as the input method.

The contributions of this study are therefore twofold:
\begin{enumerate}
    \item Demonstrating that the selected fixed reality modality (AR, AV, VR) affects the results in HRI.
    \item Demonstrating that an adaptive approach, enabling the operator to freely and dynamically change their reality modality along the RVC, improves the results in HRI compared to a fixed reality.
\end{enumerate}

The results indicated lower mental demand, effort, and frustration in AR among the fixed modalities, and with dynamic RVC changes, more objects grasped than in VR with lower mental demand, effort, and frustration than in AV and VR.
Beyond controlled experiments, the developed system underpins several academic publications and large-scale demonstrations~\cite{el_hafi_public_2025, nikkan_kogyo_shimbun_ritsumeikan_2024, tornberg_mixed_2024, garcia_ricardez_toward_2023, nikkan_kogyo_shimbun_ritsumeikan_2023}, and JST Moonshot R\&D Program Goal 1~\cite{ishiguro_realisation_2021}.

The structure of this paper is as follows:
\begin{itemize}
    \item Section~\ref{sec:previous_work} reviews XR in HRI and positions this study.
    \item Section~\ref{sec:proposed_approach} details the proposed approach and methodology.
    \item Section~\ref{sec:implementation} describes the system implementation.
    \item Section~\ref{sec:experiments} describes the experiment.
    \item Section~\ref{sec:results} presents the results.
    \item Section~\ref{sec:discussion} discusses the findings and hypotheses.
    \item Section~\ref{sec:limitations_future_work} outlines the limitations and future work.
    \item Section~\ref{sec:conclusion} concludes this study.
\end{itemize}


\section{Previous Work}
\label{sec:previous_work}

Research at the intersection of XR and HRI has grown over the past decade to improve teleoperation, task performance, safety, and trust.
However, most studies focus on fixed XR modalities and lack direct comparisons along the RVC.


\subsection{Extended Reality in Human-Robot Interaction}

Handheld AR devices occupy the user's hands and are unable to fully exclude real-world content, and projections are limited to 2D surfaces, whereas an HMD enables hands-free, stereoscopic rendering, leveraging depth perception which is critical in HRI~\cite{liu_understanding_2017, suzuki_augmented_2022}.
Only recently have XR HMDs reached consumer-grade prices, increasing XR research within HRI.
Crucially, modern XR HMDs also feature passthrough cameras displaying the physical surroundings, which earlier VR headsets lacked.
Trust is a key element in Human-Robot Collaboration~(HRC)~\cite{freedy_measurement_2007, salem_would_2015}, and XR is well established to improve it by, for instance, visualizing a robot's future intentions to provide context awareness~\cite{suzuki_augmented_2022, newbury_visualizing_2022, palmarini_designing_2018, tsamis_intuitive_2021}.

In this study, we therefore adopt a consumer-grade XR HMD with passthrough cameras to leverage hands-free, stereoscopic visualization across the entire RVC.


\subsubsection{Augmented Reality in Human-Robot Interaction}

AR is commonly used where the user is co-located with an in-sight robot, collaborating to perform tasks, plan, visualize intent, and enhance safety awareness.
For example,~\cite{gruenefeld_mind_2020} and~\cite{walker_communicating_2018} visualized a robot's future trajectory or movement in AR, allowing users to plan and avoid collisions,~\cite{lunding_ar-supported_2023} facilitated workspace awareness in collaborative assembly,~\cite{lee_interactive_2024} enabled non-experts to select automatically generated collision-free pathways using gestures, and~\cite{ong_augmented_2020} presented an AR system for intuitive robot task programming.

In this study, we include AR as a fixed condition and extend its intent visualizations, such as the future pose of the end effector, to all modalities along the RVC.


\subsubsection{Virtual Reality in Human-Robot Interaction}

In contrast, VR is commonly used when the robot is out of sight or absent, for teleoperation, programming, or simulation, and, being fully immersive, allows safe HRC without risk of injury if the task fails.
For example, in~\cite{wozniak_happily_2023}, the user teleoperates a robot by adjusting incorrectly perceived or missing objects in its exocentric perception using VR, found more useful, easier to use, and preferred over a traditional 2D screen, while~\cite{ortenzi_robot_2022} showed in virtual handovers that the more occluded the object, the longer it took to grasp.

In this study, we include VR as a fixed condition in which the robot's egocentric and exocentric data entirely replace the user's physical surroundings.


\subsubsection{Augmented Virtuality and Mixed Reality in Human-Robot Interaction}

Between complete reality and full virtuality lies MR, an umbrella term encompassing both AR and AV~\cite{milgram_augmented_1995}, where AV is the opposite of AR, including only a small amount of reality.
MR is commonly used in settings similar to AR: robot egocentric and exocentric data was visualized in MR for navigation in~\cite{ostanin_multi_2021}, and~\cite{maccio_mixed_2022} visualized the future end position of a robot's arms during collaborative assembly, suggesting that MR improves task performance and collaboration.
MR interfaces have also enabled non-expert users to visualize and correct a service robot's perception during object categorization, reducing the user's workload compared to speech-only feedback~\cite{el_hafi_system_2020, el_hafi_teaching_2021, nakamura_multimodal_2022}.
MR has also enabled on-site 6D-pose annotation, increasing accuracy compared to a PC-based system~\cite{tornberg_mixed_2024}, and~\cite{garcia_ricardez_toward_2023} proposed utilizing XR and AI to visualize robot motion and safety data in real-time.
Both served as early prototypes of the approach proposed here.
AV, to our knowledge, has not been explicitly explored: although~\cite{milgram_augmented_1995} defines AV, work such as~\cite{walker_cyber-physical_2024} and~\cite{szczurek_multimodal_2023} matches that definition yet is labeled as MR by the authors.

In this study, we explicitly include AV as a fixed condition, positioned between AR and VR along the RVC, visualizing low-fidelity representations of real-world surfaces together with the robot's egocentric and exocentric data, while excluding high-fidelity passthrough.


\subsection{Modality Selection Reasoning}

Tasks are commonly deconstructed into subtasks for evaluation, such as navigation~\cite{walker_cyber-physical_2024}, object perception~\cite{wozniak_happily_2023}, and handover~\cite{newbury_visualizing_2022,ortenzi_robot_2022}, and a given subtask has been performed in different realities, for example, navigation in both MR~\cite{walker_cyber-physical_2024,kennel-maushart_interacting_2023} and VR~\cite{huang_evaluation_2024}.
The modality is commonly chosen from the drawbacks and benefits of each reality, as in~\cite{pascher_how_2023},~\cite{maccio_mixed_2022}, and~\cite{huang_evaluation_2024}, or from prior usage and technology acceptance, as in~\cite{ong_augmented_2020},~\cite{shaaban_investigating_2024}, and~\cite{helgert_unlocking_2024}.
Tables~\ref{tab:xr_surveys_hri_summary}, \ref{tab:ar_hri_summary}, \ref{tab:mr_hri_summary}, and \ref{tab:vr_hri_summary} summarize this reasoning for surveys and for AR, MR, and VR HRI papers, where the modality is always fixed without a cross-modality comparison.

In this study, we instead compare the modalities directly within a single task and additionally enable the operator to freely select their modality along the RVC.

\begin{table}[H]
    \begin{center}
        \footnotesize
        \caption{
            Summary of XR modalities reasoning by surveys in HRI.
        }
        \label{tab:xr_surveys_hri_summary}
        \begin{tabularx}{1.0\linewidth}{lXc}
            \hline
            Title & Description & XR Modality \\
            \hline
            ~\cite{suzuki_augmented_2022} & ``Augmented reality~(AR) interfaces promise to address these challenges, as AR enables us to design expressive visual feedback without many of the constraints of physical reality.'' & AR \\
            ~\cite{makhataeva_augmented_2020} & ``Augmented Reality~(AR) has become a popular multidisciplinary research field over the last decades. [...] In robotics, AR acts as a new medium for interaction and information exchange with autonomous systems increasing the efficiency of the Human-Robot Interaction~(HRI). The most common definition of AR by Azuma~\cite{azuma_survey_1997} states that in AR ``3D virtual objects are integrated into a 3D real environment in real-time'' ''& AR/XR \\
            ~\cite{yu_mr_2022} & ``AR creates a perceptual layer to achieve the complementarity of reality and the environment. Under the premise of AR technology, Mixed Reality~(MR) is a further development, using advanced computer technology, image processing technology, and human-computer interaction technology to generate a visual environment with the characteristics of virtual and real integration, in which virtual and real objects coexist and user can interact with them in realtime~\cite{selonen_mixed_2012}'' & MR \\
            ~\cite{walker_virtual_2023} & ``VAM interfaces allow users to see 3D virtual imagery in a virtual space or contextually embedded within their environment. [...] can also be used hands-free when in the form of a head-mounted display~(HMD) that allows for more fluid and natural interactions''
            ``AR visualization techniques offer significant promise in HRI due to their capability to effortlessly communicate robot intent and ''
            ``VR techniques, however, provide unique opportunities for safe, flexible, and novel environments''
            & VAM/XR \\
            ~\cite{badia_virtual_2022} & ``Although the use of AR allows for enhancing HRC through the addition of virtual interfaces, cues, etc., its efficacy first needs to be tested. Such testing of AR interface can be easier to do in VR, where the virtual interfaces can be emulated in a range of scenarios as described above, and thereby would be tried against several options. This, in turn, results in an agile development of solutions that are not limited to a specific laboratory/experimental context.'' & VR \\
            \hline
        \end{tabularx}
    \end{center}
\end{table}

\begin{table}[H]
    \begin{center}
        \footnotesize
        \caption{
            Summary of AR reasoning in HRI papers.
        }
        \label{tab:ar_hri_summary}
        \begin{tabularx}{1.0\linewidth}{lXc}
            \hline
            Title & Description & XR Modality \\
            \hline
            ~\cite{fang_novel_2014} & ``In VR-based HRI, a virtual environment~(VE) aided by the necessary sensors provides the operator with an immersive sense of his/her presence at the real location undertaking the tasks'' ``AR can assist the users to interact intuitively with the virtual objects and spatial information for task planning within the actual working environment'' & AR \\
            ~\cite{azuma_survey_1997} & ``Recently, Virtual Reality~(VR) has been explored as a potential method to simplify demonstration collection process~\cite{george_openvr_2025},~\cite{zhang_deep_2018}, but such approaches involve additional efforts such as creating realistic VR environments, leading to further complications. We are inspired by an alternative approach, suggested by ~\cite{duan_ar2-d2_2023}, to use Augmented Reality~(AR) to enable more natural collection of demonstrations.'' & AR \\
            ~\cite{lunding_ar-supported_2023} & ``enabling the operator to reschedule tasks on the fly, and how users prefer to coordinate and collaborate with robots. To address these, we propose an Augmented Reality interface'' & AR \\
            ~\cite{ong_augmented_2020} & ``this paper presents an Augmented Reality-assisted robot programming system [...] Augmented Reality~(AR) has been widely applied to develop applications to provide guidance and/or electronic user manuals to assist users in manufacturing processes'' & AR \\
            ~\cite{walker_communicating_2018} & ``explore an alternative design space: using Augmented Reality [...] is inspired by prior research envisioning that Augmented Reality head-mounted displays (ARHMDs) might one-day support intuitive human-robot communication for collocated users'' & AR \\
            ~\cite{palmarini_designing_2018} & ``provide digital information for increased situational awareness, [...] Using AR to display information, such as robot state, progress and even intent, will enhance understanding, grounding, and thus collaboration.'' & AR \\
            ~\cite{pascher_how_2023} & ``Augmented Reality~(AR) technology to enable intuitive teleoperation'' & AR \\
            ~\cite{lee_interactive_2024} & ``AR has an advantage in human-robot cooperative motion as it has the ability to recognize human gaze and hand motions without an external device. Additionally, numerous studies have explored the utilization of Augmented Reality~(AR)'' & AR \\
            ~\cite{tsamis_intuitive_2021} & ``Through an AR display, a service robot can provide indications on both its current state, action and its safety implications, as well as on its next action and plan, aspects that can be particularly precious both in real human-robot collaboration spaces, as well as in initial, training sessions of human workers that will then collaborate with robots on a daily basis.'' & AR \\
            ~\cite{gruenefeld_mind_2020} & ``Explaining decisions and future actions by robotic AI systems has a unique geometric aspect, which can be exploited with 3D visualizations in Augmented Reality~(AR)'' & AR \\
            ~\cite{tadeja_using_2024} & ``it has been proposed to use Augmented Reality~(AR) as a promising approach to facilitating more intuitive HRC .'' & AR \\
            ~\cite{hoang_virtual_2022} & ``Recently, Augmented Reality~(AR) technology has been explored as a promising alternative to traditional programming interfaces with great potential for enabling fluent human-robot interaction'' & AR \\
            ~\cite{newbury_visualizing_2022} & ``Recent advancements in graphics and hardware technology have led to increased popularity of Augmented Reality~(AR). AR presents new opportunities for robotics applications, and it is especially promising for Human-Robot Interaction~\cite{makhataeva_augmented_2020}. Using AR, robots can communicate their intent to users through visual displays, allowing users to consider the robot's plans and act accordingly.'' & AR \\
            \hline
        \end{tabularx}
    \end{center}
\end{table}

\begin{table}[H]
    \begin{center}
        \footnotesize
        \caption{
            Summary of MR reasoning in HRI papers.
        }
        \label{tab:mr_hri_summary}
        \begin{tabularx}{1.0\linewidth}{lXc}
            \hline
            Title & Description & XR Modality \\
            \hline
            ~\cite{schmidt_augmented_2022} & ``there are several examples of usage of Augmented Reality~(AR) and Mixed Reality~(MR) in industry'' & AR/MR \\
            ~\cite{ostanin_multi_2021} & ``human-robot interaction~(HRI) requires intuitive and efficient interfaces. One of the best ways to implement this in HRI is augmented and Mixed Reality, according to \cite{goodrich_human-robot_2008}.'' & AR/MR \\
            ~\cite{kennel-maushart_interacting_2023} & ``MR-headsets [...] a more recently developed type of interface, which allows for a bigger variety of interactions.'' & MR \\
            ~\cite{shaaban_investigating_2024} & ``MR has been explored as a tool for effective communication between humans and robots [...] hold the potential to make the robot's internal state intuitively understandable to human collaborators by projecting its planned actions as holograms.'' & MR \\
            ~\cite{maccio_mixed_2022} & ``Unlike Augmented Reality~(AR), which is limited to the overlay of digital visual content onto the observed scene, MR creates immersive experiences by bringing interactive and spatially contextualized holograms into the real world.'' & MR \\
            ~\cite{tornberg_mixed_2024} & ``the key benefits of using MR is the freedom of moving between several points of view to resolve visual ambiguities from 3D to 2D projection'' & MR \\
            ~\cite{pant_mixed-sense_2024} & ``Mixed Reality~(MR) sensor emulation framework'' ``It represents the spectrum encompassing both Augmented Reality~(AR) and Virtual Reality~(VR), synergistically combining elements of the physical and digital world.'' & MR \\
            ~\cite{szczurek_multimodal_2023} & ``There have been attempts to use VR HMDs to present the information in robotic intervention scenarios without success due to the complexity and intrusiveness of such headsets and used VR controllers, which often caused motion sickness and were difficult to deploy in the field.'' & MR \\
            ~\cite{walker_cyber-physical_2024} & ``design and evaluation of an immersive Mixed Reality interface that is developed for mobile robots that are part of a larger human-robot field team'' & MR \\
            \hline
        \end{tabularx}
    \end{center}
\end{table}

\begin{table}[H]
    \begin{center}
        \footnotesize
        \caption{
            Summary of VR reasoning in HRI papers.
        }
        \label{tab:vr_hri_summary}
        \begin{tabularx}{1.0\linewidth}{lXc}
            \hline
            Title & Description & XR Modality \\
            \hline
            ~\cite{liu_understanding_2017} & ``Although the use of AR allows for enhancing HRC through the addition of virtual interfaces, cues, etc., its efficacy first needs to be tested. Such testing of AR interface can be easier to do in VR, where the virtual interfaces can be emulated in a range of scenarios as described above, and thereby would be tried against several options. This, in turn, results in an agile development of solutions that are not limited to a specific laboratory/experimental context.'' & VR \\
            ~\cite{huang_evaluation_2024} & ``[...] leveraging Virtual Reality~(VR) technology, which can deliver immersive user experiences with intuitive controls'' & VR \\
            ~\cite{wozniak_happily_2023} & ``It is not always trivial to decide which technology is better when it comes to HRI applications. Despite this, there is a lack of studies that compare the effectiveness of VAM and screen-based interfaces for HRI.'' ``We chose VR because it provides an immersive remote interaction with the robot.'' & VR \\
            ~\cite{rauso_learning_2024} & ``Virtual Reality to provide demonstrations to the robotic system in a safe and simplified manner'' & VR \\
            ~\cite{ortenzi_robot_2022} & Missing explanation. ``To this end, we performed an experiment in Virtual Reality~(VR) where a simulated robot offers tools with different degrees of visual occlusion [of the object handed over].'' & VR \\
            ~\cite{helgert_unlocking_2024} & ``Virtual reality has become an increasingly accepted solution in recent years to deal with these limitations and problems.'' & VR \\
            ~\cite{luo_user-customizable_2024} & ``user-customizable shared control by directly communicating the internal arbitration parameters to an operator for refinement [...] in Virtual Reality~(VR)'' & VR \\
            ~\cite{hein_virtual_2024} & ``VR simulators utilize high-quality 3D models of anatomical structures in fully synthetic environments'' & VR \\
            \hline
        \end{tabularx}
    \end{center}
\end{table}


\section{Proposed Approach}
\label{sec:proposed_approach}

This study proposes an adaptive framework enabling both fixed and dynamic use of XR interfaces along the RVC, guided by Hypotheses H1 to H3 (Section~\ref{sec:introduction}).


\subsection{Reality-Virtuality Manager}
\label{sec:reality_virtuality_manager}

\begin{figure}[t]
    \centering
    \includegraphics[width=1.0\linewidth]{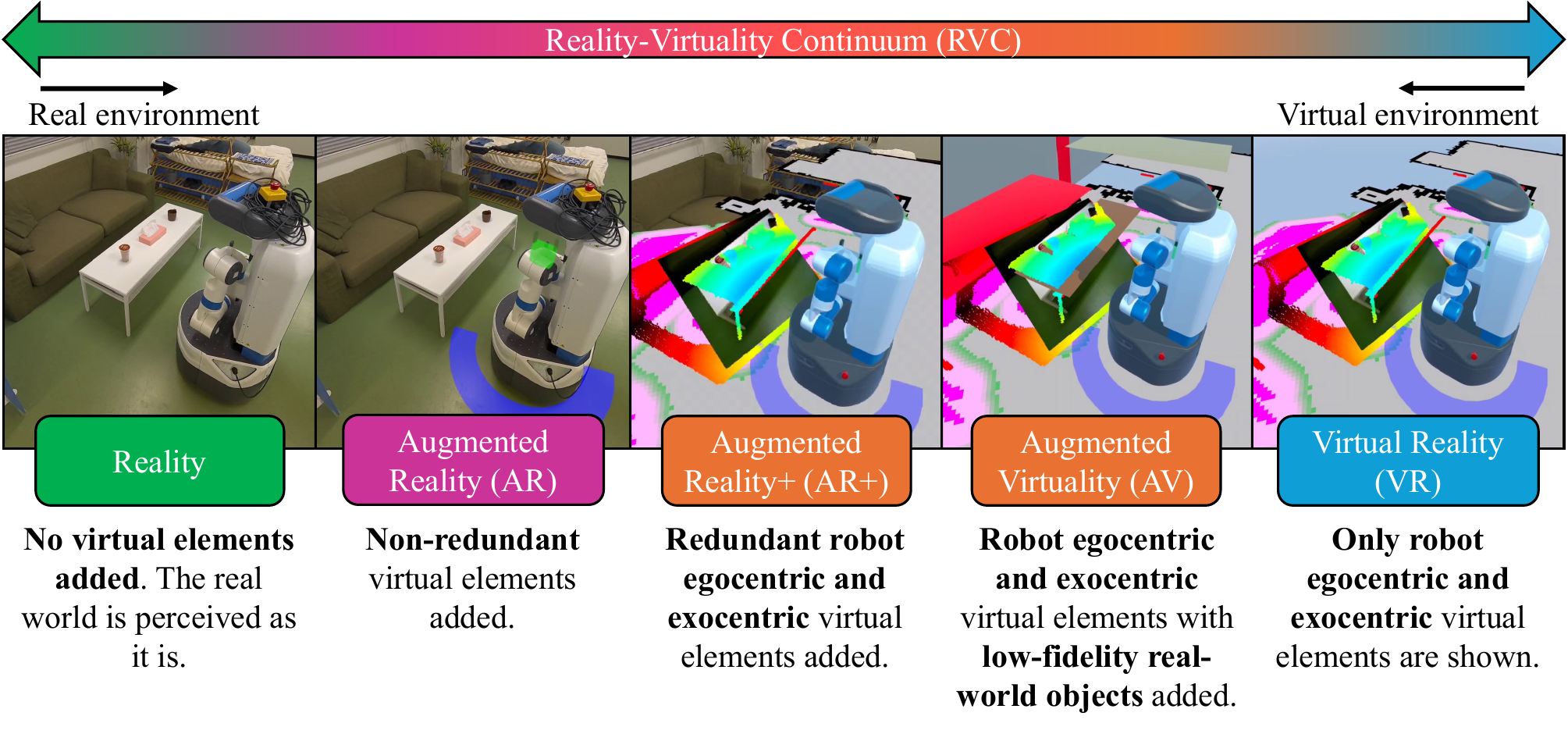}
    \caption{
        Milgram's Reality-Virtuality Continuum, a linear spectrum spanning from a fully real environment to a fully immersive virtual environment.
    }
    \label{fig:approach_rvc}
\end{figure}

To enable an operator to change the XR modality, a Reality-Virtuality Manager is proposed, which sets the level of reality either as fixed or dynamically changed at runtime through controller-based inputs.
An XR HMD interface lets the user move freely, leverage depth perception, and instruct the robot to navigate or manipulate, visualizing the robot's egocentric and exocentric data through five preset realities:

\begin{itemize}
    \item Reality:
          No visualizations are shown, and the user only perceives the real world through the passthrough cameras.
    \item Augmented Reality~(AR):
          No redundant information is shown, ensuring that the high-fidelity passthrough video is not occluded.
    \item Augmented Reality+~(AR+):
          Egocentric and exocentric data is visualized, providing a redundant and detailed view of the robot's status, while maintaining passthrough video.
    \item Augmented Virtuality~(AV):
          All egocentric and exocentric data is shown, with virtual objects limited to low-fidelity ones, primarily the AR planes.
    \item Virtual Reality~(VR):
          Only egocentric and exocentric data is shown, and the real world is entirely occluded and replaced with a blue skybox.
\end{itemize}

As evaluations of XR-based HRI commonly target a single subtask, such as navigation~\cite{walker_cyber-physical_2024, huang_evaluation_2024}, perception~\cite{wozniak_happily_2023}, or handovers~\cite{ortenzi_robot_2022, newbury_visualizing_2022}, often on independent tasks~\cite{walker_communicating_2018,gruenefeld_mind_2020}, this study promotes a multi-subtask pick-and-place experiment in a shared, multi-room simulated service environment.


\subsection{Proposed Methodology for Assessing HRI in RVC}

To systematically explore HRI within the RVC, a within-participant design was employed to reduce inter-individual variability, with two interaction paradigms: fixed modality, where the user operates exclusively in one of three predefined XR conditions (AR, AV, VR), mirroring existing XR literature, and dynamic modality (RVC), where the user transitions freely across modalities in real time.
This distinction evaluates whether fixed constraints hinder adaptability and whether user-directed transitions enhance performance and experience.

Visual feedback was categorized into two types of robot perception embedded within the XR HMD view: egocentric, the robot's internal perception, including its joint states, localization, and Point of View~(PoV), and exocentric, its perception of the surrounding world, such as cost maps and depth information.
The study combines two fundamental HRI subtasks in a single pick-and-place service scenario: navigation, a low-dimensional (3 Degrees of Freedom~(DOF)) subtask moving the robot to a designated location, and manipulation, a high-dimensional (6 DOF) subtask requiring precise control of the end effector's pose.

A mixed-methods approach quantified the influence of both configurations: the quantitative indicators were the number of objects grasped/placed (task throughput), subtask time (temporal efficiency), and time spent in each modality during the dynamic RVC condition (modality preference), while subjective experiences were assessed using the NASA-TLX~\cite{hart_development_1988} workload index and the SUS~\cite{brooke_sus_1996}.


\section{Implementation}
\label{sec:implementation}


\begin{figure}[t]
    \centering
    \includegraphics[width=1.0\linewidth]{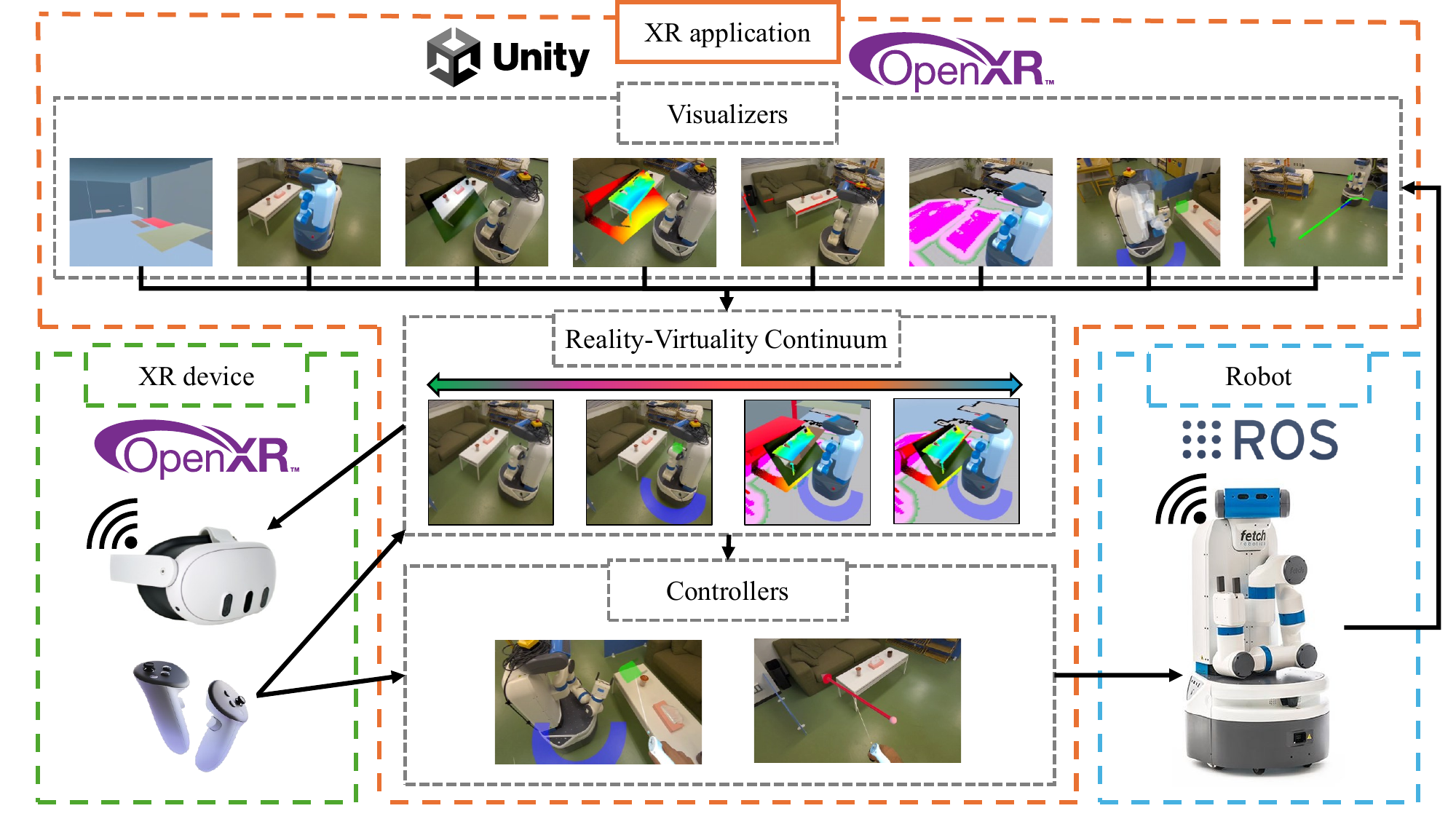}
    \caption{
        Implementation diagram.
        The XR device and robot communicate wirelessly via TCP, and the robot's egocentric and exocentric data is visualized along the RVC.
        The user can control the robot's navigation and manipulation, as well as the level of virtual content along the RVC using the XR motion controllers.
    }
    \label{fig:implementation_diagram}
\end{figure}

The implementation integrates a robotic platform with an immersive XR interface, visualized in Fig.~\ref{fig:implementation_diagram}.
The robot selected was the Fetch Mobile Manipulator~\cite{wise_fetch_2016}, a single-arm mobile robot capable of both indoor navigation and object manipulation, operating on ROS~\cite{quigley_ros_2009}.
For the XR component, the Meta Quest 3\footnote{Meta Quest 3 (\url{https://www.meta.com/jp/en/quest/quest-3/})} was selected, an XR-capable HMD supporting controller-based interaction (hereon XR motion controllers), hands-free hand tracking, and OpenXR\footnote{Khronos OpenXR (\url{https://www.khronos.org/openxr/})}, allowing the software to operate on a broad range of compliant headsets.
Unity 6\footnote{Unity 6 (\url{https://unity.com/releases/unity-6})}, which supports OpenXR, was used to develop the XR interface from Unity's Mixed Reality template project\footnote{Unity OpenXR: Meta (\url{https://docs.unity3d.com/Packages/com.unity.xr.meta-openxr@2.1/manual/index.html})}.
The open-source ROS TCP Endpoint\footnote{ROS TCP Endpoint (\url{https://github.com/Unity-Technologies/ROS-TCP-Endpoint})}, launched as a ROS node, paired with the ROS TCP Connector\footnote{ROS TCP Connector (\url{https://github.com/Unity-Technologies/ROS-TCP-Connector})} in Unity, handles the serialization and deserialization of ROS-native messages in both directions, synchronizing sensor data, robot state, and user input in real time.
The robot-side ROS environment was deployed using a containerized software development environment for robotics~\cite{el_hafi_abstraction-rich_2018, el_hafi_software_2022}, whereas the XR application itself relies only on generic ROS and OpenXR interfaces, keeping it robot- and headset-agnostic.


\subsection{AR Plane Integration}
\label{sec:ar_plane_integration}

\begin{figure}[t]
    \centering
    \includegraphics[width=1.0\linewidth]{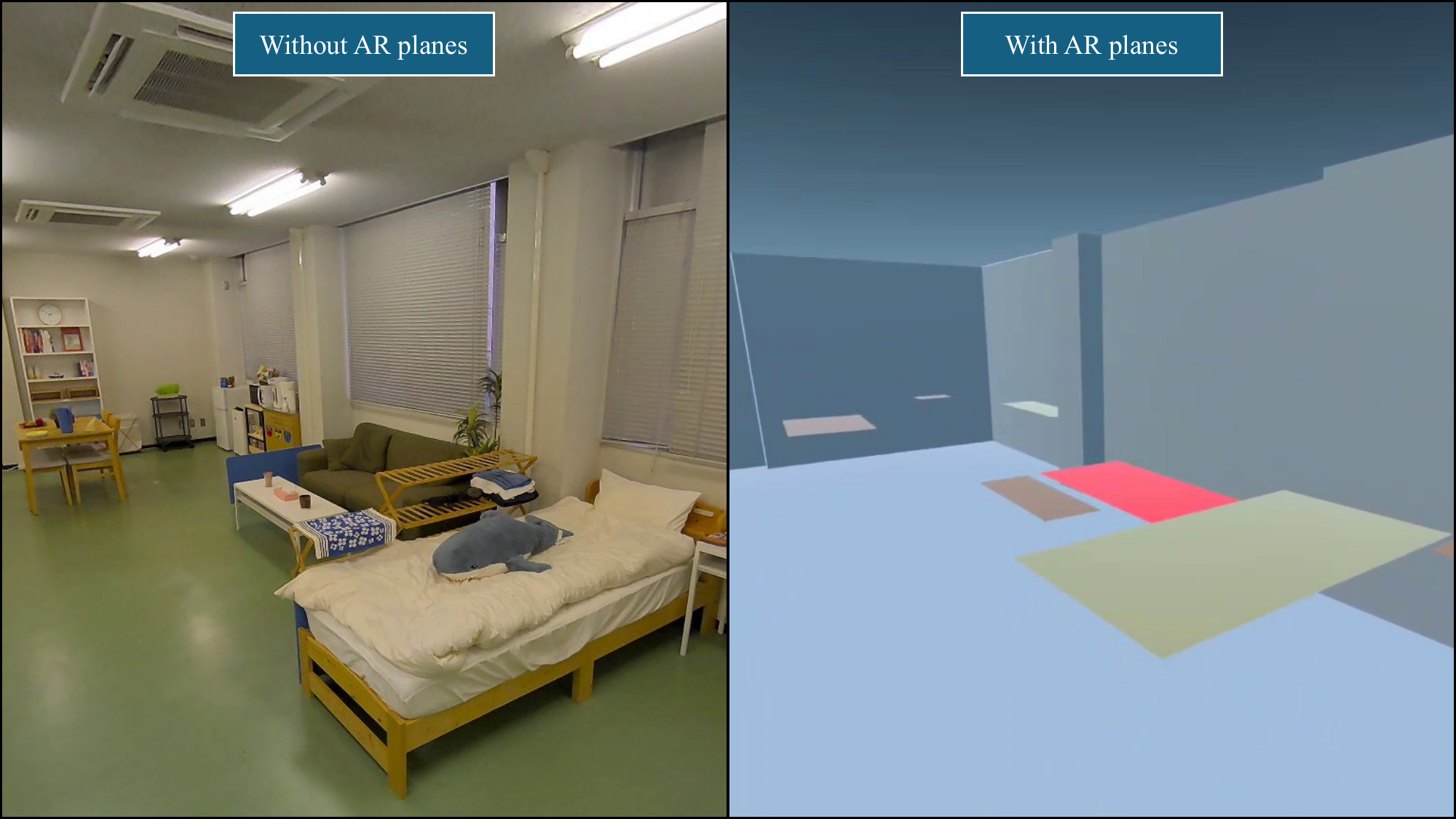}
    \caption{
        XR HMD's PoV of AR planes.
    }
    \label{fig:implementation_view_ar_planes}
\end{figure}

The Meta Quest 3 features built-in environment segmentation: the user scans the physical space, and the headset recognizes walls, floors, ceilings, windows, and objects such as tables, which are stored persistently across sessions and rooms.
Through Unity's OpenXR support, each object is accessible through its top surface as an AR plane managed by the ARPlaneManager, and an ARPlanePainter creates a color-coded digital twin of each plane (Fig.~\ref{fig:implementation_view_ar_planes}).
Furthermore, each plane's 2D dimensions and Z-position create a padded 3D box passed to the ROS planner relative to a parent Transformation~(TF), commonly the \texttt{baselink}, to enhance collision avoidance.
As these planes bring real-world structure into the virtual environment, they are shown in AV but hidden in VR (Section~\ref{sec:reality_virtuality_manager}).


\subsection{Digital Twin}
\label{sec:digital_twin}

\begin{figure}[t]
    \centering
    \begin{minipage}[t]{0.49\linewidth}
        \centering
        \frame{\includegraphics[width=\linewidth]{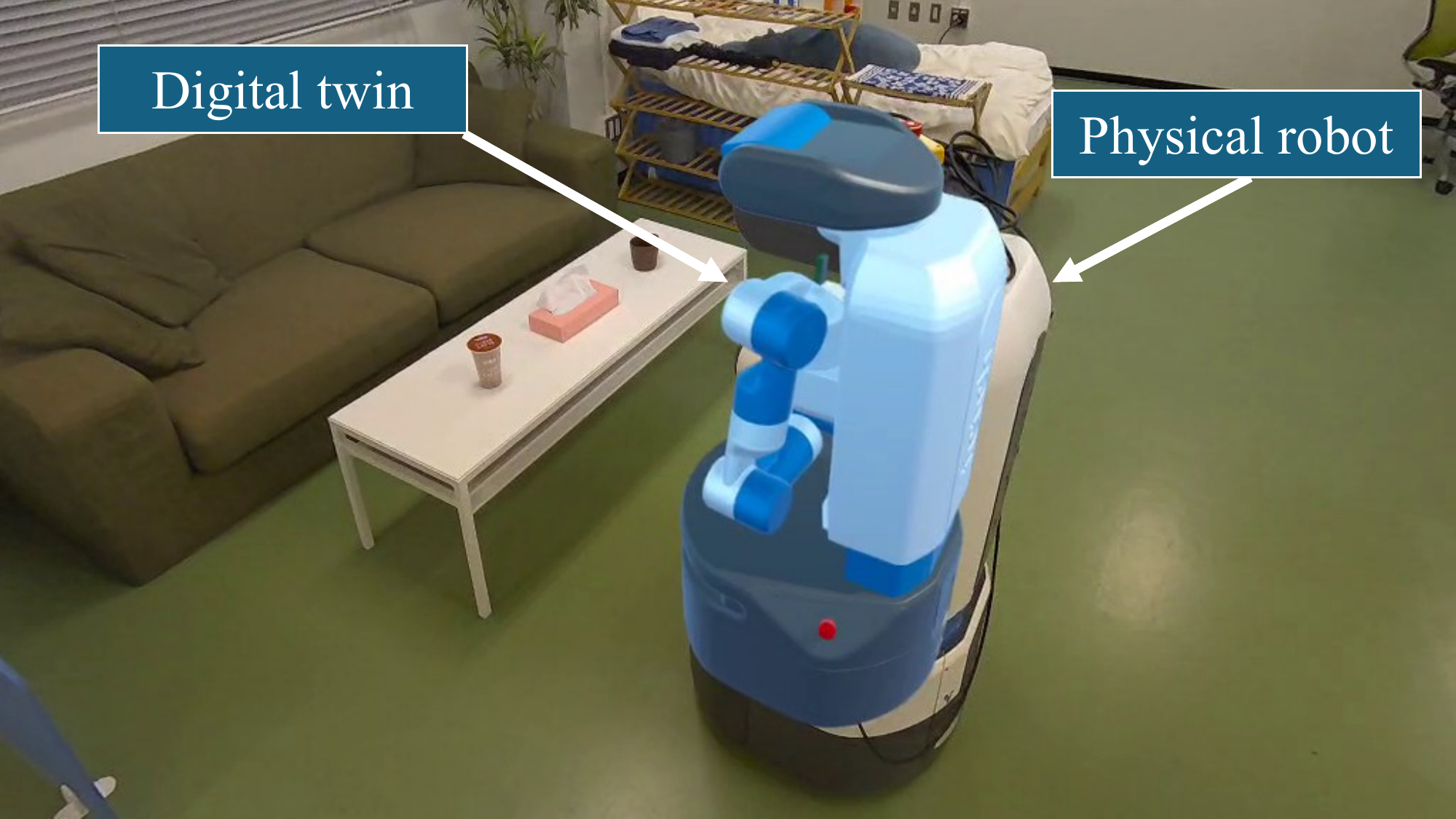}}
        \caption{
            XR HMD's PoV of the Fetch robot's digital twin.
        }
        \label{fig:implementation_view_digital_twin}
    \end{minipage}
    \hfill
    \begin{minipage}[t]{0.49\linewidth}
        \centering
        \frame{\includegraphics[width=\linewidth]{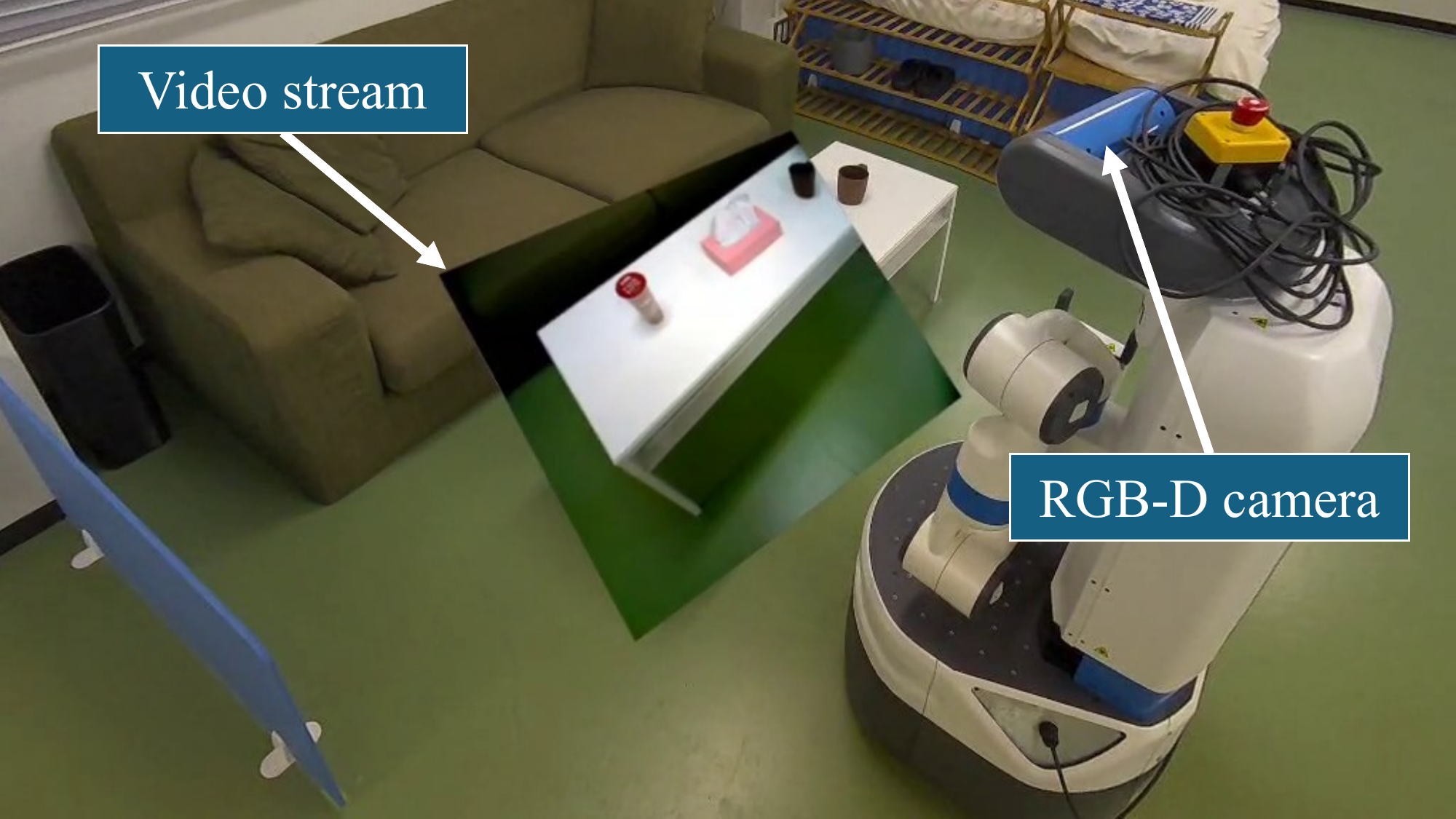}}
        \caption{
            XR HMD's PoV of the robot egocentric video stream.
        }
        \label{fig:implementation_view_image_viewer}
    \end{minipage}

    \vspace{1em}

    \begin{minipage}[t]{0.49\linewidth}
        \centering
        \frame{\includegraphics[width=\linewidth]{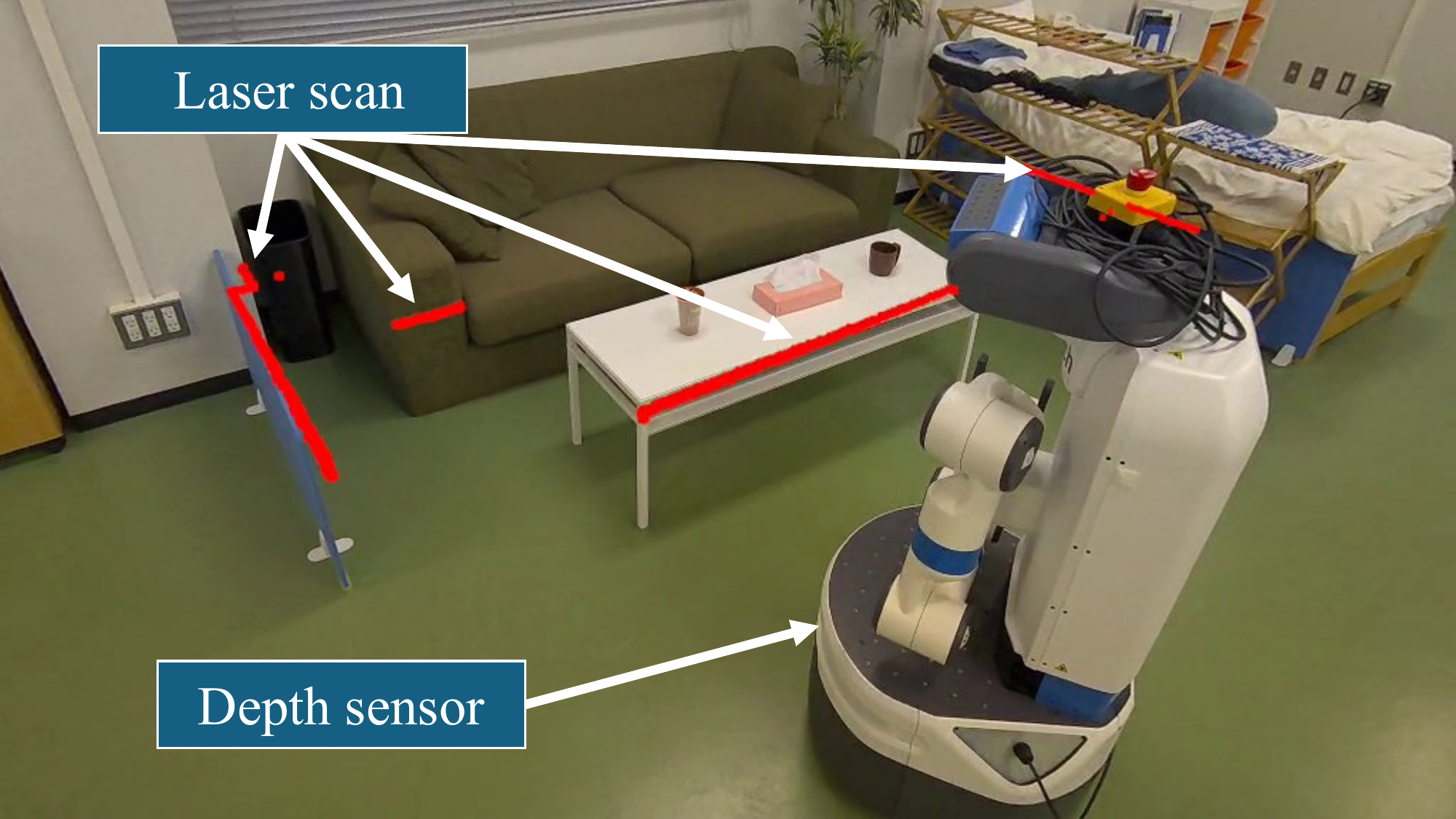}}
        \caption{
            XR HMD's PoV of the robot exocentric laser scan.
        }
        \label{fig:implementation_view_laser_scan}
    \end{minipage}
    \hfill
    \begin{minipage}[t]{0.49\linewidth}
        \centering
        \frame{\includegraphics[width=\linewidth]{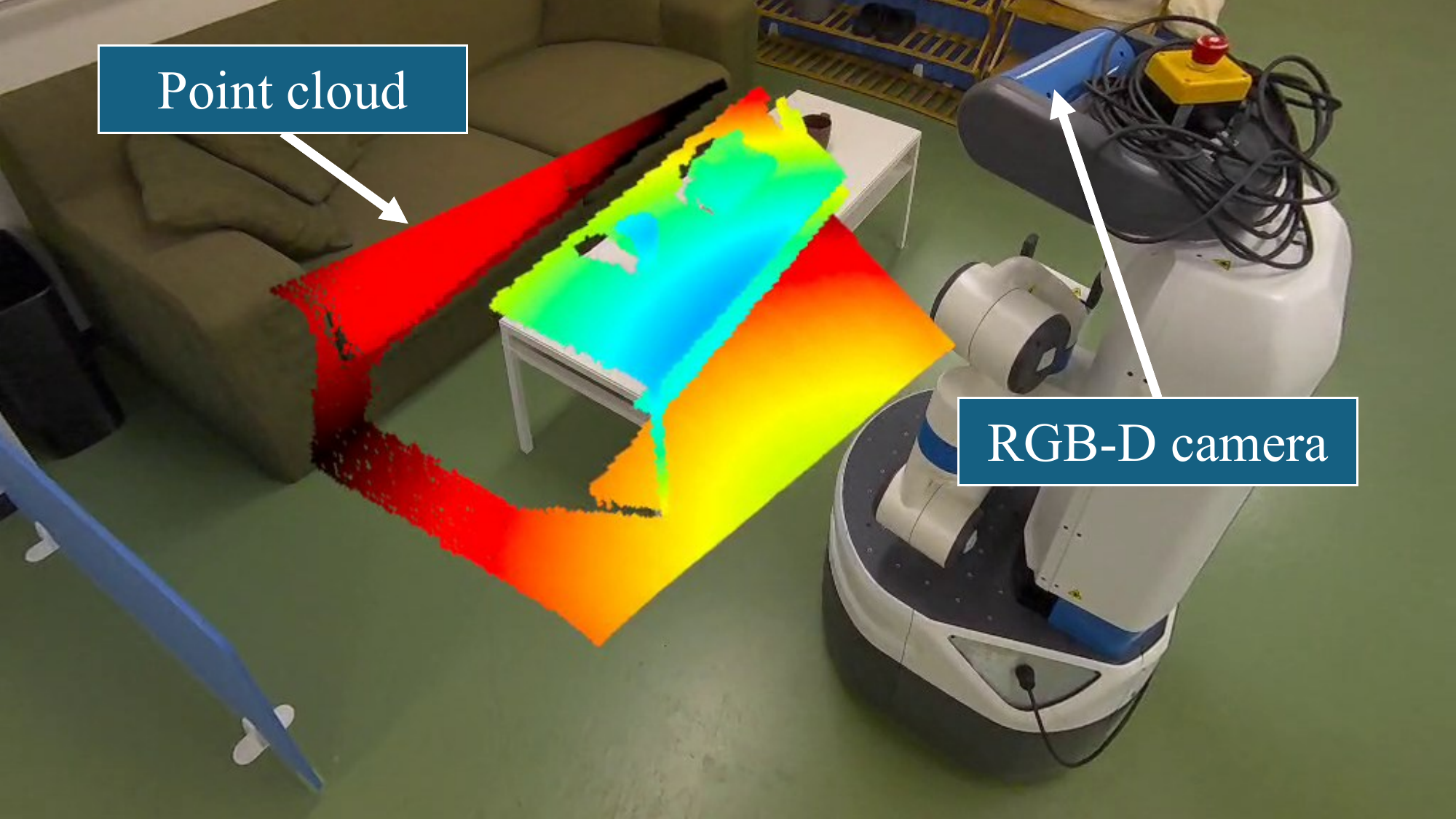}}
        \caption{
            XR HMD's PoV of the robot exocentric point cloud.
        }
        \label{fig:implementation_view_point_cloud}
    \end{minipage}
\end{figure}

To visualize the robot's current state, a digital twin is rendered as an overlay on the physical world, enabling users to intuitively perceive the robot's posture and ongoing actions (Fig.~\ref{fig:implementation_view_digital_twin}).
The Unity URDF Importer\footnote{URDF Importer (\url{https://github.com/Unity-Technologies/URDF-Importer})} converts the robot's Unified Robot Description Format~(URDF) file into a Unity-compatible GameObject, initialized by a generic wrapper script providing a shared API, \texttt{Robot.sc}.
The twin is synchronized via the \texttt{/tf} and \texttt{/joint\_states} topics, with positions and rotations converted from ROS's FLU (Front-Left-Up) coordinates to Unity's RUF (Right-Up-Front) conventions, and its links and joints serve as references for other modules.


\subsection{Data Visualizers}


\subsubsection{Egocentric: Video Stream}
\label{sec:egocentric_video_stream}

The robot's egocentric view visualizes the robot's own PoV from its head-mounted RGB-D camera, helping users understand what the robot perceives (Fig.~\ref{fig:implementation_view_image_viewer}), which can enhance decision-making, teleoperation, and trust~\cite{newbury_visualizing_2022, aubert_designing_2018}.
The compressed ROS image topic is deserialized in Unity into a 2D texture rendered on a canvas whose dimensions are computed from the camera's Field of View~(FoV) using pinhole projection, keeping the apparent angular size constant at any distance from its parent TF.


\subsubsection{Exocentric: Laser Scan}
\label{sec:exocentric_laser_scan}

The robot's exocentric data (RGB-D depth data, the base laser scan, and the global and local cost maps) is visualized through a Point Cloud Visualizer class, based on a Unity Particle System, rendering arbitrary 3D points as baked meshes relative to a parent TF.
The laser scan script filters invalid points from the raw 2D LiDAR data and passes each computed point position to the visualizer (Fig.~\ref{fig:implementation_view_laser_scan}), optionally colored by distance.


\subsubsection{Exocentric: Point Cloud}
\label{sec:exocentric_point_cloud}

The RGB-D camera's depth data is visualized from the raw point cloud byte array, converting each point to its numerical value, excluding invalid points, and passing positions and colors to the visualizer (Fig.~\ref{fig:implementation_view_point_cloud}).
If the Jet colormap is selected, each point's color is set as a gradient on the distance between 0.5~m and 1.5~m, corresponding to the robot's grasping range.


\subsubsection{Exocentric: Cost Map}
\label{sec:exocentric_cost_map}

\begin{figure}[t]
    \centering
    \begin{minipage}[t]{0.49\linewidth}
        \centering
        \frame{\includegraphics[width=\linewidth]{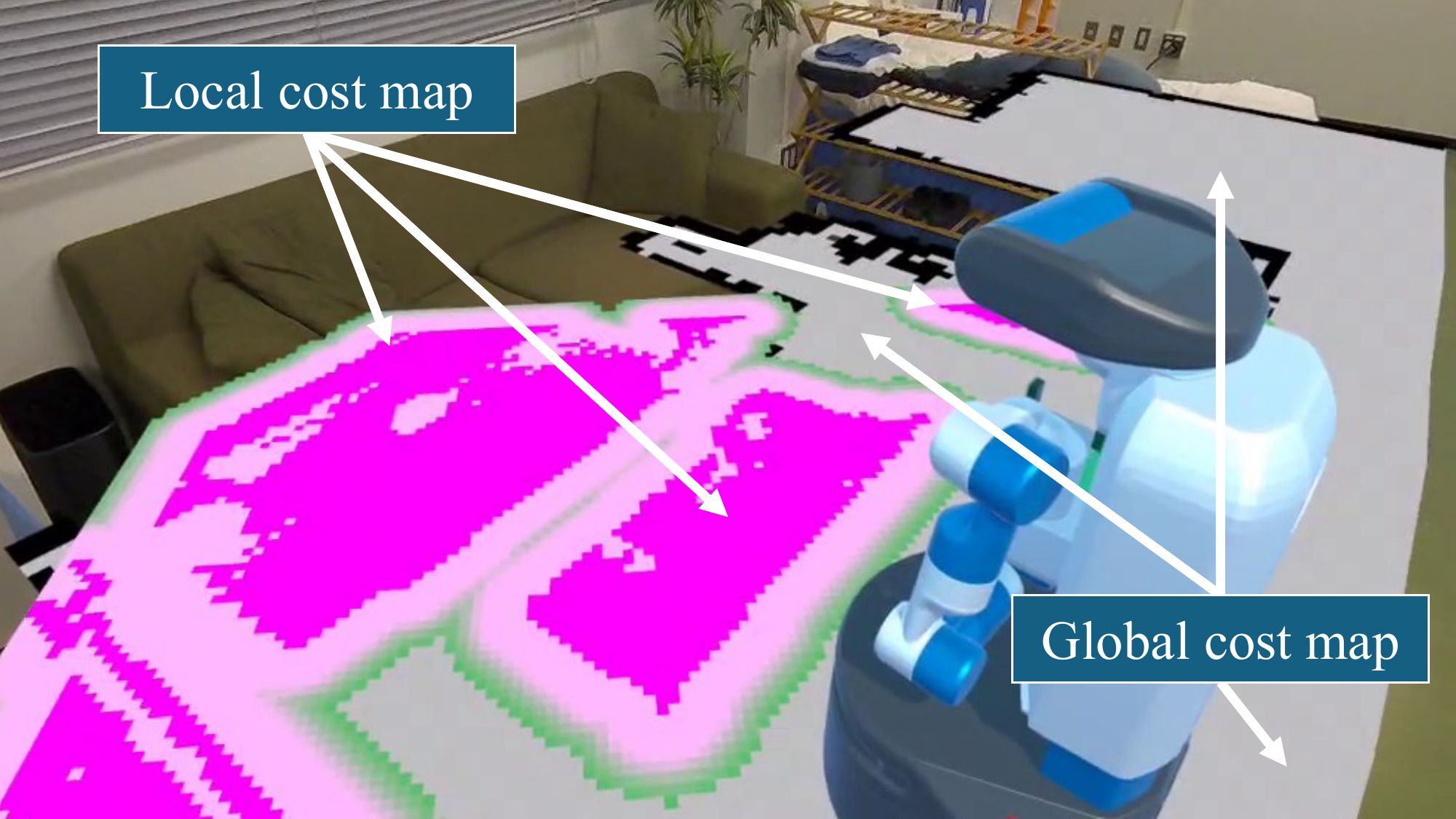}}
        \caption{
            XR HMD's PoV of the robot exocentric cost maps.
            The gray to black colors indicate the global cost map, and the purple to green color scale indicates the local cost map.
        }
        \label{fig:implementation_view_cost_map}
    \end{minipage}
    \hfill
    \begin{minipage}[t]{0.49\linewidth}
        \centering
        \frame{\includegraphics[width=\linewidth]{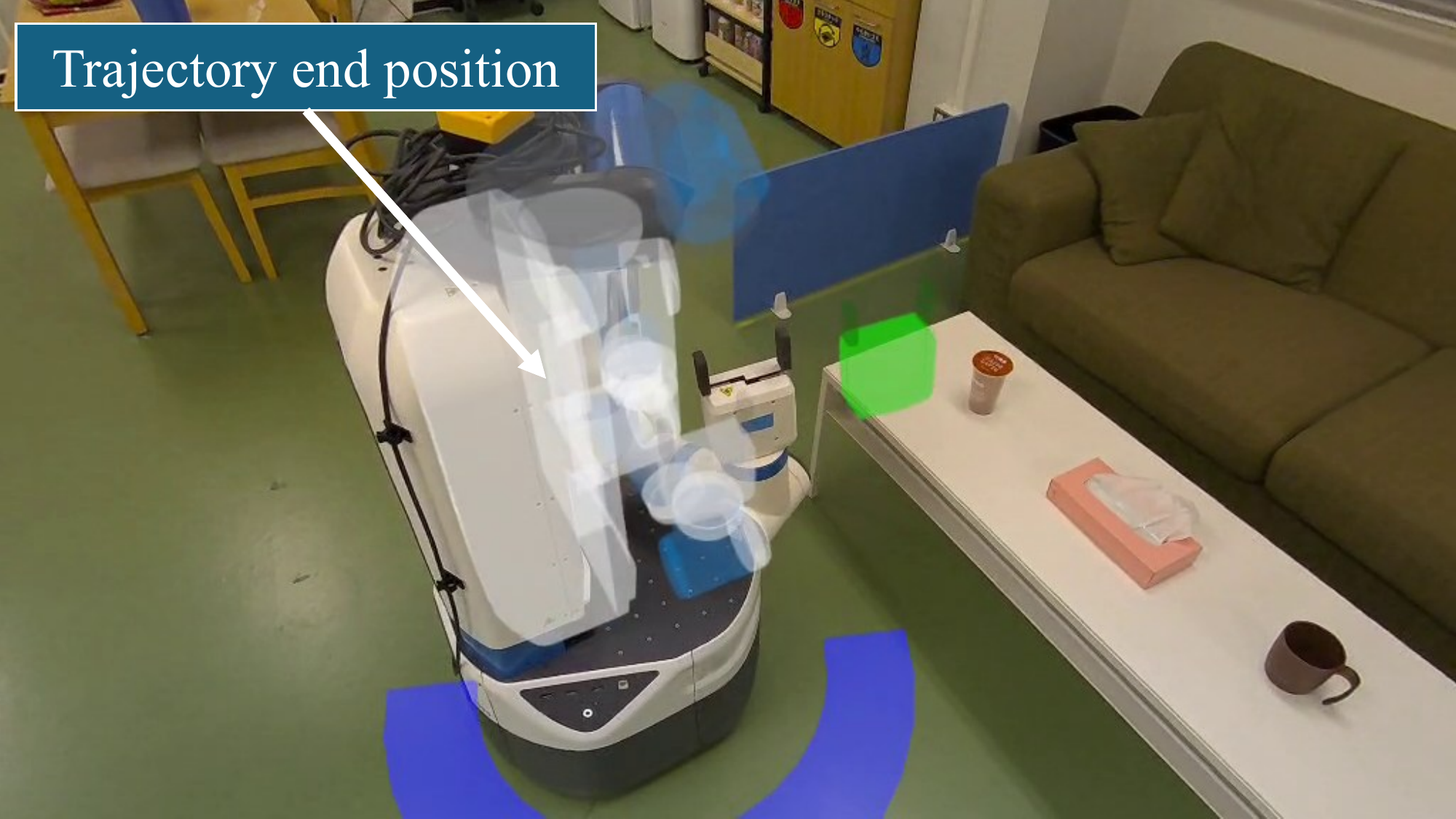}}
        \caption{
            XR HMD's PoV of a trajectory's end position as a semi-transparent digital clone.
        }
        \label{fig:implementation_view_trajectory_end_position}
    \end{minipage}

    \vspace{1em}

    \begin{minipage}[t]{0.49\linewidth}
        \centering
        \frame{\includegraphics[width=\linewidth]{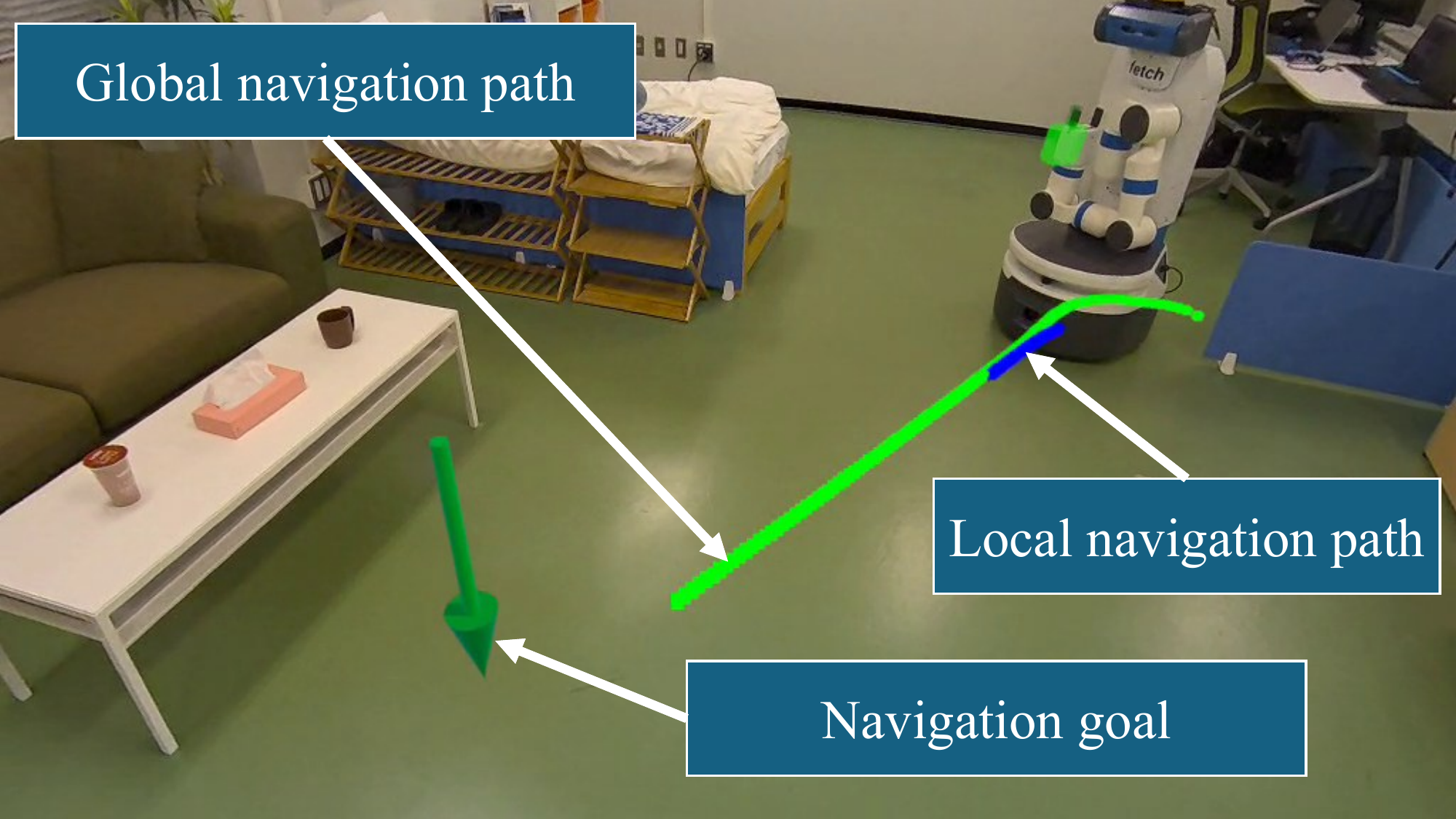}}
        \caption{
            XR HMD's PoV of the robot's trajectory goal and paths.
            The green arrow represents the position goal, the green line represents the global path, and the blue line represents the local path.
        }
        \label{fig:implementation_view_navigation}
    \end{minipage}
    \hfill
    \begin{minipage}[t]{0.49\linewidth}
        \centering
        \frame{\includegraphics[width=\linewidth]{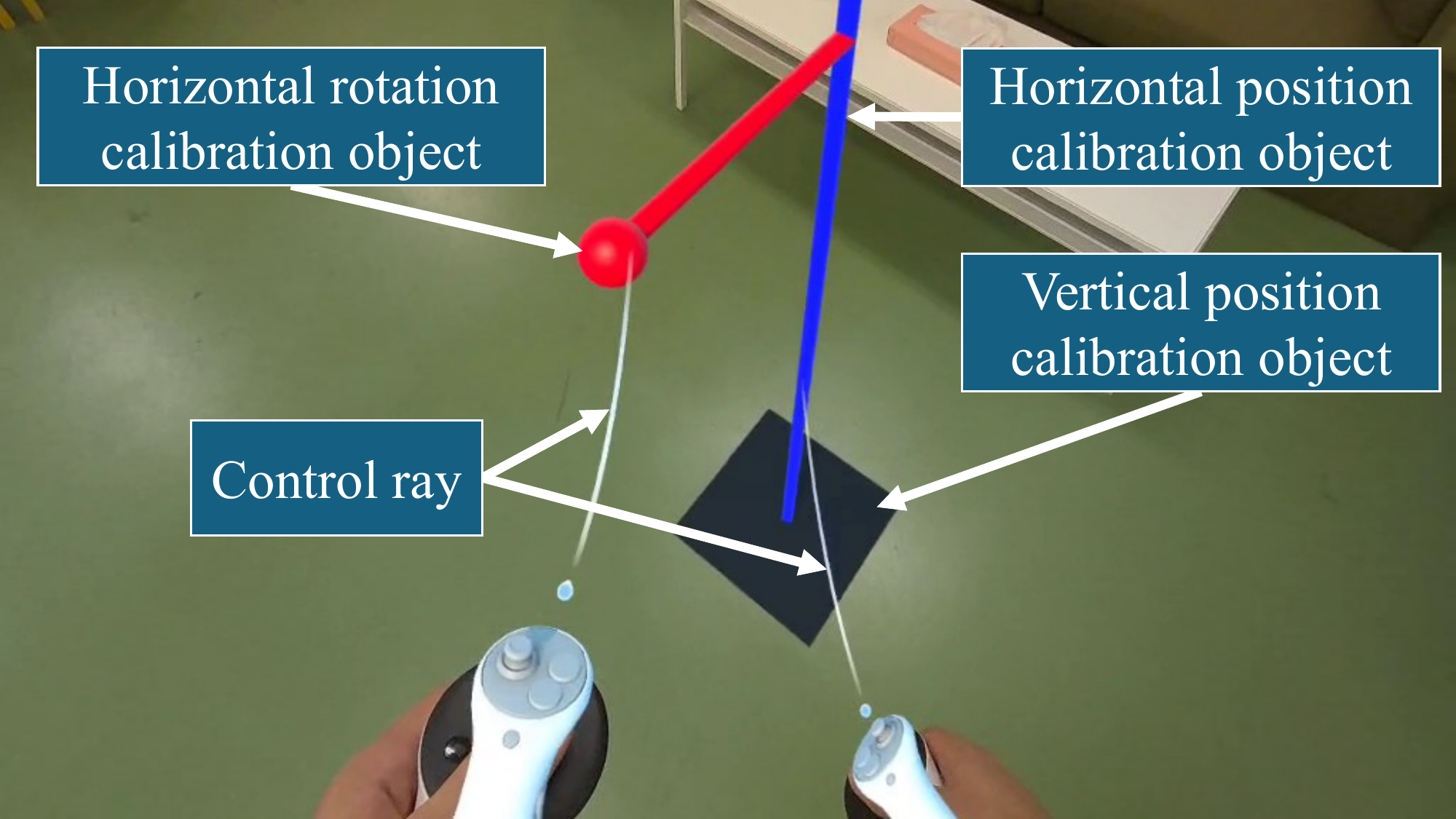}}
        \caption{
            XR HMD's PoV of the hand-eye world frame calibration objects.
        }
        \label{fig:implementation_view_calibration}
    \end{minipage}
\end{figure}

A robot's cost map is a two-dimensional collision-probability matrix central to motion planning and obstacle avoidance.
Two types are employed, the static, pre-scanned global cost map and the dynamically updated local cost map (Fig.~\ref{fig:implementation_view_cost_map}).
Upon receiving the initial cost map from ROS, the data is cropped to the bounding box enclosing all valid cells, reducing memory and rendering overhead, after which incremental updates refresh the baked mesh rendering the occupancy grid to the user.


\subsection{Intent Visualizers}


\subsubsection{Egocentric: Trajectory End Position Visualization}
\label{sec:egocentric_trajectory_end_position_visualization}

As visualization of intent enhances both user trust and the fluency of task execution in AR~\cite{newbury_visualizing_2022, maccio_mixed_2022}, the anticipated future pose of the robot's end effector is visualized as a semi-transparent robot clone (Fig.~\ref{fig:implementation_view_trajectory_end_position}), adapting~\cite{maccio_mixed_2022}: the clone is instantiated at the robot's current joint positions, updated to the final configuration of the MoveIt planner's collision-aware trajectory, then destroyed after a delay from the trajectory metadata to avoid visual clutter.


\subsubsection{Egocentric: Navigation Goal Visualization}
\label{sec:egocentric_navigation_goal_visualization}

During semi- or fully autonomous navigation, the end goal is visualized as an arrow pointing to the ground, bobbing sinusoidally.
Since the goal has no result callback, the robot's distance to the goal is checked, and the visualization is removed within a threshold distance.


\subsubsection{Exocentric: Navigation Path Visualization}

When a navigation goal is issued, the global plan, calculated once using the global cost map, defines a high-level trajectory avoiding known static obstacles, while the local plan is recalculated continuously using the local cost map to generate collision-free paths in real time.
Both plans are visualized from the \texttt{Path} messages' waypoints, each rendered as a cube, with the global plan colored green and the local plan blue (Fig.~\ref{fig:implementation_view_navigation}), inspired by RViz, enabling users to assess long-range intent and short-range adaptation simultaneously.


\subsection{Controllers}
\label{sec:controllers}

\begin{figure}[t]
    \centering
    \frame{\includegraphics[width=1.0\linewidth]{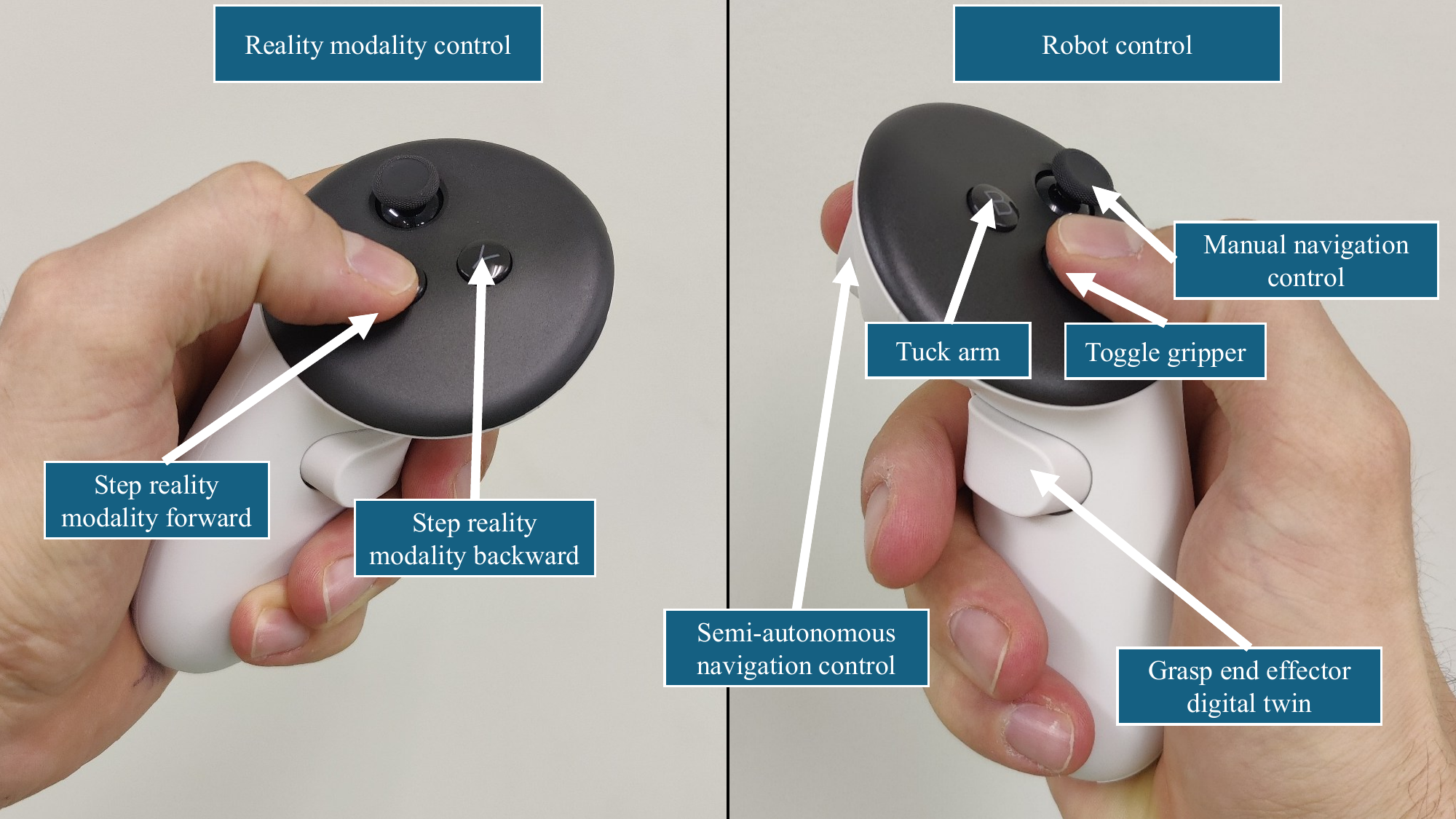}}
    \caption{
        The XR motion controller's button mapping, where the right hand controls the robot, and the left hand controls the level of reality along the RVC.
    }
    \label{fig:implementation_xr_motion_controllers_mapping}
\end{figure}

To facilitate flexible, task-specific control, a set of modular controllers is implemented, each addressing a distinct functional requirement at a different level of autonomy (Fig.~\ref{fig:implementation_xr_motion_controllers_mapping}).


\subsubsection{Gripper Controller}
\label{sec:gripper_controller}

The gripper controller actuates the end effector's fingers, managing discrete states (open, opening, closed, closing, and grasping) broadcast to the central \texttt{Robot} script.
The primary button on the right XR motion controller toggles the gripper, with the script subscribing to the ROS gripper action server's result topic to keep the toggle synchronized with the physical gripper and context-aware.


\subsubsection{Base Link Controller}
\label{sec:base_link_controller}

\begin{figure}[t]
    \centering
    \begin{minipage}[t]{0.49\linewidth}
        \centering
        \frame{\includegraphics[width=\linewidth]{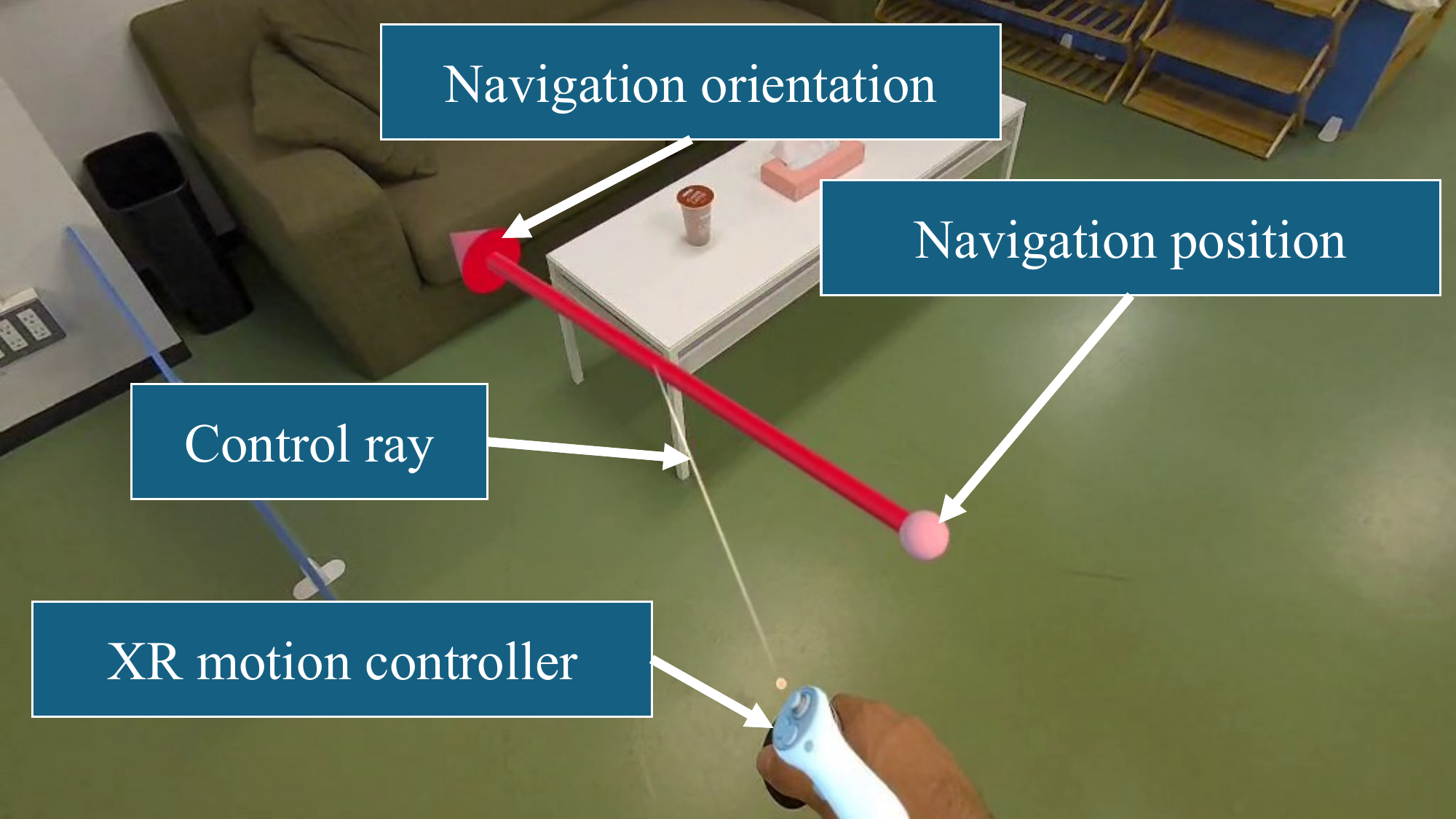}}
        \caption{
            XR HMD's PoV of a navigation's end position as an arrow, with its base corresponding to the navigation goal position, and heading direction as the navigation orientation goal.
        }
        \label{fig:implementation_view_controller_navigation}
    \end{minipage}
    \hfill
    \begin{minipage}[t]{0.49\linewidth}
        \centering
        \frame{\includegraphics[width=\linewidth]{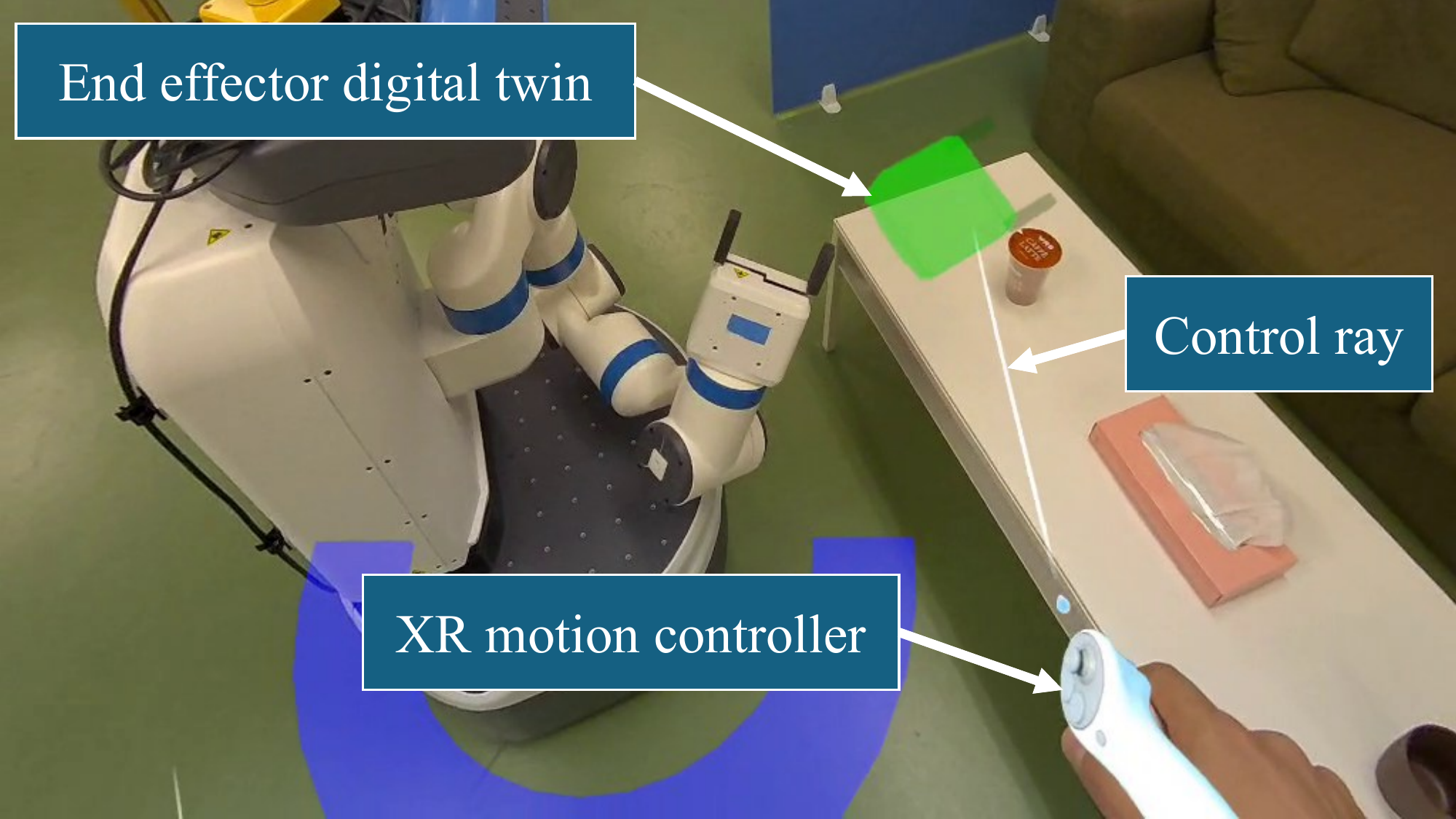}}
        \caption{
            XR HMD's PoV of the end effector digital twin.
            Once the digital twin is grasped using the XR motion controllers, the arm moves towards the pose of the digital twin.
        }
        \label{fig:implementation_view_controller_end_effector}
    \end{minipage}
\end{figure}

In addition to the semi-autonomous navigation controller of~\cite{torck_control_2025} (Fig.~\ref{fig:implementation_view_controller_navigation}), a manual base link controller enables driving the robot's mobile base with the right joystick, while rotation around the Z-axis (yaw) is intentionally disabled for safety, as unintentional rotation while the arm is extended can result in collisions.


\subsubsection{End Effector Controller}
\label{sec:end_effector_controller}

The end effector controller operates through a digital twin (Fig.~\ref{fig:implementation_view_controller_end_effector}) grasped and moved using the XR motion controllers, in two modes: in position control, releasing the twin sends its pose as the desired future pose, executed via a motion plan using inverse kinematics, while in velocity control, the end effector ``chases'' the grasped twin, halting when it is released.
Through preliminary testing, the velocity controller was favored due to its predictability and responsiveness.
Grasping the twin broadcasts the event to the central \texttt{Robot} script, setting the robot to \texttt{manipulation mode}.


\subsubsection{Head Controller}
\label{sec:head_controller}

The head controller operates automatically: in \texttt{manipulation mode}, it publishes actions orienting the robot's head towards the physical end effector, providing a consistent line of sight, essential for occluded teleoperation.
Upon release of the digital twin, the head remains fixed at the last requested position.


\subsubsection{Tuck Arm Commander}
\label{sec:tuck_arm_commander}

The tuck arm commander issues an action plan positioning the robot's arm and torso for safe navigation, as an extended arm may be erroneously detected as an obstacle and incorporated into the local occupancy map, rendering semi-autonomous navigation ineffective.
The AR planes of Section~\ref{sec:ar_plane_integration} are transformed into 3D bounding boxes, padded by 7~cm against registration inaccuracies, and transmitted to the MoveIt planner, which avoids them.


\subsection{Reality-Virtuality Manager}
\label{sec:implementation_reality_virtuality_manager}

\begin{figure}[t]
    \centering
    \begin{minipage}[t]{0.49\linewidth}
        \centering
        \frame{\includegraphics[width=\linewidth]{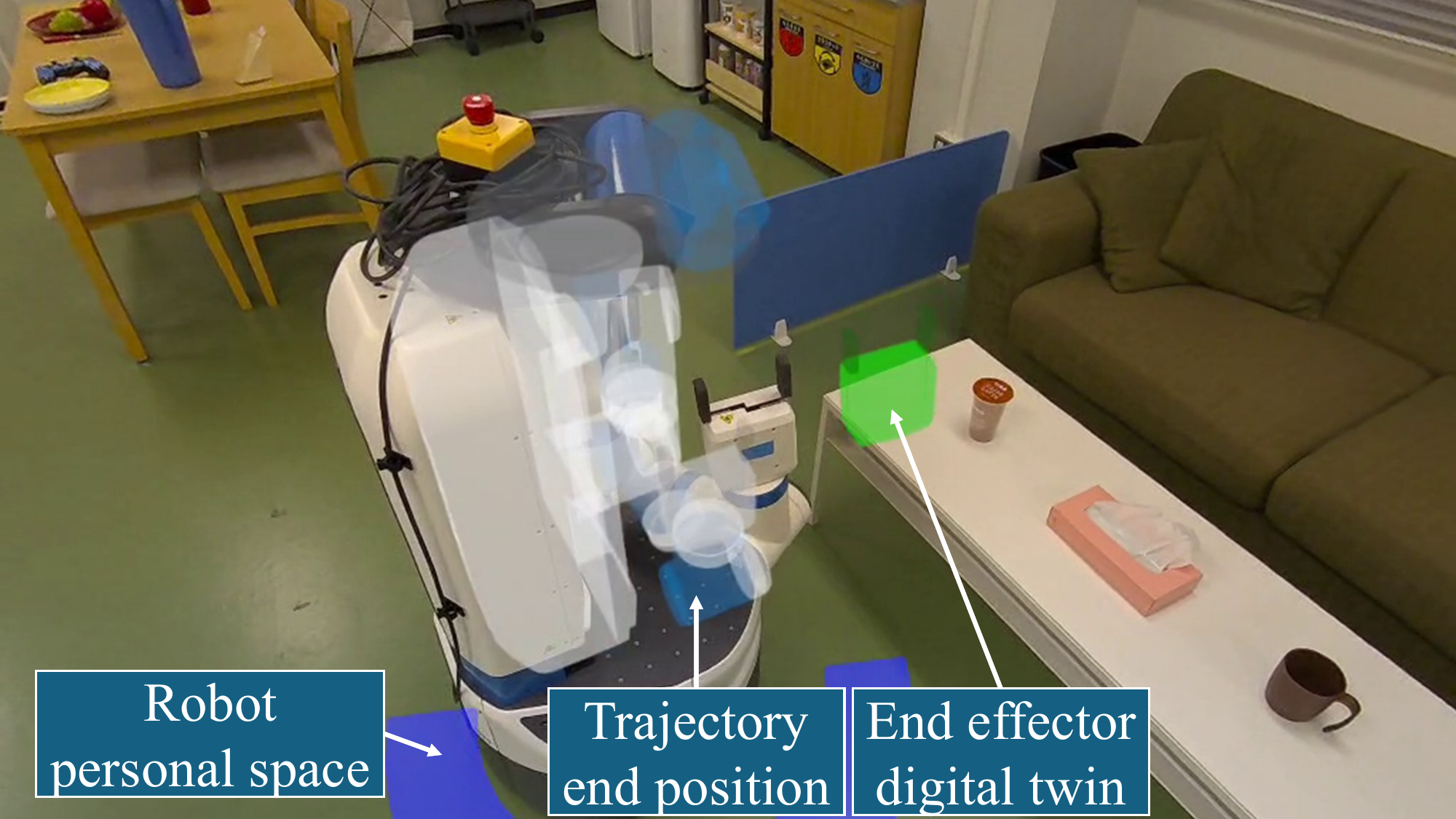}}
        \caption{
            Augmented Reality~(AR).
        }
        \label{fig:approach_rvc_augmented_reality}
    \end{minipage}
    \hfill
    \begin{minipage}[t]{0.49\linewidth}
        \centering
        \frame{\includegraphics[width=\linewidth]{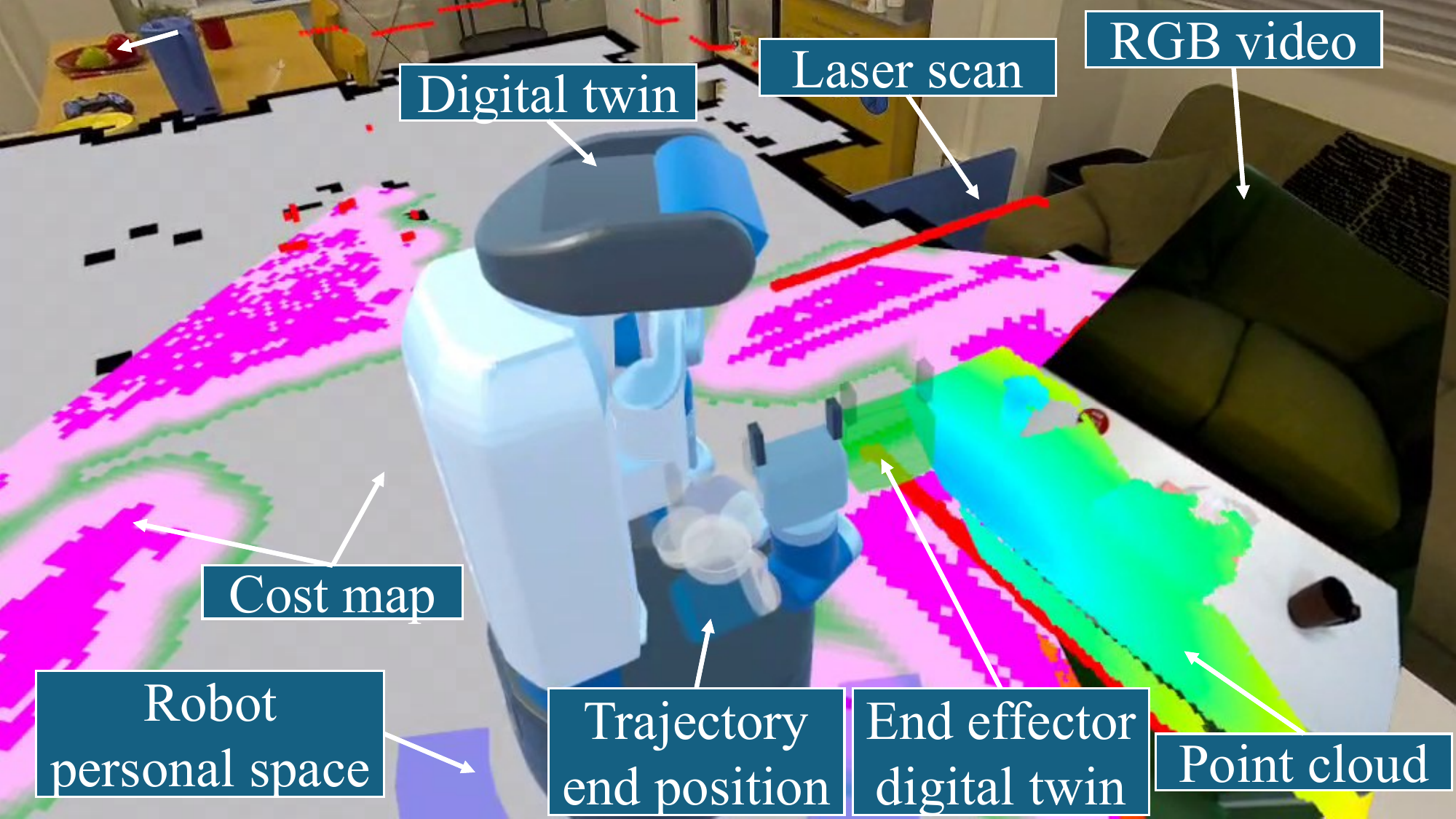}}
        \caption{
            Augmented Reality+~(AR+).
        }
        \label{fig:approach_rvc_augmented_reality_plus}
    \end{minipage}

    \vspace{1em}

    \begin{minipage}[t]{0.49\linewidth}
        \centering
        \frame{\includegraphics[width=\linewidth]{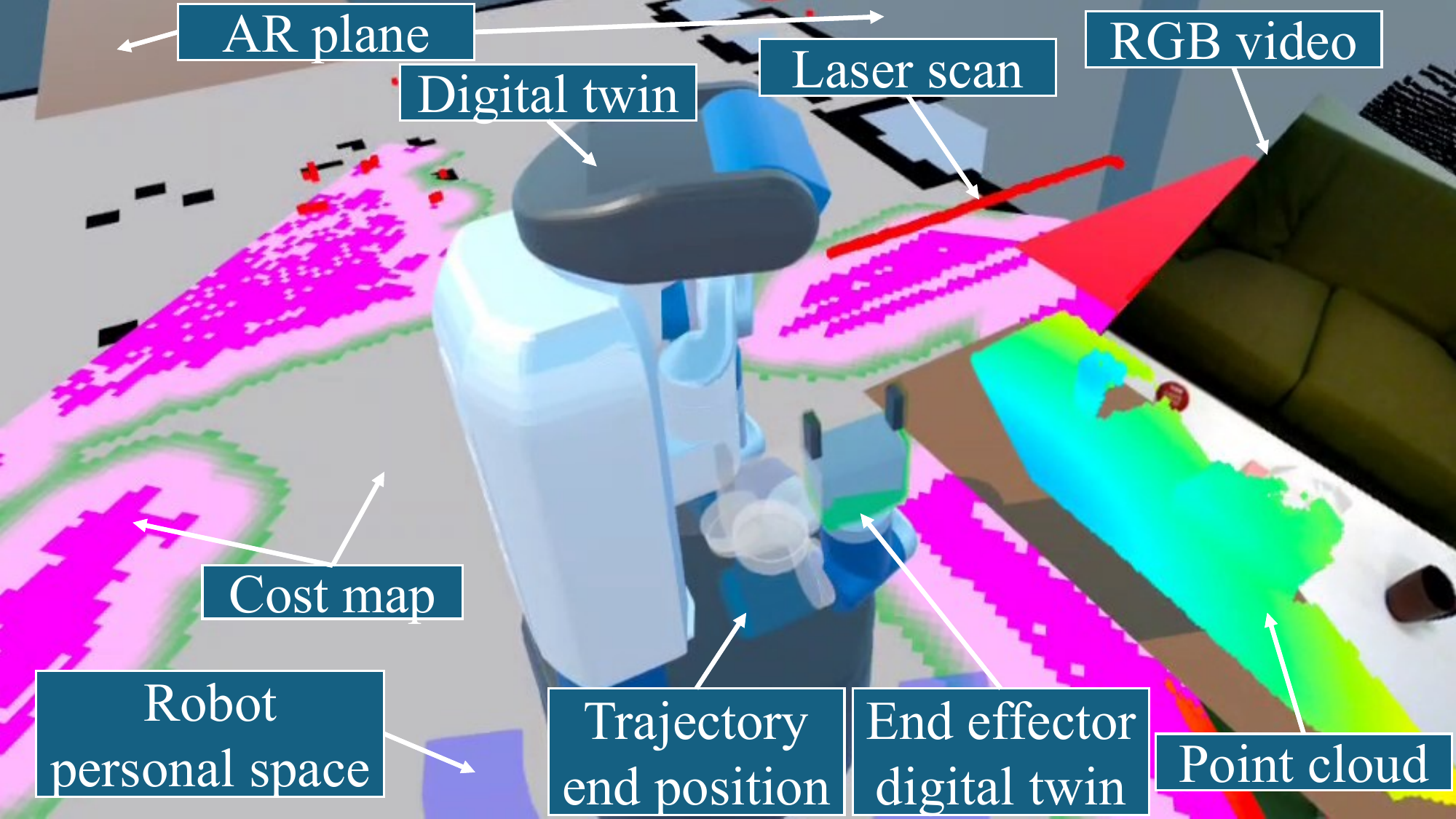}}
        \caption{
            Augmented Virtuality~(AV).
        }
        \label{fig:approach_rvc_augmented_virtuality}
    \end{minipage}
    \hfill
    \begin{minipage}[t]{0.49\linewidth}
        \centering
        \frame{\includegraphics[width=\linewidth]{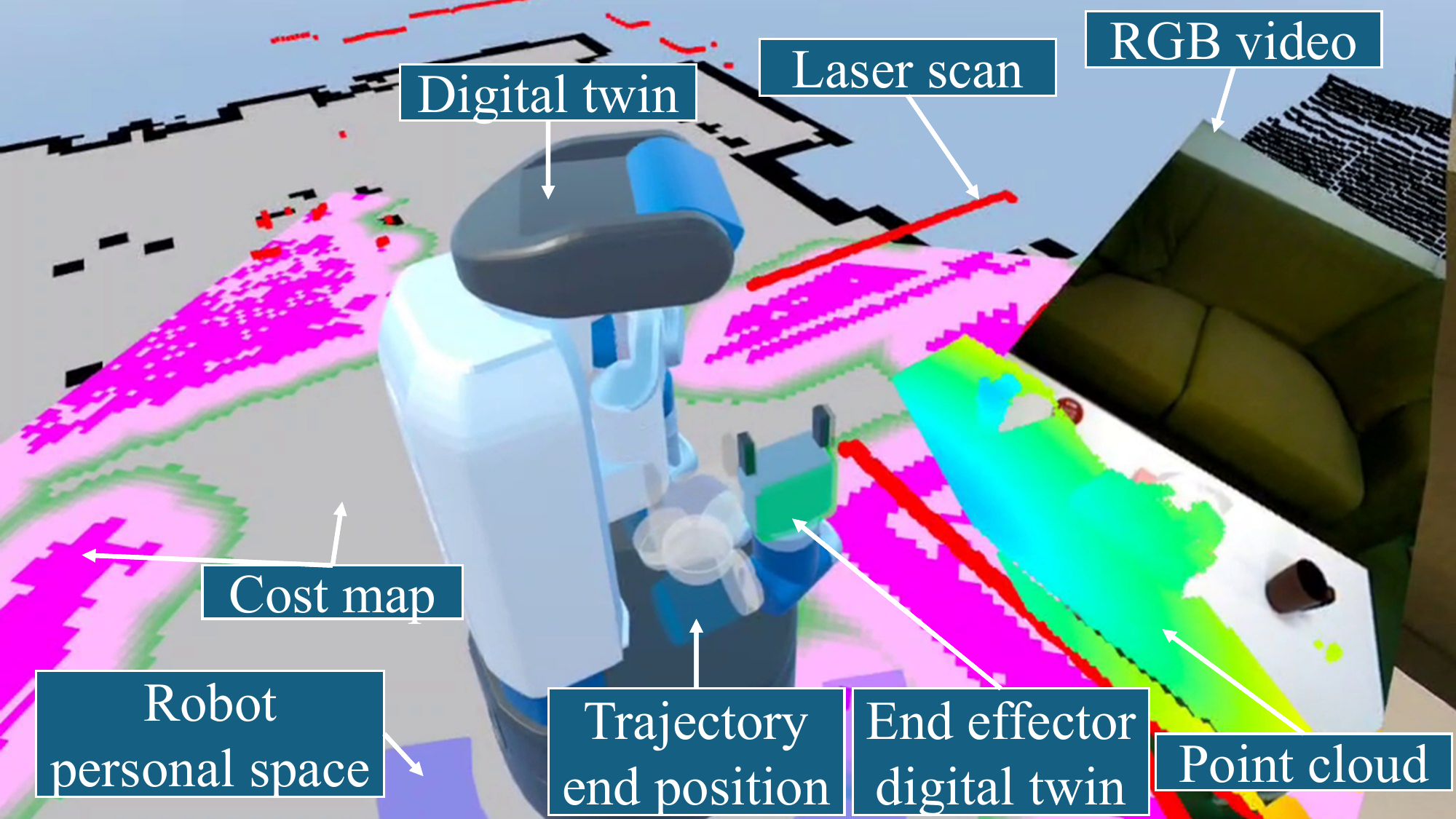}}
        \caption{
            Virtual Reality~(VR).
        }
        \label{fig:approach_rvc_virtual_reality}
    \end{minipage}
\end{figure}

All of the above visualizations together situate the user within AV.
The RVC manager therefore controls the user's current reality, implementing the five preset realities of Section~\ref{sec:reality_virtuality_manager} by showing or hiding the corresponding visualizations, as illustrated in Figs.~\ref{fig:approach_rvc_augmented_reality}, \ref{fig:approach_rvc_augmented_reality_plus}, \ref{fig:approach_rvc_augmented_virtuality}, and \ref{fig:approach_rvc_virtual_reality}.

\subsection{World Frame Calibration}
\label{sec:world_frame_calibration}

A hand-eye calibration process aligns the ROS and Unity world frames: the global static cost map is visualized in XR, and the user manually aligns it with the real world through three XR objects (Fig.~\ref{fig:implementation_view_calibration}) controlling the X-Y coordinates, the Z coordinate, and the rotation around the Z axis.
Rotation around the X and Y axes cannot be adjusted.
However, since both sides assume a horizontal floor, such misalignments are negligible.
Once aligned, the location is stored relative to one of the AR planes, and the inverse transformation restores the alignment when the application starts.


\section{Experiment}
\label{sec:experiments}

This section describes the experiment conducted to evaluate the hypotheses, starting from the preliminary experiment that motivated the choice of input method, followed by the setup, procedure, protocol, and collected data.


\subsection{Preliminary Experiment}

\begin{figure}[t]
    \centering
    \includegraphics[width=1.0\linewidth]{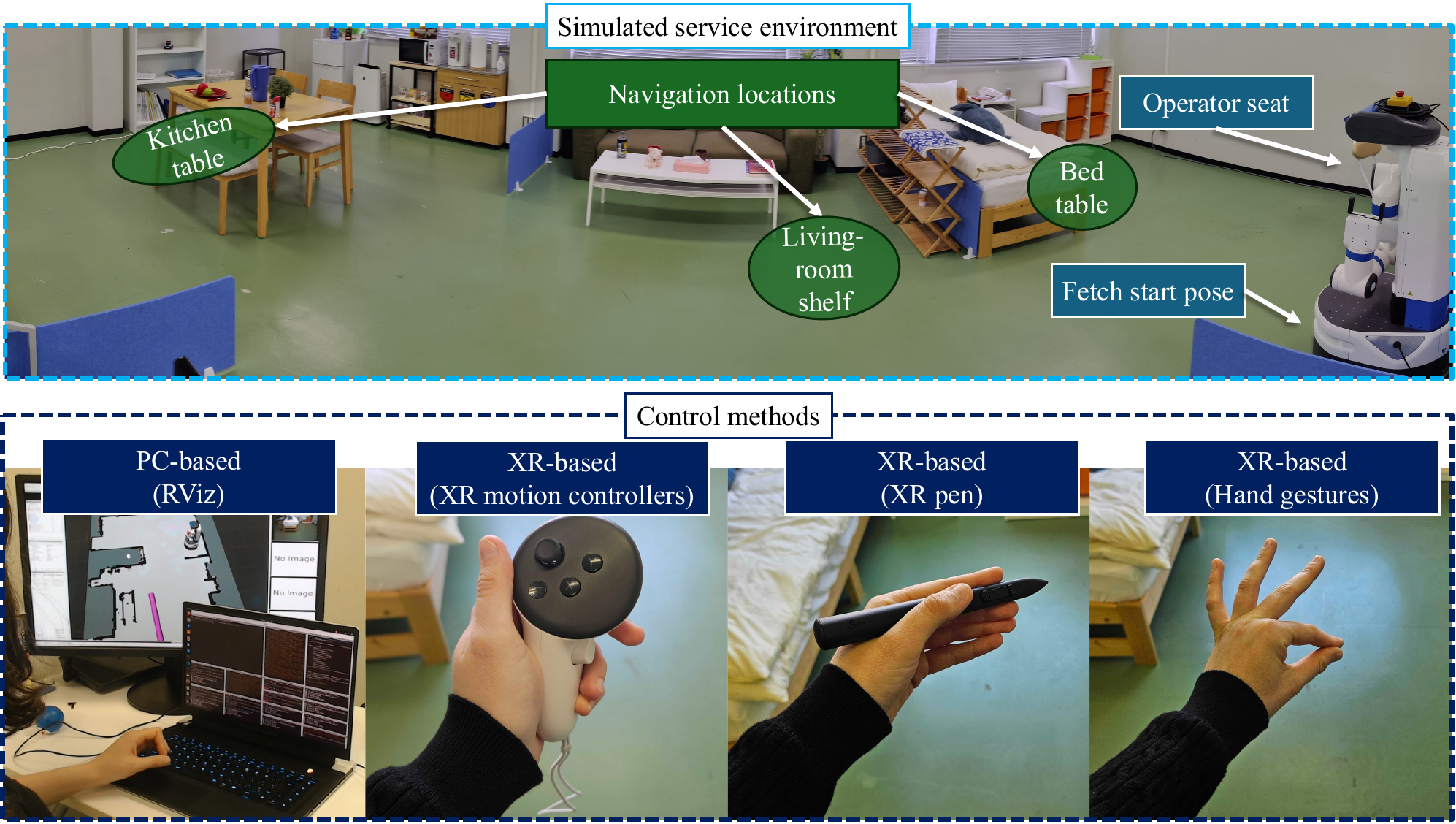}
    \caption{
        Overview of the preliminary experiment's system and controller types~\cite{torck_control_2025}, in which the participant semi-autonomously navigated the robot to three predefined target locations, each corresponding to a position where the robot could feasibly perform a simulated grasp.
    }
    \label{fig:experiment_preliminary}
\end{figure}

A preliminary experiment, reported in~\cite{torck_control_2025}, evaluated suitable controller types for navigating a robot in a semi-autonomous service environment.
It compared a traditional 2D, PC-based interface (RViz) with an XR-based interface supporting three types of XR controllers (XR motion controllers, hand gestures, and an XR pen, Fig.~\ref{fig:experiment_preliminary}), with participants navigating the robot to one of three locations for a simulated item pickup.
Quantitative metrics revealed no significant difference in task completion time across controller types, with the XR pen yielding the shortest task selection times, while the qualitative NASA-TLX~\cite{hart_development_1988} and SUS~\cite{brooke_sus_1996} results indicated a preference for the XR motion controllers.


\subsection{Setup}
\label{sec:experiment_b}

\begin{figure}[t]
    \centering
    \includegraphics[width=1.0\linewidth]{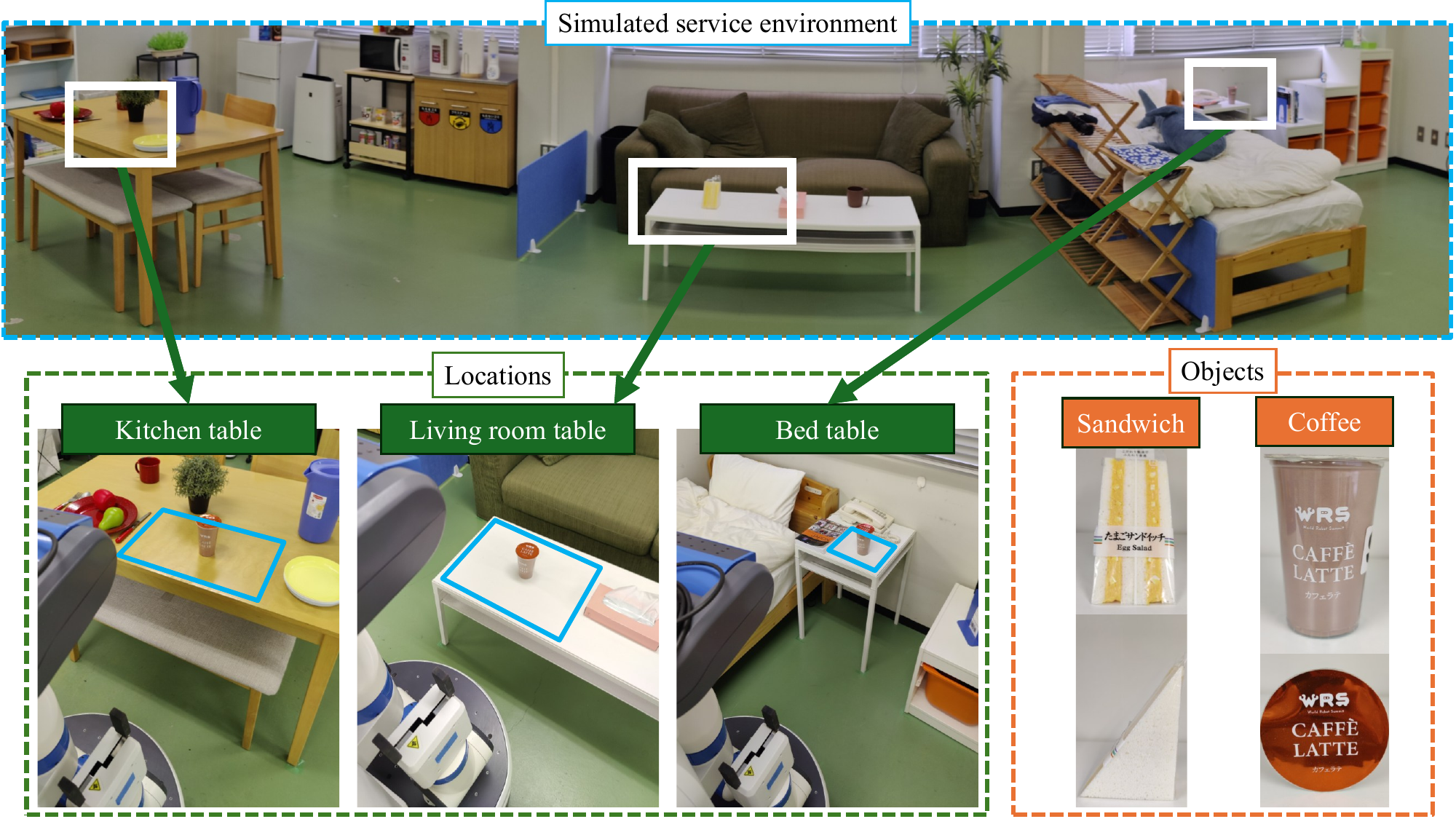}
    \caption{
        Scene overview of the experiment, showing the simulated service environment with the three locations between which participants transport the two objects.
    }
    \label{fig:experiment_scene}
\end{figure}

Participants performed a semi-autonomous pick-and-place task simulating a real-world service scenario using the XR motion controllers, first navigating the robot to a 3D $(x_{r}, y_{r}, \alpha_{r})$ location where it could reach the object, as in the preliminary experiment, then grasping the item in 6D $(x_{e}, y_{e}, z_{e}, \alpha_{e}, \beta_{e}, \gamma_{e})$ and placing it at one of the defined locations.
The participant was either restricted to a fixed reality (AR, AV, or VR) or free to change the level of immersion along the RVC, illustrated in Fig.~\ref{fig:approach_rvc}, including the intermediate step between AR and AV referred to as AR+.
Each reality condition never visualized redundant information: in AR, the camera's RGB-D feed would occlude the high-fidelity real-world view and is hidden, whereas in VR, the same data is non-redundant and visualized.

In the experiment scene, two objects (a coffee cup and a sandwich, Fig.~\ref{fig:experiment_scene}) had to be relocated between three areas: the kitchen table, the living room table, and the bedside table.
The objects were sourced from the World Robot Summit~(WRS) Future Convenience Store Challenge~(FCSC) 2025 dataset\footnote{World Robot Summit Future Convenience Store Challenge (\url{https://worldrobotsummit.org/en/wrs2025/fcsc/})} for their contrasting properties: the coffee cup is rigid and axially symmetric but contains fluid, shifting its center of gravity, while the sandwich is soft, deformable, and asymmetrical, complicating grasp planning.
Similarly, the three target locations introduce diverse constraints.
The kitchen table is cluttered, challenging the participants' robot exocentric perception~\cite{tornberg_mixed_2024,wozniak_happily_2023}, the living room table is low, requiring a downward arm extension, and the bedside table is in a constrained area, impeding maneuverability and viewing angles.


\subsection{Procedure}
\label{sec:procedure}

Each trial required the participant to perform as many successful pick-and-place cycles as possible within a 10-minute time frame, with the two objects randomly placed at two of the three locations, leaving one unoccupied as the drop-off point.
Once the object was grasped, the participant lifted the end effector away from the pickup surface, drove the robot backwards, tucked the arm, and semi-autonomously directed the robot to the empty location by pointing and pressing, before placing the object on its original face.
If the object was dropped, placed on an incorrect face, or the safety stop was activated due to an actual or potential collision, a failure was recorded.
After collisions, the participant manually moved the robot to a safe location, and after drops or misplacements, the operator reset the scene.
Each of the ten participants completed four 10-minute trials, one per XR condition in an order randomized between participants, with the RVC condition allowing dynamic adjustment using the XR motion controller buttons (Section~\ref{sec:reality_virtuality_manager}).


\subsection{Protocol}

Upon arrival, each participant completed a pre-questionnaire and received an explanation of the experiment's goals, task structure, and evaluation criteria.
Once ready, the participant was fitted with the XR headset and completed a familiarization session until comfortable, after which the trials commenced.
After each trial, participants completed the NASA-TLX~\cite{hart_development_1988} and SUS~\cite{brooke_sus_1996} questionnaires, along with an open-ended question on their experience, and after all four trials, a final questionnaire asked participants to indicate their preferred XR condition and reflect on their overall experience.
The experiments were conducted under the Ritsumeikan University Research Ethics Guidelines, established by an independent committee.
Written informed consent was obtained from all participants before the experiment, and all collected data was anonymized.


\subsection{Data}

The data was acquired in three stages.
First, the pre-questionnaire gathered demographics, vision, and health screening information to exclude individuals with limitations affecting task performance.
Second, each trial recorded the number of grasps and placed objects, object drops, objects placed on an incorrect face, collisions, and time spent manipulating and navigating, along with the modality in use during the RVC condition, followed by the post-trial questionnaires.
Finally, the post-experiment questionnaire captured the preferred XR modality ($\in$ [AR, AV, VR, RVC]) and an open-ended reflection on the overall experience.


\section{Results}
\label{sec:results}

For each dependent variable of the experiment, a one-way repeated-measures ANOVA was conducted, with normality of residuals assessed via the Shapiro-Wilk test and sphericity via Mauchly's test.
When violations of sphericity were detected ($p \leq 0.05$), the Greenhouse-Geisser correction was applied, and significant main effects were followed by paired t-tests with Holm-Bonferroni correction, with the significance level set at $\alpha = 0.05$.


\subsection{Fixed Reality Modality}


\subsubsection{Time Navigating}

\begin{figure}[H]
    \centering
    \includegraphics[width=0.50\linewidth]{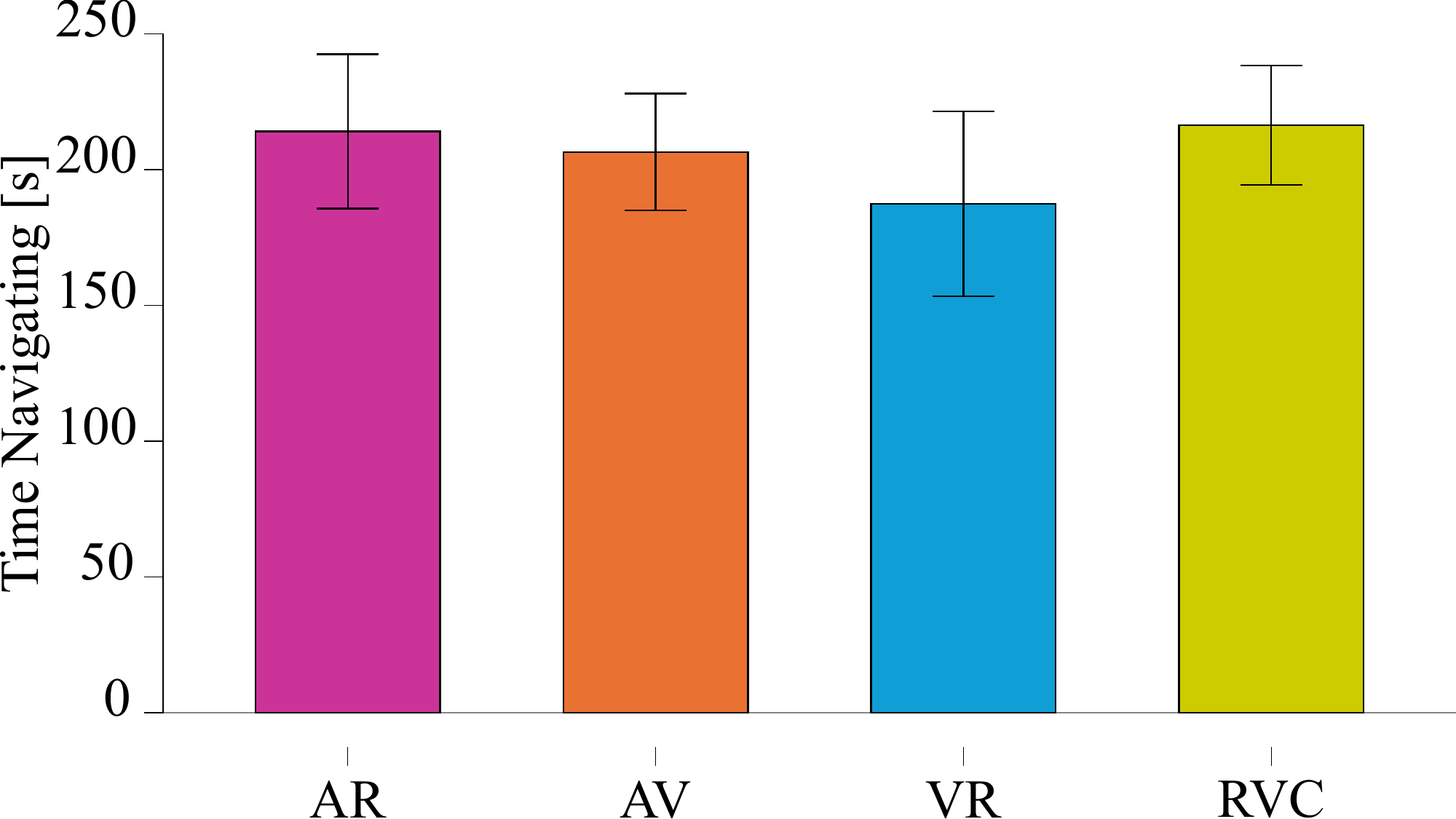}
    \caption{
        Mean Time Navigating (in seconds) with 95\% confidence intervals across conditions.
        No statistically significant differences were found between conditions.
    }
    \label{fig:results_fixed_time_navigating}
\end{figure}

No statistically significant main effect of condition on navigation time was found, $F(3, 27) = 1.217$, $p = .323$ (Tables~\ref{tab:results_fixed_time_navigating_within_subjects} and \ref{tab:results_fixed_time_navigating_descriptives}, Fig.~\ref{fig:results_fixed_time_navigating}).

\begin{table}[H]
    \centering
    \footnotesize
    \caption{Within-Subjects Analysis of Variance for Time Spent Navigating Across Conditions}
    \label{tab:results_fixed_time_navigating_within_subjects}
    {
        \begin{tabularx}{1.0\linewidth}{lrRRRR}
            \toprule
            Cases & Sum of Squares & df & Mean Square & F & p \\
            \cmidrule[0.4pt]{1-6}
            Time Navigating [s] & $5199.066$ & $3$ & $1733.022$ & $1.217$ & $0.323$ \\
            Residuals & $38460.168$ & $27$ & $1424.451$ & $ $ & $ $ \\
            \bottomrule
        \end{tabularx}
    }
\end{table}

\begin{table}[H]
    \centering
    \footnotesize
    \caption{Descriptive Statistics for Time Spent Navigating Across Conditions}
    \label{tab:results_fixed_time_navigating_descriptives}
    {
        \begin{tabularx}{1.0\linewidth}{lRRRRr}
            \toprule
            Time Navigating [s] & N & Mean & SD & SE & Coefficient of variation \\
            \cmidrule[0.4pt]{1-6}
            AR & $10$ & $214.106$ & $35.848$ & $11.336$ & $0.167$ \\
            AV & $10$ & $206.549$ & $36.248$ & $11.463$ & $0.175$ \\
            VR & $10$ & $187.368$ & $56.075$ & $17.732$ & $0.299$ \\
            RVC & $10$ & $216.334$ & $29.030$ & $9.180$ & $0.134$ \\
            \bottomrule
        \end{tabularx}
    }
\end{table}


\subsubsection{Time Manipulating}

\begin{figure}[H]
    \centering
    \includegraphics[width=0.50\linewidth]{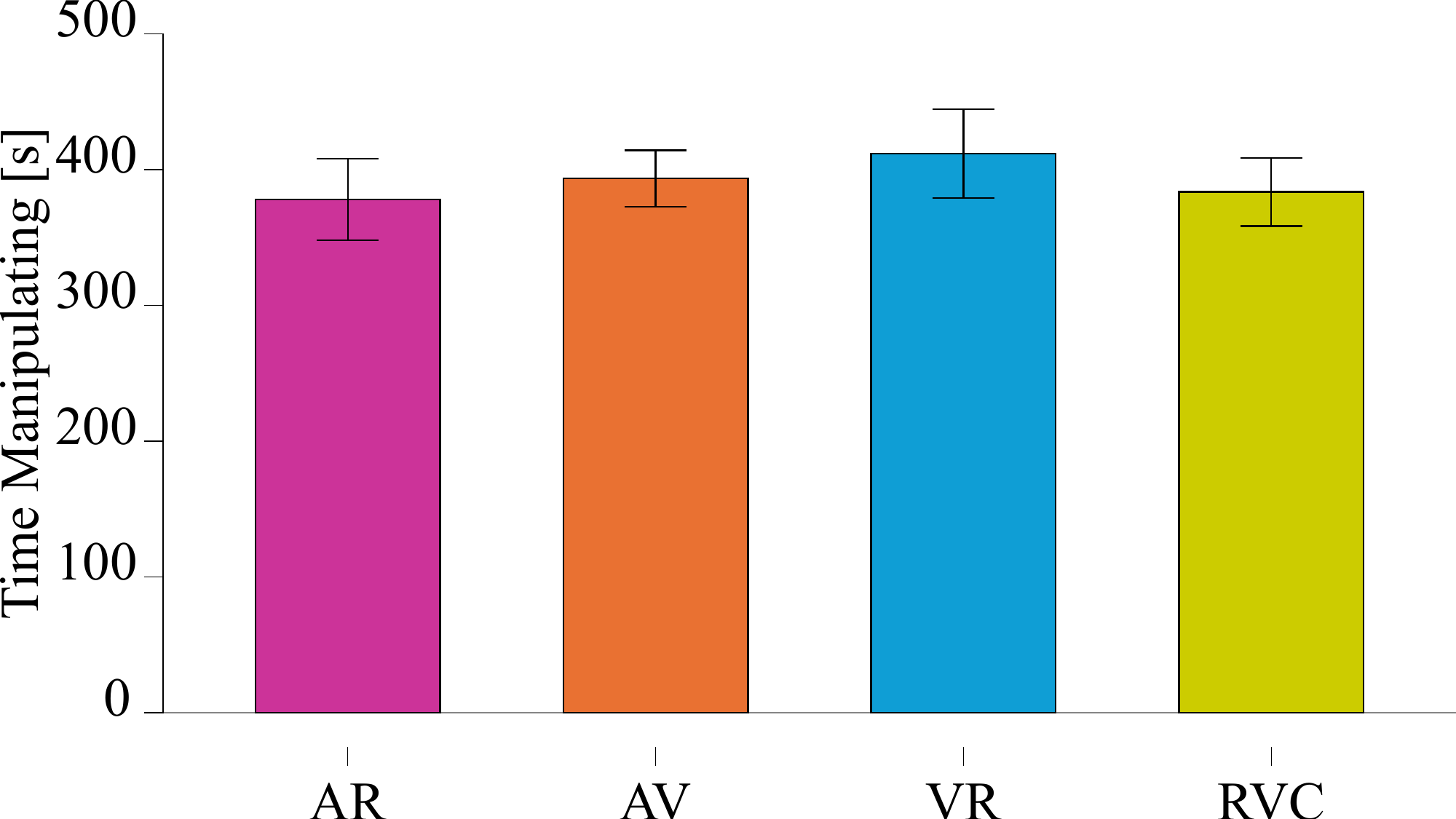}
    \caption{
        Mean Time Manipulating (in seconds) with 95\% confidence intervals across conditions.
        No statistically significant differences were found between conditions.
    }
    \label{fig:results_fixed_time_manipulating}
\end{figure}

No statistically significant differences in manipulation time were found across conditions, $F(3, 27) = 1.493$, $p = .239$ (Tables~\ref{tab:results_fixed_time_manipulating_within_subjects} and \ref{tab:results_fixed_time_manipulating_descriptives}, Fig.~\ref{fig:results_fixed_time_manipulating}).

\begin{table}[H]
    \centering
    \footnotesize
    \caption{Within-Subjects Analysis of Variance for Time Spent Manipulating Across Conditions}
    \label{tab:results_fixed_time_manipulating_within_subjects}
    {
        \begin{tabularx}{1.0\linewidth}{lrRRRR}
            \toprule
            Cases & Sum of Squares & df & Mean Square & F & p \\
            \cmidrule[0.4pt]{1-6}
            Time Manipulating [s] & $6640.947$ & $3$ & $2213.649$ & $1.493$ & $0.239$ \\
            Residuals & $40025.749$ & $27$ & $1482.435$ & $ $ & $ $ \\
            \bottomrule
        \end{tabularx}
    }
\end{table}

\begin{table}[H]
    \centering
    \footnotesize
    \caption{Descriptive Statistics for Time Spent Manipulating Across Conditions}
    \label{tab:results_fixed_time_manipulating_descriptives}
    {
        \begin{tabularx}{1.0\linewidth}{lRRRRr}
            \toprule
            Time Manipulating [s] & N & Mean & SD & SE & Coefficient of variation \\
            \cmidrule[0.4pt]{1-6}
            AR & $10$ & $377.974$ & $34.631$ & $10.951$ & $0.092$ \\
            AV & $10$ & $393.451$ & $36.248$ & $11.463$ & $0.092$ \\
            VR & $10$ & $411.901$ & $55.562$ & $17.570$ & $0.135$ \\
            RVC & $10$ & $383.666$ & $29.030$ & $9.180$ & $0.076$ \\
            \bottomrule
        \end{tabularx}
    }
\end{table}


\subsubsection{Objects Grasped}

\begin{figure}[H]
    \centering
    \includegraphics[width=0.50\linewidth]{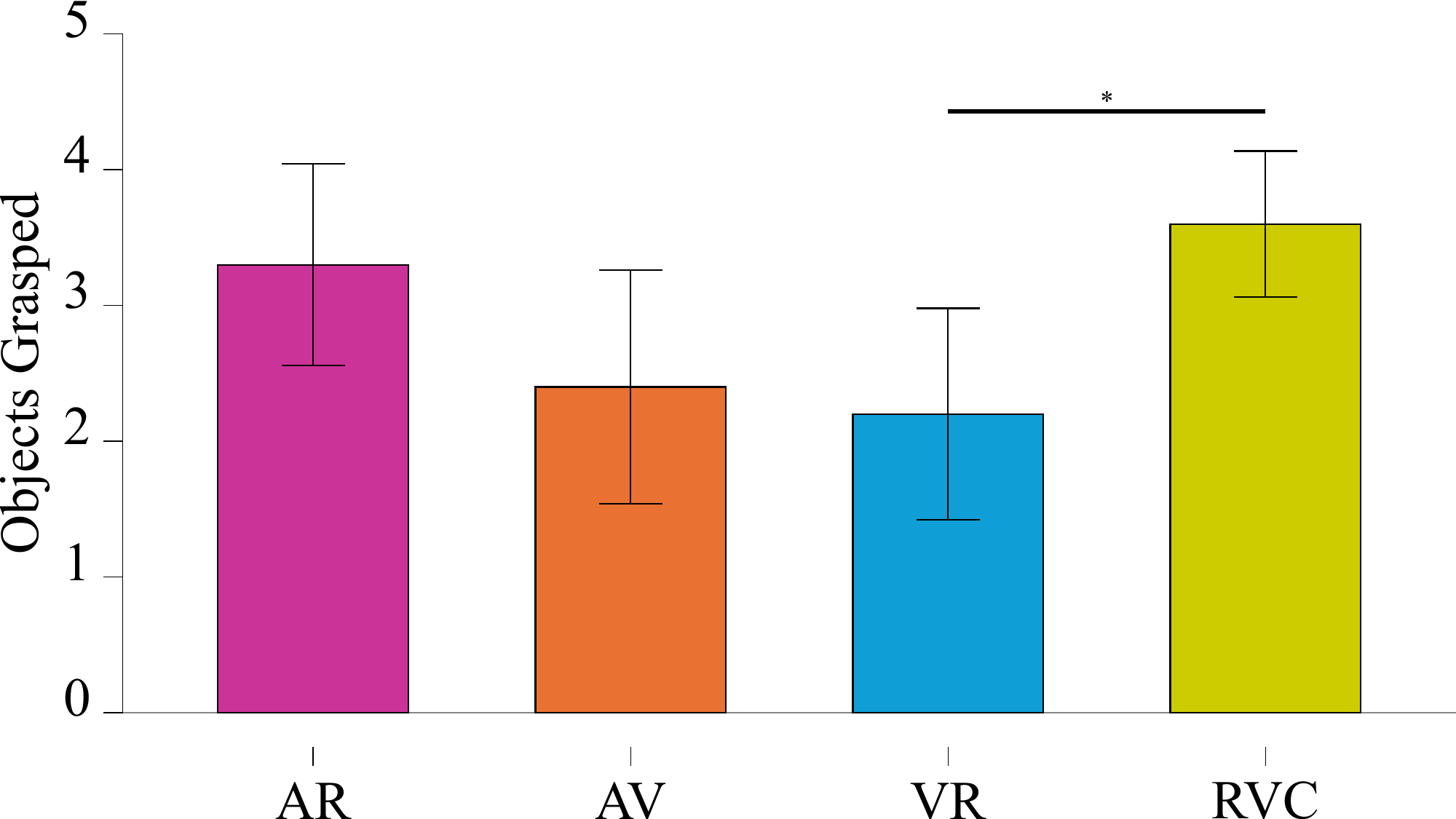}
    \caption{
        Mean Objects Grasped with 95\% confidence intervals across conditions.
        A statistically significant main effect was found, and pairwise differences are denoted by $*$.
    }
    \label{fig:results_fixed_objects_grasped}
\end{figure}

A statistically significant main effect of condition on the number of objects grasped was found, $F(3, 27) = 4.325$, $p = .013$ (Table~\ref{tab:results_fixed_objects_grasped_within_subjects}, Fig.~\ref{fig:results_fixed_objects_grasped}).
The mean number of grasped objects was highest in RVC ($M = 3.600$), followed by AR ($M = 3.300$), AV ($M = 2.400$), and VR ($M = 2.200$) (Table~\ref{tab:results_fixed_objects_grasped_descriptives}).
Holm-corrected post hoc comparisons (Table~\ref{tab:results_fixed_objects_grasped_post_hoc}) revealed a significant difference between VR and RVC, $t(9) = -3.500$, $p_{holm} = .040$, with no other contrasts reaching significance.

\begin{table}[H]
    \centering
    \footnotesize
    \caption{Within-Subjects Analysis of Variance for Objects Grasped Across Conditions}
    \label{tab:results_fixed_objects_grasped_within_subjects}
    {
        \begin{tabularx}{1.0\linewidth}{lRRRRR}
            \toprule
            Cases & Sum of Squares & df & Mean Square & F & p \\
            \cmidrule[0.4pt]{1-6}
            Objects Grasped & $13.875$ & $3$ & $4.625$ & $4.325$ & $0.013$ \\
            Residuals & $28.875$ & $27$ & $1.069$ & $ $ & $ $ \\
            \bottomrule
        \end{tabularx}
    }
\end{table}

\begin{table}[H]
    \centering
    \footnotesize
    \caption{Descriptive Statistics for Objects Grasped Across Conditions}
    \label{tab:results_fixed_objects_grasped_descriptives}
    {
        \begin{tabularx}{1.0\linewidth}{lRRRRr}
            \toprule
            Objects Grasped & N & Mean & SD & SE & Coefficient of variation \\
            \cmidrule[0.4pt]{1-6}
            AR & $10$ & $3.300$ & $1.252$ & $0.396$ & $0.379$ \\
            AV & $10$ & $2.400$ & $0.966$ & $0.306$ & $0.403$ \\
            VR & $10$ & $2.200$ & $1.398$ & $0.442$ & $0.636$ \\
            RVC & $10$ & $3.600$ & $1.075$ & $0.340$ & $0.299$ \\
            \bottomrule
        \end{tabularx}
    }
\end{table}

\begin{table}[H]
    \centering
    \footnotesize
    \caption{Post Hoc Comparisons for Objects Grasped Across Conditions}
    \label{tab:results_fixed_objects_grasped_post_hoc}
    {
        \begin{tabularx}{1.0\linewidth}{llrRRRR}
            \toprule
            $ $ & $ $ & Mean Difference & SE & df & t & p$_{holm}$ \\
            \cmidrule[0.4pt]{1-7}
            AR & AV & $0.900$ & $0.526$ & $9$ & $1.711$ & $0.364$ \\
            $ $ & VR & $1.100$ & $0.433$ & $9$ & $2.538$ & $0.127$ \\
            & RVC & $-0.300$ & $0.423$ & $9$ & $-0.709$ & $0.992$ \\
            AV & VR & $0.200$ & $0.573$ & $9$ & $0.349$ & $0.992$ \\
            $ $ & RVC & $-1.200$ & $0.389$ & $9$ & $-3.087$ & $0.065$ \\
            VR & RVC & $-1.400$ & $0.400$ & $9$ & $-3.500$ & $0.040$ \\
            \bottomrule
        \end{tabularx}
    }
\end{table}


\subsubsection{Objects Placed}

\begin{figure}[H]
    \centering
    \includegraphics[width=0.50\linewidth]{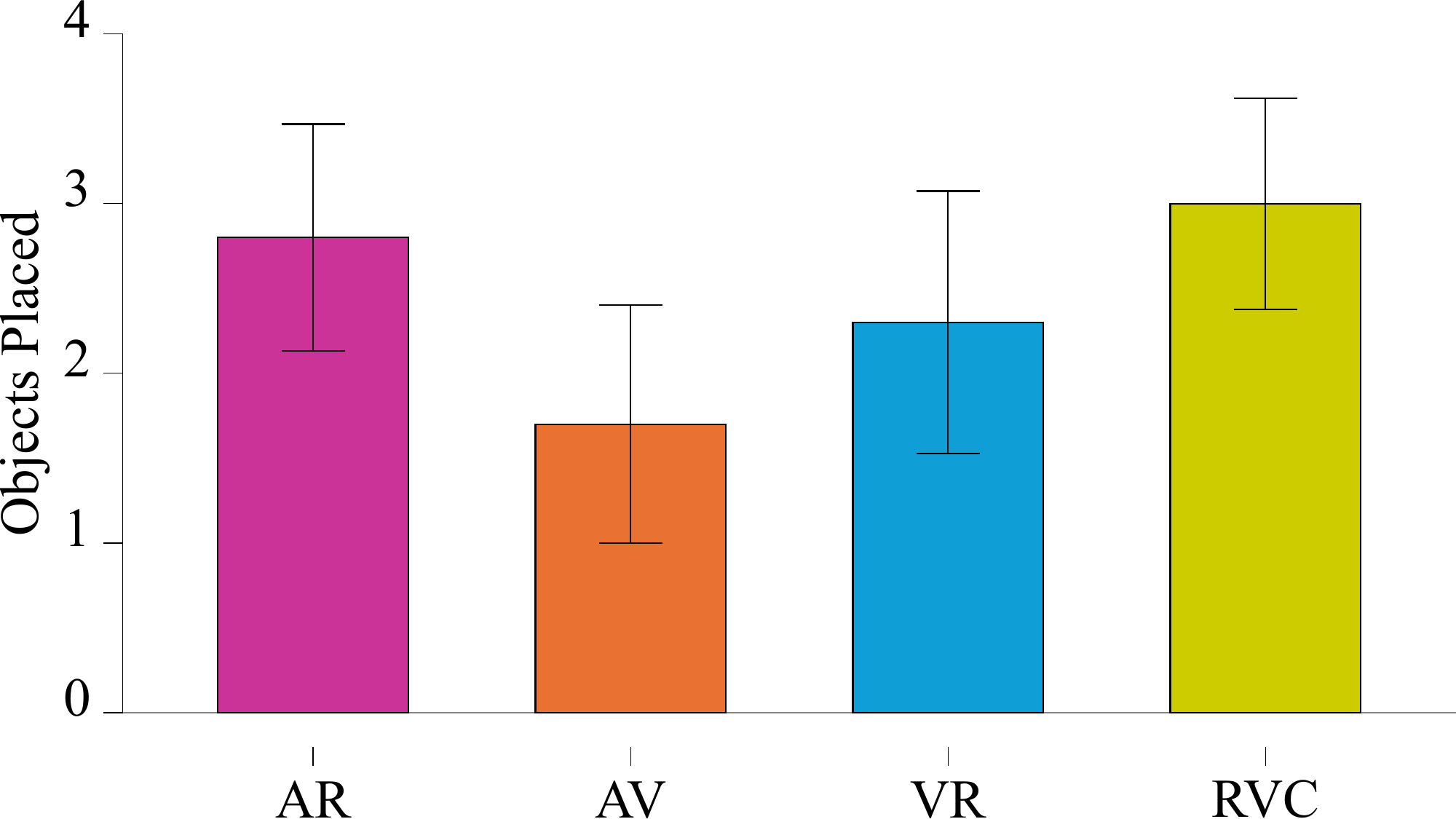}
    \caption{
        Mean Objects Placed with 95\% confidence intervals across conditions.
        A statistically significant main effect was found, though no pairwise differences reached significance.
    }
    \label{fig:results_fixed_objects_placed}
\end{figure}

A statistically significant main effect of condition on the number of objects placed was found, $F(3, 27) = 3.579$, $p = .027$ (Table~\ref{tab:results_fixed_objects_placed_within_subjects}, Fig.~\ref{fig:results_fixed_objects_placed}).
The highest mean was achieved under RVC ($M = 3.000$), followed by AR ($M = 2.800$), VR ($M = 2.300$), and AV ($M = 1.700$) (Table~\ref{tab:results_fixed_objects_placed_descriptives}).
Despite the main effect, Holm-corrected post hoc comparisons (Table~\ref{tab:results_fixed_objects_placed_post_hoc}) revealed no significant pairwise differences, the largest being between AV and RVC, $t(9) = -2.899$, $p_{holm} = .106$.

\begin{table}[H]
    \centering
    \footnotesize
    \caption{Within-Subjects Analysis of Variance for Objects Placed Across Conditions}
    \label{tab:results_fixed_objects_placed_within_subjects}
    {
        \begin{tabularx}{1.0\linewidth}{lRRRRR}
            \toprule
            Cases & Sum of Squares & df & Mean Square & F & p \\
            \cmidrule[0.4pt]{1-6}
            Objects Placed & $10.100$ & $3$ & $3.367$ & $3.579$ & $0.027$ \\
            Residuals & $25.400$ & $27$ & $0.941$ & $ $ & $ $ \\
            \bottomrule
        \end{tabularx}
    }
\end{table}

\begin{table}[H]
    \centering
    \footnotesize
    \caption{Descriptive Statistics for Objects Placed Across Conditions}
    \label{tab:results_fixed_objects_placed_descriptives}
    {
        \begin{tabularx}{1.0\linewidth}{lRRRRr}
            \toprule
            Objects Placed & N & Mean & SD & SE & Coefficient of variation \\
            \cmidrule[0.4pt]{1-6}
            AR & $10$ & $2.800$ & $1.135$ & $0.359$ & $0.405$ \\
            AV & $10$ & $1.700$ & $0.949$ & $0.300$ & $0.558$ \\
            VR & $10$ & $2.300$ & $1.059$ & $0.335$ & $0.461$ \\
            RVC & $10$ & $3.000$ & $1.054$ & $0.333$ & $0.351$ \\
            \bottomrule
        \end{tabularx}
    }
\end{table}

\begin{table}[H]
    \centering
    \footnotesize
    \caption{Post Hoc Comparisons for Objects Placed Across Conditions}
    \label{tab:results_fixed_objects_placed_post_hoc}
    {
        \begin{tabularx}{1.0\linewidth}{llrRRRR}
            \toprule
            $ $ & $ $ & Mean Difference & SE & df & t & p$_{holm}$ \\
            \cmidrule[0.4pt]{1-7}
            AR & AV & $1.100$ & $0.407$ & $9$ & $2.703$ & $0.121$ \\
            $ $ & VR & $0.500$ & $0.500$ & $9$ & $1.000$ & $0.687$ \\
            & RVC & $-0.200$ & $0.359$ & $9$ & $-0.557$ & $0.687$ \\
            AV & VR & $-0.600$ & $0.452$ & $9$ & $-1.327$ & $0.652$ \\
            $ $ & RVC & $-1.300$ & $0.448$ & $9$ & $-2.899$ & $0.106$ \\
            VR & RVC & $-0.700$ & $0.423$ & $9$ & $-1.655$ & $0.529$ \\
            \bottomrule
        \end{tabularx}
    }
\end{table}


\subsubsection{Objects Placed on Incorrect Face}

\begin{figure}[H]
    \centering
    \includegraphics[width=0.50\linewidth]{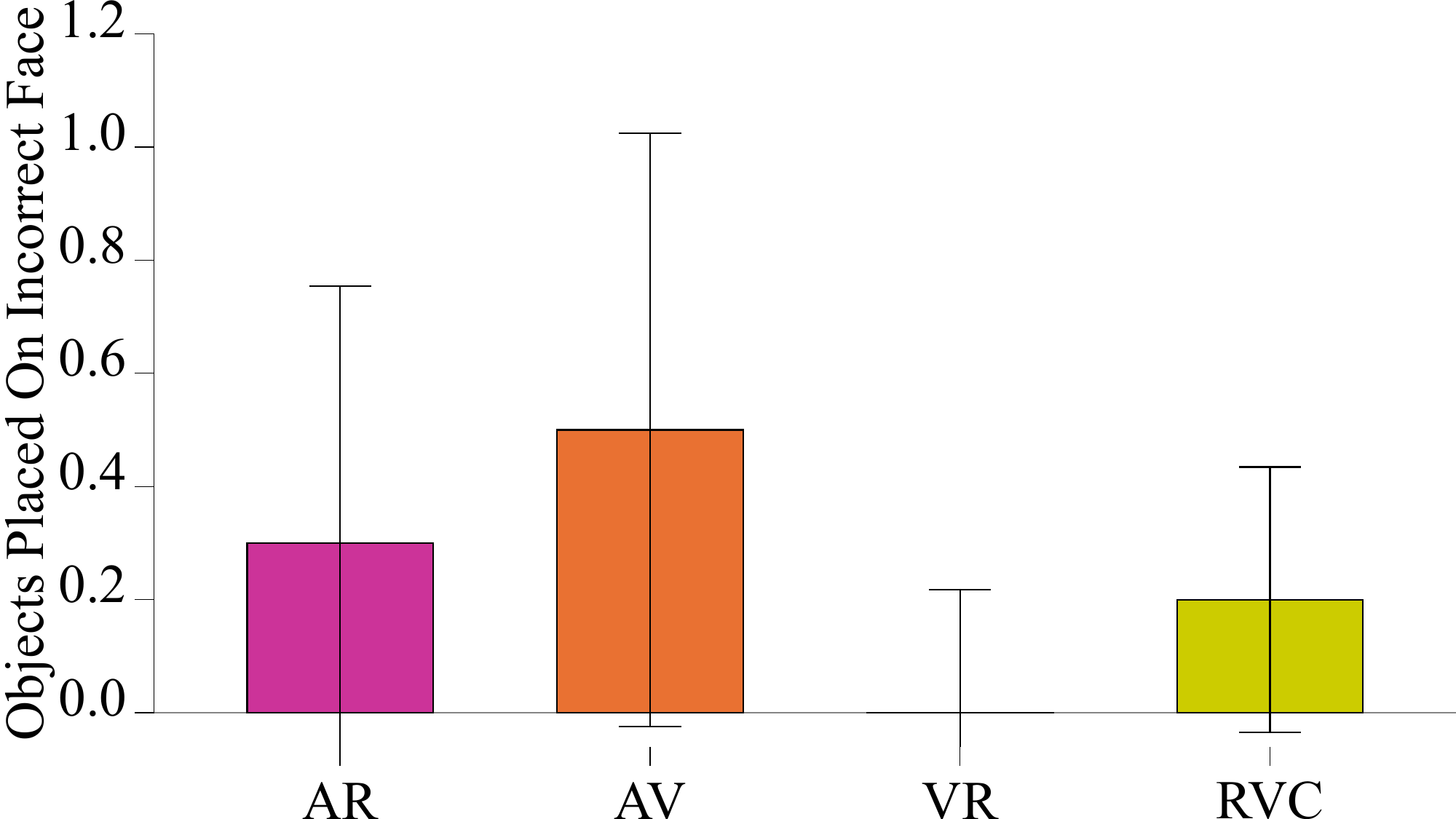}
    \caption{
        Mean Objects Placed on Incorrect Face with 95\% confidence intervals across conditions.
        No statistically significant main effect was found.
    }
    \label{fig:results_fixed_incorrect_face}
\end{figure}

No significant main effect of condition on the number of objects placed on an incorrect face was found, $F(3, 27) = 1.519$, $p = .232$ (Tables~\ref{tab:results_fixed_objects_placed_on_incorrect_face_within_subjects} and \ref{tab:results_fixed_objects_placed_on_incorrect_face_descriptives}, Fig.~\ref{fig:results_fixed_incorrect_face}).

\begin{table}[H]
    \centering
    \footnotesize
    \caption{Within-Subjects Analysis of Variance for Objects Placed on Incorrect Face Across Conditions}
    \label{tab:results_fixed_objects_placed_on_incorrect_face_within_subjects}
    {
        \begin{tabularx}{1.0\linewidth}{lrRrRR}
            \toprule
            Cases & Sum of Squares & df & Mean Square & F & p \\
            \cmidrule[0.4pt]{1-6}
            Objects Placed on Incorrect Face & $1.300$ & $3$ & $0.433$ & $1.519$ & $0.232$ \\
            Residuals & $7.700$ & $27$ & $0.285$ & $ $ & $ $ \\
            \bottomrule
        \end{tabularx}
    }
\end{table}

\begin{table}[H]
    \centering
    \footnotesize
    \caption{Descriptive Statistics for Objects Placed on Incorrect Face Across Conditions}
    \label{tab:results_fixed_objects_placed_on_incorrect_face_descriptives}
    {
        \begin{tabularx}{1.0\linewidth}{lRRRRr}
            \toprule
            Objects Placed on Incorrect Face & N & Mean & SD & SE & Coefficient of variation \\
            \cmidrule[0.4pt]{1-6}
            AR & $10$ & $0.300$ & $0.675$ & $0.213$ & $2.250$ \\
            AV & $10$ & $0.500$ & $0.707$ & $0.224$ & $1.414$ \\
            VR & $10$ & $0.000$ & $0.000$ & $0.000$ & NaN \\
            RVC & $10$ & $0.200$ & $0.422$ & $0.133$ & $2.108$ \\
            \bottomrule
        \end{tabularx}
    }
\end{table}


\subsubsection{Objects Dropped}

\begin{figure}[H]
    \centering
    \includegraphics[width=0.50\linewidth]{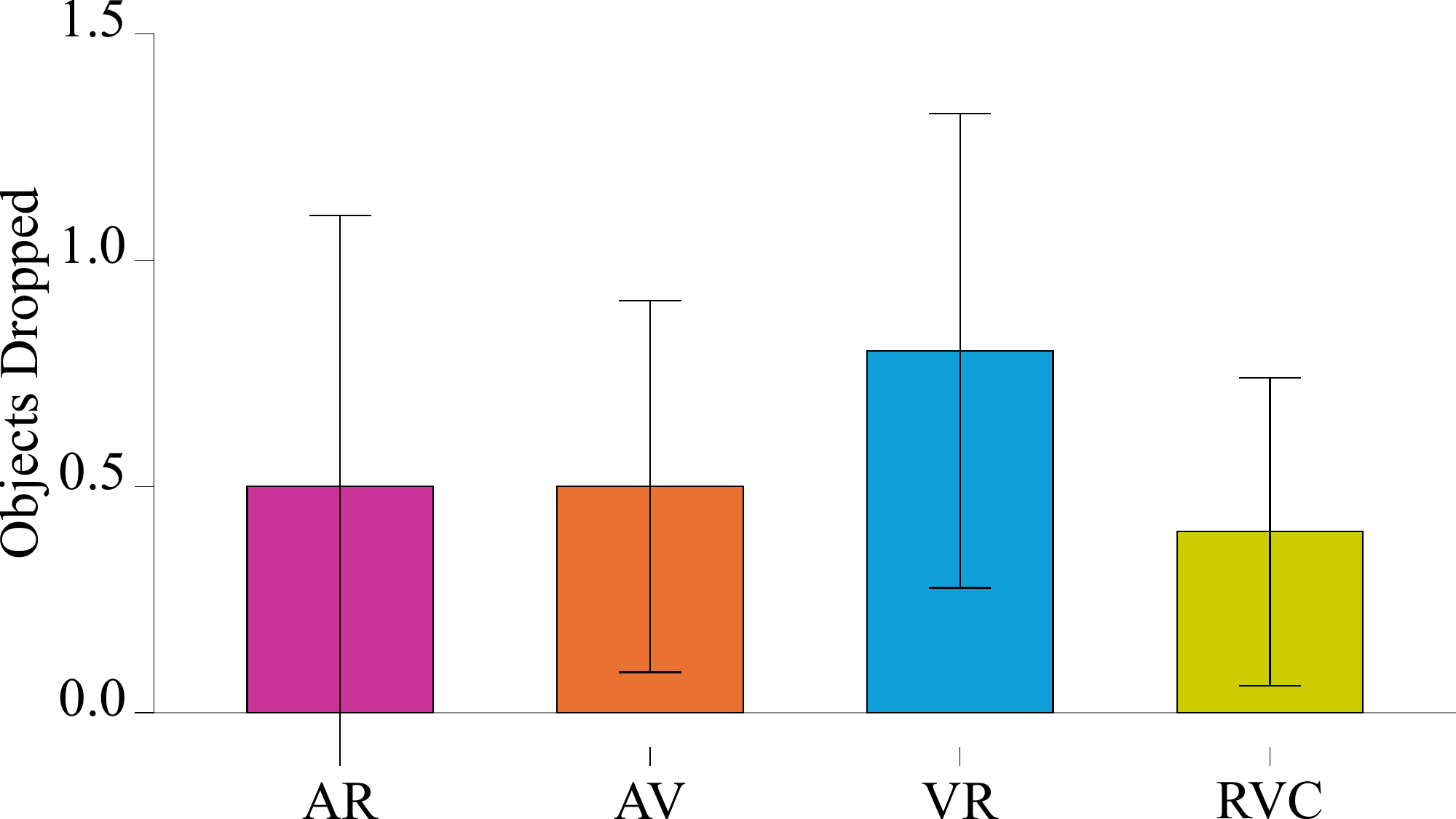}
    \caption{
        Mean Objects Dropped with 95\% confidence intervals across conditions.
        No statistically significant main effect was found.
    }
    \label{fig:results_fixed_objects_dropped}
\end{figure}

With Greenhouse-Geisser correction, no significant main effect of condition on the number of objects dropped was found, $F(2.022, 18.198) = 0.669$, $p = .526$ (Tables~\ref{tab:results_fixed_objects_dropped_within_subjects} and \ref{tab:results_fixed_objects_dropped_descriptives}, Fig.~\ref{fig:results_fixed_objects_dropped}).

\begin{table}[H]
    \centering
    \footnotesize
    \caption{Within-Subjects Analysis of Variance for Objects Dropped Across Conditions}
    \label{tab:results_fixed_objects_dropped_within_subjects}
    {
        \begin{tabularx}{1.0\linewidth}{Lrrrrrr}
            \toprule
            Cases & Sphericity Correction & Sum of Squares & df & Mean Square & F & p \\
            \cmidrule[0.4pt]{1-7}
            Objects Dropped & Greenhouse-Geisser & $0.900$ & $2.022$ & $0.445$ & $0.669$ & $0.526$ \\
            Residuals & Greenhouse-Geisser & $12.100$ & $18.198$ & $0.665$ & $ $ & $ $ \\
            \bottomrule
        \end{tabularx}
    }
\end{table}

\begin{table}[H]
    \centering
    \footnotesize
    \caption{Descriptive Statistics for Objects Dropped Across Conditions}
    \label{tab:results_fixed_objects_dropped_descriptives}
    {
        \begin{tabularx}{1.0\linewidth}{lRRRRr}
            \toprule
            Objects Dropped & N & Mean & SD & SE & Coefficient of variation \\
            \cmidrule[0.4pt]{1-6}
            AR & $10$ & $0.500$ & $0.707$ & $0.224$ & $1.414$ \\
            AV & $10$ & $0.500$ & $0.707$ & $0.224$ & $1.414$ \\
            VR & $10$ & $0.800$ & $1.135$ & $0.359$ & $1.419$ \\
            RVC & $10$ & $0.400$ & $0.516$ & $0.163$ & $1.291$ \\
            \bottomrule
        \end{tabularx}
    }
\end{table}


\subsubsection{Collisions}

\begin{figure}[H]
    \centering
    \includegraphics[width=0.50\linewidth]{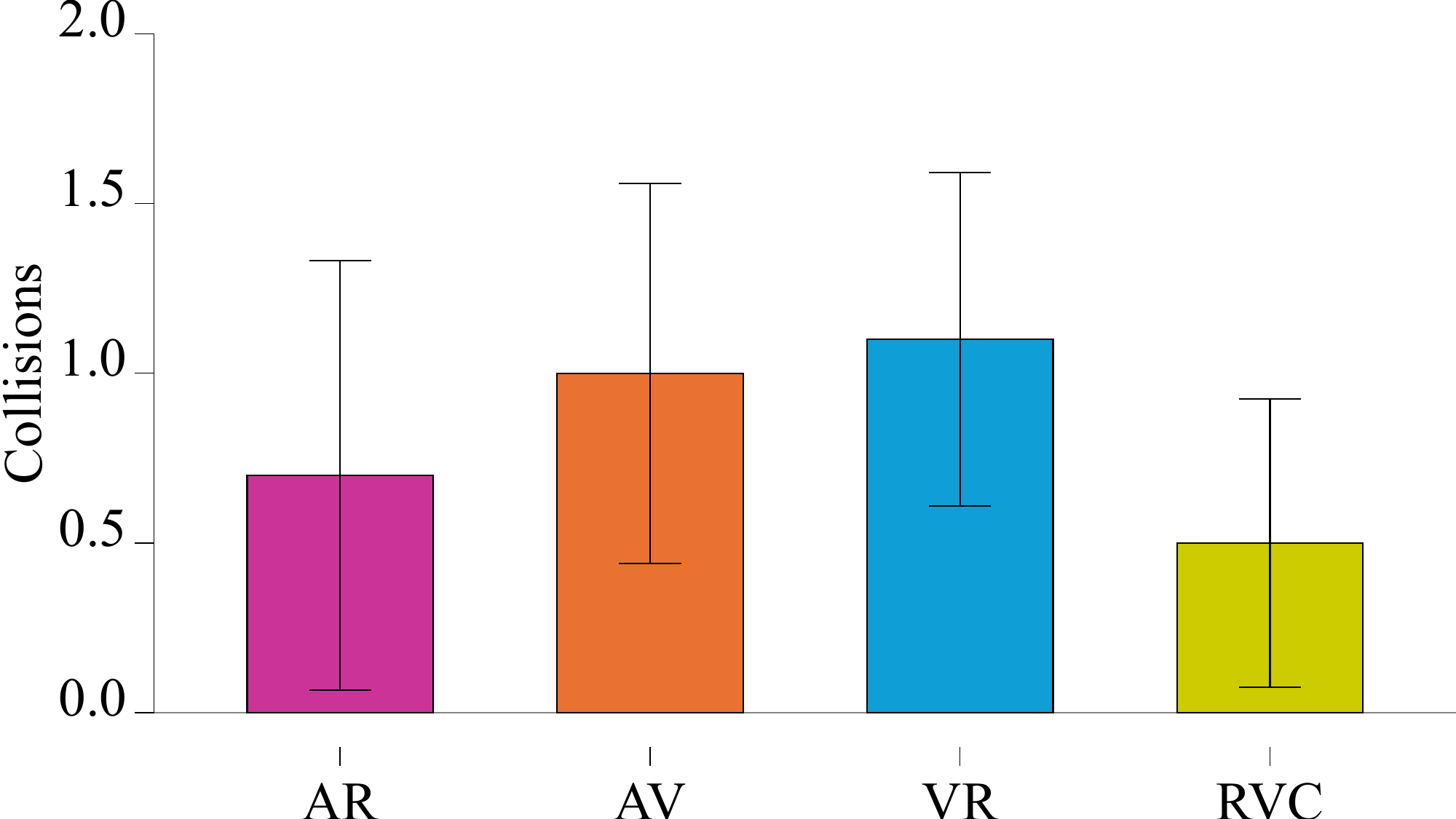}
    \caption{
        Mean Collisions with 95\% confidence intervals across conditions.
        No statistically significant main effect was found.
    }
    \label{fig:results_fixed_collisions}
\end{figure}

No statistically significant differences in collision frequency were found between conditions, $F(3, 27) = 1.367$, $p = .274$ (Tables~\ref{tab:results_fixed_collisions_within_subjects} and \ref{tab:results_fixed_collisions_descriptives}, Fig.~\ref{fig:results_fixed_collisions}).

\begin{table}[H]
    \centering
    \footnotesize
    \caption{Within-Subjects Analysis of Variance for Collisions Across Conditions}
    \label{tab:results_fixed_collisions_within_subjects}
    {
        \begin{tabularx}{1.0\linewidth}{lRRRRR}
            \toprule
            Cases & Sum of Squares & df & Mean Square & F & p \\
            \cmidrule[0.4pt]{1-6}
            Collisions & $2.275$ & $3$ & $0.758$ & $1.367$ & $0.274$ \\
            Residuals & $14.975$ & $27$ & $0.555$ & $ $ & $ $ \\
            \bottomrule
        \end{tabularx}
    }
\end{table}

\begin{table}[H]
    \centering
    \footnotesize
    \caption{Descriptive Statistics for Collisions Across Conditions}
    \label{tab:results_fixed_collisions_descriptives}
    {
        \begin{tabularx}{1.0\linewidth}{lRRRRr}
            \toprule
            Collisions & N & Mean & SD & SE & Coefficient of variation \\
            \cmidrule[0.4pt]{1-6}
            AR & $10$ & $0.700$ & $0.483$ & $0.153$ & $0.690$ \\
            AV & $10$ & $1.000$ & $1.247$ & $0.394$ & $1.247$ \\
            VR & $10$ & $1.100$ & $1.101$ & $0.348$ & $1.000$ \\
            RVC & $10$ & $0.500$ & $0.707$ & $0.224$ & $1.414$ \\
            \bottomrule
        \end{tabularx}
    }
\end{table}


\subsubsection{NASA-TLX}

\begin{figure}[H]
    \centering
    \includegraphics[width=1.0\linewidth]{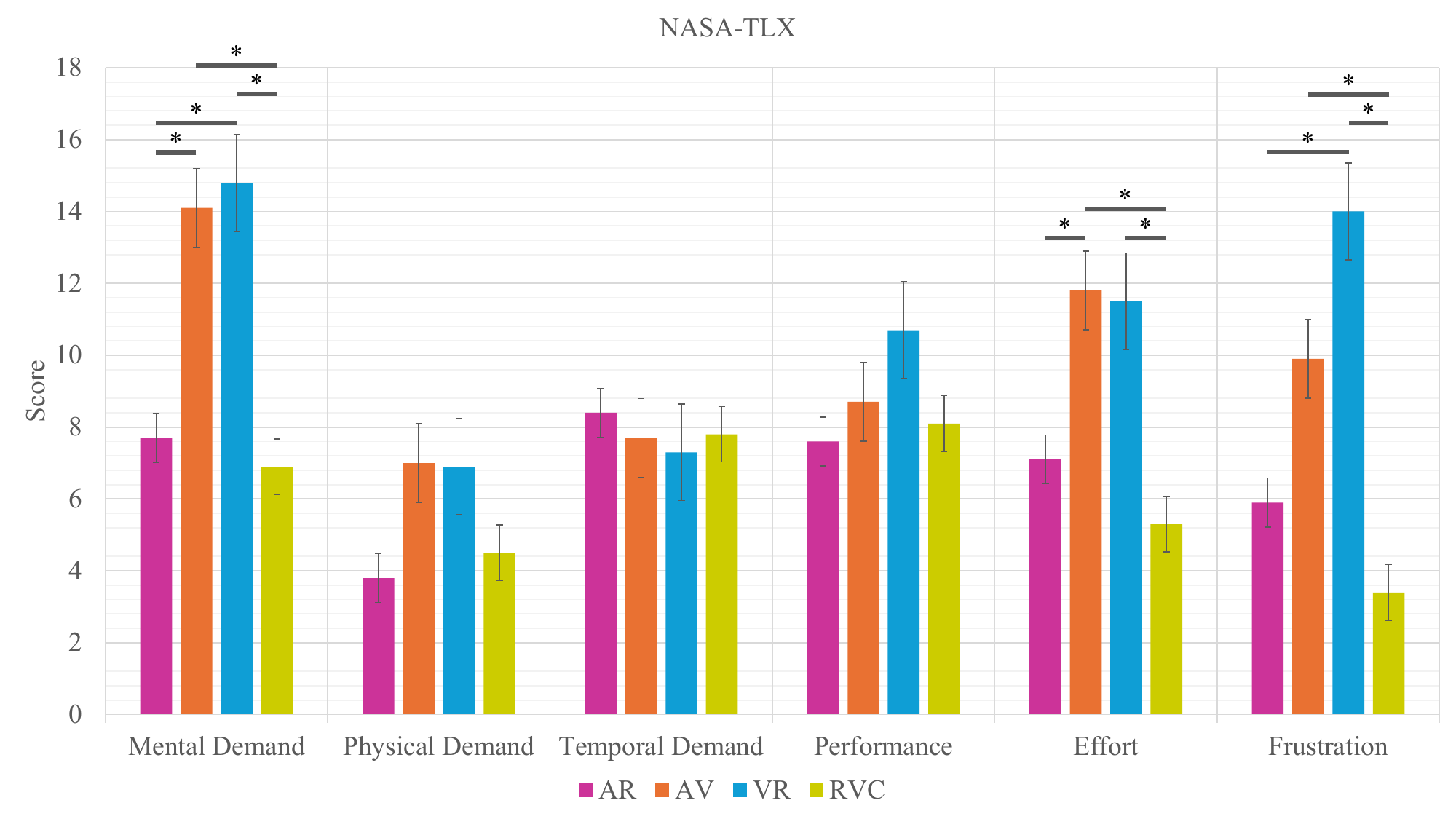}
    \caption{
        NASA-TLX with 95\% confidence intervals across conditions.
        A statistically significant main effect was found, and pairwise differences are denoted by $*$.
    }
    \label{fig:results_fixed_nasa_tlx}
\end{figure}

A significant main effect of condition on mental demand was found after Greenhouse-Geisser correction, $F(1.579, 14.210) = 18.881$, $p < .001$, with ratings highest in VR ($M = 14.800$) and AV ($M = 14.100$), and substantially lower in AR ($M = 7.700$) and RVC ($M = 6.900$) (Tables~\ref{tab:results_fixed_nasa_tlx_mental_within_subjects} and \ref{tab:results_fixed_nasa_tlx_mental_descriptive}, Fig.~\ref{fig:results_fixed_nasa_tlx}).
Post hoc comparisons (Table~\ref{tab:results_fixed_nasa_tlx_mental_post_hoc}) confirmed significantly lower mental demand in AR compared to AV ($p_{holm} = .022$) and VR ($p_{holm} = .003$), no difference between AR and RVC ($p_{holm} = .305$), and significant differences between RVC and both AV ($p_{holm} = .007$) and VR ($p_{holm} < .001$).

For physical demand, the main effect was also significant, $F(3, 27) = 5.311$, $p = .005$, with AV and VR rated highest, although no pairwise comparison reached significance after Holm correction (Tables~\ref{tab:results_fixed_nasa_tlx_physical_within_subjects}, \ref{tab:results_fixed_nasa_tlx_physical_descriptive}, and \ref{tab:results_fixed_nasa_tlx_physical_post_hoc}).
No significant main effect was found for temporal demand, $F(3, 27) = 0.351$, $p = .788$, nor for perceived performance, $F(3, 27) = 0.892$, $p = .458$ (Tables~\ref{tab:results_fixed_nasa_tlx_temporal_within_subjects}, \ref{tab:results_fixed_nasa_tlx_temporal_descriptive}, \ref{tab:results_fixed_nasa_tlx_performance_within_subjects}, and \ref{tab:results_fixed_nasa_tlx_performance_descriptive}).

A significant effect was found for effort, $F(3, 27) = 11.833$, $p < .001$, with the highest ratings in AV ($M = 11.800$) and VR ($M = 11.500$) and the lowest in RVC ($M = 5.300$) (Tables~\ref{tab:results_fixed_nasa_tlx_effort_within_subjects} and \ref{tab:results_fixed_nasa_tlx_effort_descriptive}).
Post hoc analysis (Table~\ref{tab:results_fixed_nasa_tlx_effort_post_hoc}) revealed that effort in AR was significantly lower than in AV ($p_{holm} = .011$) but not VR ($p_{holm} = .067$), while both AV and VR were significantly higher than RVC ($p_{holm} = .001$ and $p_{holm} = .004$, respectively).

Frustration also showed a significant main effect after Greenhouse-Geisser correction, $F(1.741, 15.665) = 12.702$, $p < .001$, reported highest in VR ($M = 14.000$), followed by AV ($M = 9.900$), AR ($M = 5.900$), and RVC ($M = 3.400$) (Tables~\ref{tab:results_fixed_nasa_tlx_frustration_within_subjects} and \ref{tab:results_fixed_nasa_tlx_frustration_descriptive}).
Post hoc comparisons (Table~\ref{tab:results_fixed_nasa_tlx_frustration_post_hoc}) revealed significantly higher frustration in VR compared to AR ($p_{holm} = .004$) and RVC ($p_{holm} < .001$), as well as a significant difference between AV and RVC ($p_{holm} = .004$).


\subsubsection{System Usability Scale}

\begin{figure}[H]
    \centering
    \includegraphics[width=0.5\linewidth]{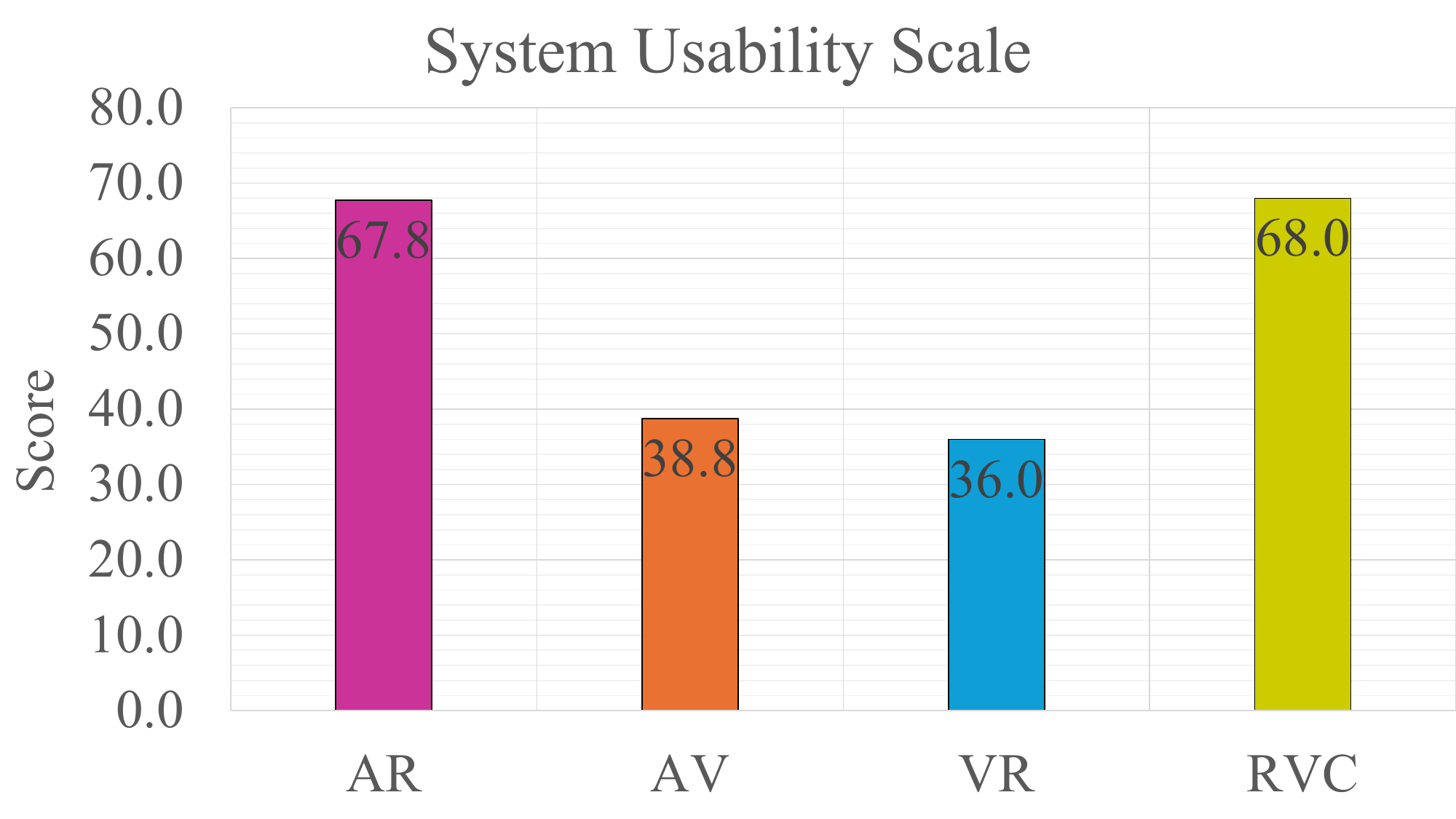}
    \caption{
        System Usability Scale~(SUS) scores across XR conditions.
    }
    \label{fig:results_fixed_sus}
\end{figure}

The SUS scores, summarized in Table~\ref{tab:sus_scores} and Fig.~\ref{fig:results_fixed_sus}, were highest for RVC (68.0), closely followed by AR (67.8), while AV and VR yielded substantially lower ratings of 38.8 and 36.0, respectively.

\begin{table}[H]
    \centering
    \footnotesize
    \caption{System Usability Scale Scores Across Conditions}
    \label{tab:sus_scores}
    {
        \begin{tabularx}{1.0\linewidth}{LRRRR}
            \toprule
            Condition & AR & AV & VR & RVC \\
            \cmidrule[0.4pt]{1-5}
            Score & $67.8$ & $38.8$ & $36.0$ & $68.0$ \\
            \bottomrule
        \end{tabularx}
    }
\end{table}


\subsection{Dynamic Reality Modalities}


\subsubsection{Time in Modality}

\begin{figure}[H]
    \centering
    \includegraphics[width=0.50\linewidth]{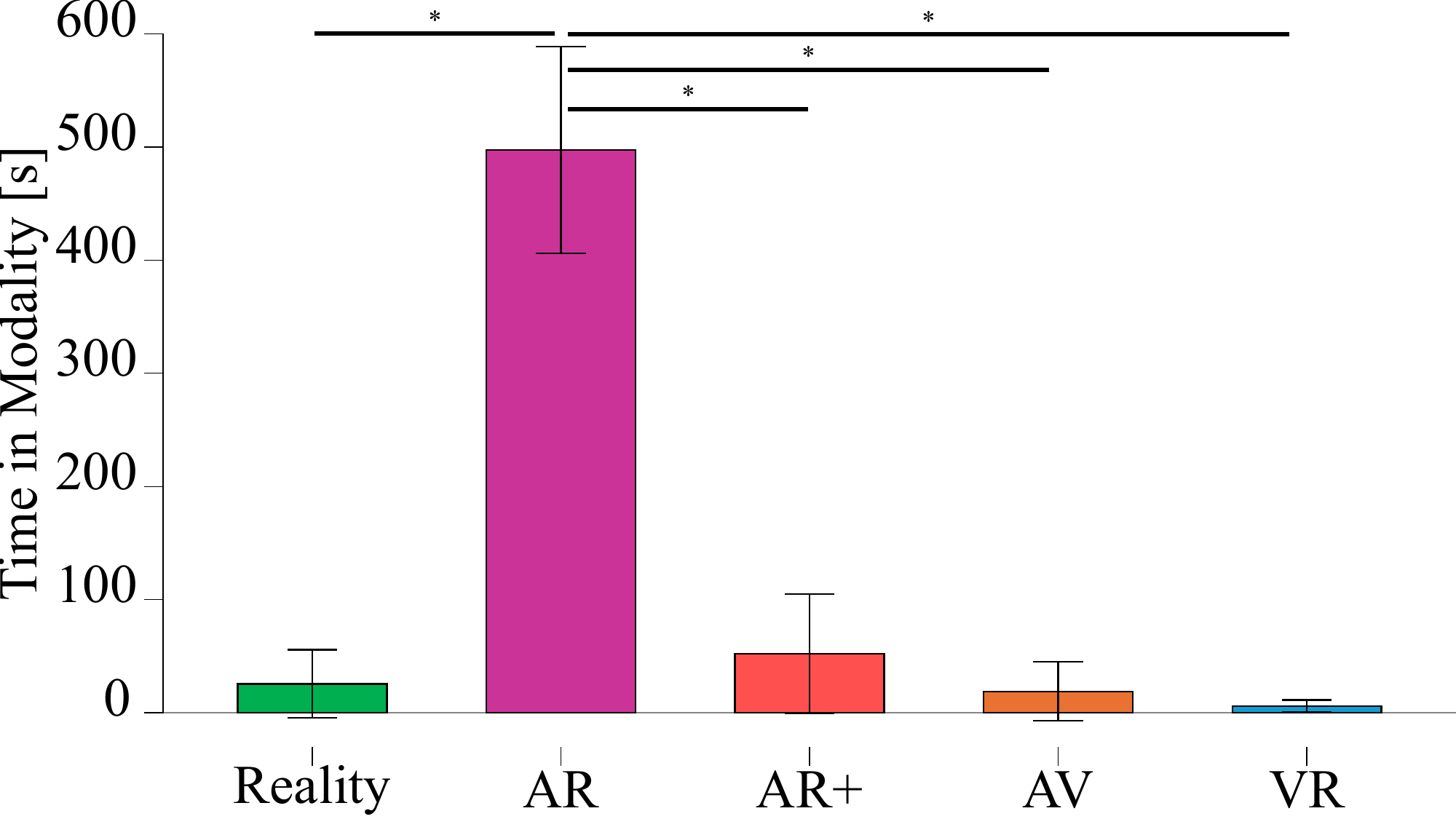}
    \caption{
        Mean Time in Modality with 95\% confidence intervals across modalities.
        A statistically significant main effect was found, and pairwise differences are denoted by $*$.
    }
    \label{fig:results_dynamic_time_in_modality}
\end{figure}

With Greenhouse-Geisser correction applied, a significant main effect of modality on time spent was found during the dynamic RVC condition, $F(1.202, 10.819) = 90.125$, $p < .001$ (Table~\ref{tab:results_dynamic_time_in_reality_within_subjects}, Fig.~\ref{fig:results_dynamic_time_in_modality}).
Participants spent the most time in AR ($M = 497.530$), followed by AR+, Reality, AV, and VR (Table~\ref{tab:results_dynamic_time_in_modality_descriptives}), with time spent in AR differing significantly from all other modalities (all $p_{holm} < .001$) and no significant differences between the remaining modalities (Table~\ref{tab:results_dynamic_time_in_modality_post_hoc}).

\begin{table}[H]
    \centering
    \footnotesize
    \caption{Within-Subjects Analysis of Variance for Time in Modality Across Modalities}
    \label{tab:results_dynamic_time_in_reality_within_subjects}
    {
        \begin{tabularx}{1.0\linewidth}{lrrRrRR}
            \toprule
            Cases & Sphericity Correction & Sum of Squares & df & Mean Square & F & p \\
            \cmidrule[0.4pt]{1-7}
            Time in Modality [s] & Greenhouse-Geisser & $1.793\times10^{+6}$ & $1.202$ & $1.491\times10^{+6}$ & $90.125$ & $<$ .001 \\
            Residuals & Greenhouse-Geisser & $179043.752$ & $10.819$ & $16548.988$ & $ $ & $ $ \\
            \bottomrule
        \end{tabularx}
    }
\end{table}

\begin{table}[H]
    \centering
    \footnotesize
    \caption{Descriptive Statistics for Time in Modality Across Modalities}
    \label{tab:results_dynamic_time_in_modality_descriptives}
    {
        \begin{tabularx}{1.0\linewidth}{lRRRRr}
            \toprule
            Time in Modality [s] & N & Mean & SD & SE & Coefficient of variation \\
            \cmidrule[0.4pt]{1-6}
            Reality & $10$ & $25.473$ & $37.754$ & $11.939$ & $1.482$ \\
            AR & $10$ & $497.530$ & $114.134$ & $36.092$ & $0.229$ \\
            AR+ & $10$ & $52.105$ & $65.756$ & $20.794$ & $1.262$ \\
            AV & $10$ & $18.906$ & $32.765$ & $10.361$ & $1.733$ \\
            VR & $10$ & $5.986$ & $6.668$ & $2.109$ & $1.114$ \\
            \bottomrule
        \end{tabularx}
    }
\end{table}

\begin{table}[H]
    \centering
    \footnotesize
    \caption{Post Hoc Comparisons for Time in Modality Across Modalities}
    \label{tab:results_dynamic_time_in_modality_post_hoc}
    {
        \begin{tabularx}{1.0\linewidth}{llrRRRR}
            \toprule
            $ $ & $ $ & Mean Difference & SE & df & t & p$_{holm}$ \\
            \cmidrule[0.4pt]{1-7}
            Reality & AR & $-472.056$ & $44.357$ & $9$ & $-10.642$ & $<$ .001 \\
            $ $ & AR+ & $-26.632$ & $20.568$ & $9$ & $-1.295$ & $0.683$ \\
            & AV & $6.568$ & $13.414$ & $9$ & $0.490$ & $0.683$ \\
            & VR & $19.487$ & $11.468$ & $9$ & $1.699$ & $0.494$ \\
            AR & AR+ & $445.425$ & $55.931$ & $9$ & $7.964$ & $<$ .001 \\
            $ $ & AV & $478.624$ & $45.113$ & $9$ & $10.610$ & $<$ .001 \\
            & VR & $491.543$ & $37.036$ & $9$ & $13.272$ & $<$ .001 \\
            AR+ & AV & $33.199$ & $14.190$ & $9$ & $2.340$ & $0.264$ \\
            $ $ & VR & $46.119$ & $19.967$ & $9$ & $2.310$ & $0.264$ \\
            AV & VR & $12.919$ & $10.475$ & $9$ & $1.233$ & $0.683$ \\
            \bottomrule
        \end{tabularx}
    }
\end{table}


\subsubsection{Time Navigating}

\begin{figure}[H]
    \centering
    \includegraphics[width=0.50\linewidth]{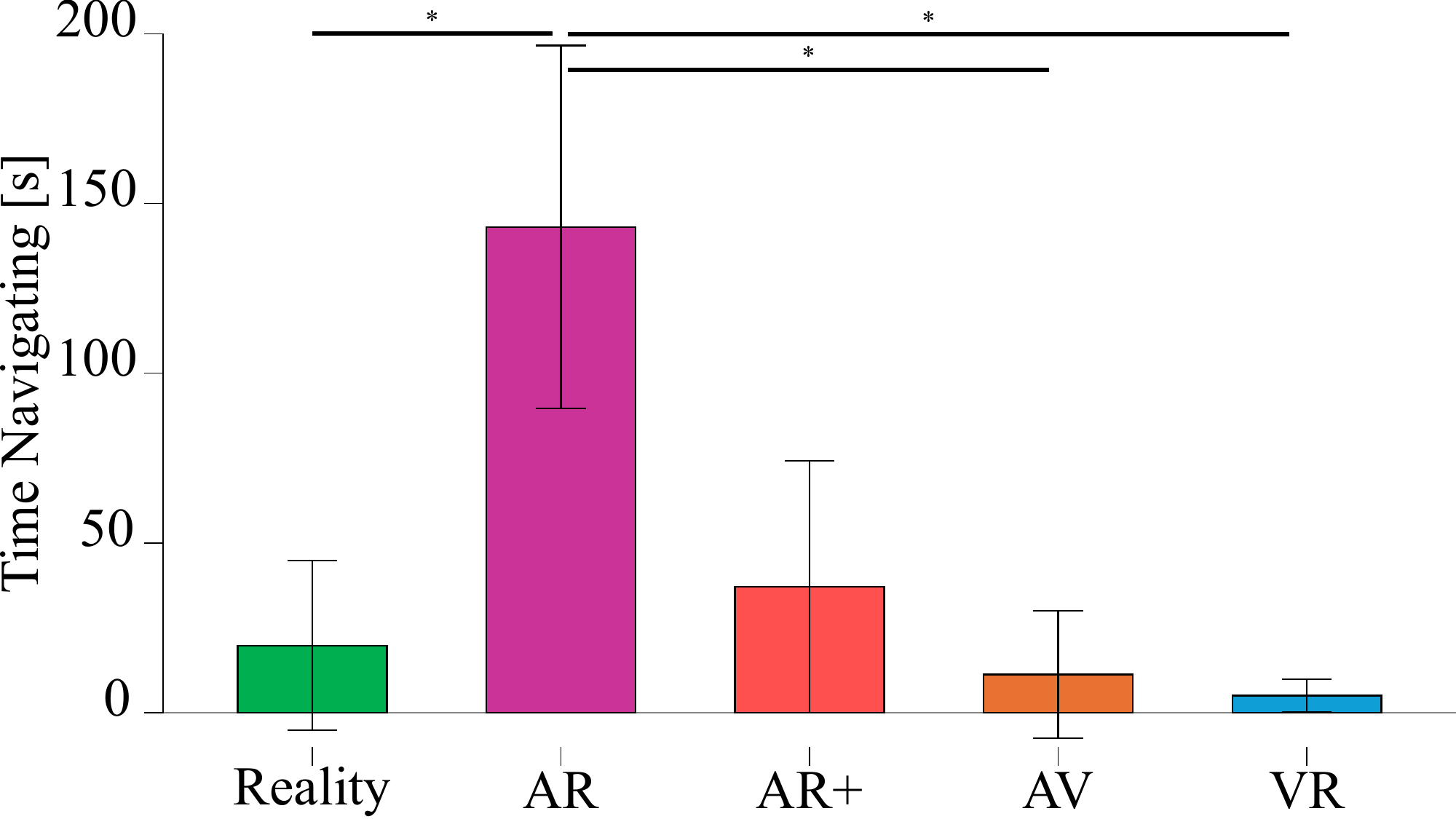}
    \caption{
        Mean Time Navigating with 95\% confidence intervals across modalities.
        A statistically significant main effect was found, and pairwise differences are denoted by $*$.
    }
    \label{fig:results_dynamic_time_navigating}
\end{figure}

A statistically significant main effect of modality on Time Navigating was observed with Greenhouse-Geisser correction, $F(1.493, 13.435) = 15.914$, $p < .001$ (Table~\ref{tab:results_dynamic_time_navigating_within_subjects}, Fig.~\ref{fig:results_dynamic_time_navigating}).
The highest mean was recorded under AR ($M = 143.108$) and the lowest under VR ($M = 5.028$) (Table~\ref{tab:results_dynamic_time_navigating_descriptives}).
Post hoc comparisons (Table~\ref{tab:results_dynamic_time_navigating_post_hoc}) indicated that navigation times under AR were significantly higher than in Reality ($p_{holm} = .012$), AV ($p_{holm} = .006$), and VR ($p_{holm} < .001$), with no significant differences among the remaining modalities.

\begin{table}[H]
    \centering
    \footnotesize
    \caption{Within-Subjects Analysis of Variance for Time Navigating Across Modalities}
    \label{tab:results_dynamic_time_navigating_within_subjects}
    {
        \begin{tabularx}{1.0\linewidth}{lrrRrRR}
            \toprule
            Cases & Sphericity Correction & Sum of Squares & df & Mean Square & F & p \\
            \cmidrule[0.4pt]{1-7}
            Time Navigating & Greenhouse-Geisser & $130435.193$ & $1.493$ & $87377.560$ & $15.914$ & $<$ .001 \\
            Residuals & Greenhouse-Geisser & $73767.332$ & $13.435$ & $5490.687$ & $ $ & $ $ \\
            \bottomrule
        \end{tabularx}
    }
\end{table}

\begin{table}[H]
    \centering
    \footnotesize
    \caption{Descriptive Statistics for Time Navigating Across Modalities}
    \label{tab:results_dynamic_time_navigating_descriptives}
    {
        \begin{tabularx}{1.0\linewidth}{lRRRRr}
            \toprule
            Time Navigating & N & Mean & SD & SE & Coefficient of variation \\
            \cmidrule[0.4pt]{1-6}
            Reality & $10$ & $19.808$ & $33.656$ & $10.643$ & $1.699$ \\
            AR & $10$ & $143.108$ & $65.788$ & $20.804$ & $0.460$ \\
            AR+ & $10$ & $37.150$ & $48.145$ & $15.225$ & $1.296$ \\
            AV & $10$ & $11.245$ & $23.391$ & $7.397$ & $2.080$ \\
            VR & $10$ & $5.028$ & $6.247$ & $1.975$ & $1.242$ \\
            \bottomrule
        \end{tabularx}
    }
\end{table}

\begin{table}[H]
    \centering
    \footnotesize
    \caption{Post Hoc Comparisons for Time Navigating Across Modalities}
    \label{tab:results_dynamic_time_navigating_post_hoc}
    {
        \begin{tabularx}{1.0\linewidth}{llrRRRR}
            \toprule
            $ $ & $ $ & Mean Difference & SE & df & t & p$_{holm}$ \\
            \cmidrule[0.4pt]{1-7}
            Reality & AR & $-123.300$ & $27.279$ & $9$ & $-4.520$ & $0.012$ \\
            $ $ & AR+ & $-17.342$ & $16.253$ & $9$ & $-1.067$ & $0.941$ \\
            & AV & $8.563$ & $13.592$ & $9$ & $0.630$ & $0.941$ \\
            & VR & $14.780$ & $10.780$ & $9$ & $1.371$ & $0.814$ \\
            AR & AR+ & $105.958$ & $34.893$ & $9$ & $3.037$ & $0.099$ \\
            $ $ & AV & $131.863$ & $25.660$ & $9$ & $5.139$ & $0.006$ \\
            & VR & $138.079$ & $20.893$ & $9$ & $6.609$ & $<$ .001 \\
            AR+ & AV & $25.905$ & $13.842$ & $9$ & $1.871$ & $0.470$ \\
            $ $ & VR & $32.122$ & $14.956$ & $9$ & $2.148$ & $0.362$ \\
            AV & VR & $6.217$ & $7.822$ & $9$ & $0.795$ & $0.941$ \\
            \bottomrule
        \end{tabularx}
    }
\end{table}


\subsubsection{Time Manipulating}

\begin{figure}[H]
    \centering
    \includegraphics[width=0.50\linewidth]{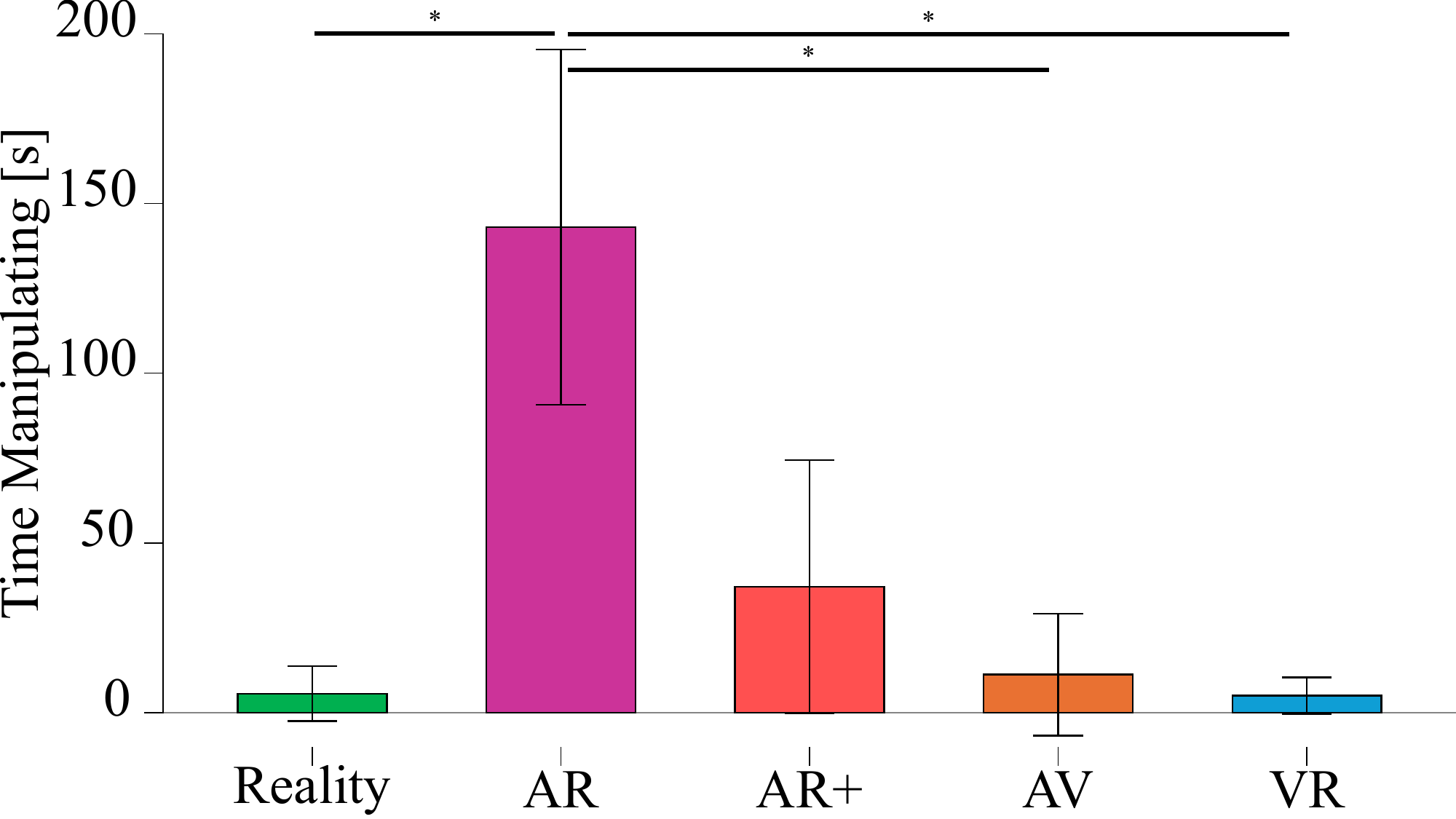}
    \caption{
        Mean Time Manipulating with 95\% confidence intervals across modalities.
        A statistically significant main effect was found, and pairwise differences are denoted by $*$.
    }
    \label{fig:results_dynamic_time_manipulating}
\end{figure}

A significant main effect of modality on Time Manipulating was observed with Greenhouse-Geisser correction, $F(1.192, 10.730) = 19.529$, $p < .001$ (Table~\ref{tab:results_dynamic_time_manipulating_within_subjects}, Fig.~\ref{fig:results_dynamic_time_manipulating}).
AR resulted in the highest average manipulation time ($M = 143.108$) (Table~\ref{tab:results_dynamic_time_manipulating_descriptives}), significantly higher than in Reality ($p_{holm} = .002$), AV ($p_{holm} = .005$), and VR ($p_{holm} < .001$), with no other significant differences after correction (Table~\ref{tab:results_dynamic_time_manipulating_post_hoc}).

\begin{table}[H]
    \centering
    \footnotesize
    \caption{Within-Subjects Analysis of Variance for Time Manipulating Across Modalities}
    \label{tab:results_dynamic_time_manipulating_within_subjects}
    {
        \begin{tabularx}{1.0\linewidth}{Lrrrrrr}
            \toprule
            Cases & Sphericity Correction & Sum of Squares & df & Mean Square & F & p \\
            \cmidrule[0.4pt]{1-7}
            Time Manipulating & Greenhouse-Geisser & $138674.410$ & $1.192$ & $116316.682$ & $19.529$ & $<$ .001 \\
            Residuals & Greenhouse-Geisser & $63908.829$ & $10.730$ & $5956.128$ & $ $ & $ $ \\
            \bottomrule
        \end{tabularx}
    }
\end{table}

\begin{table}[H]
    \centering
    \footnotesize
    \caption{Descriptive Statistics for Time Manipulating Across Modalities}
    \label{tab:results_dynamic_time_manipulating_descriptives}
    {
        \begin{tabularx}{1.0\linewidth}{lRRRRr}
            \toprule
            Time Manipulating & N & Mean & SD & SE & Coefficient of variation \\
            \cmidrule[0.4pt]{1-6}
            Reality & $10$ & $5.661$ & $9.909$ & $3.133$ & $1.750$ \\
            AR & $10$ & $143.108$ & $65.788$ & $20.804$ & $0.460$ \\
            AR+ & $10$ & $37.150$ & $48.145$ & $15.225$ & $1.296$ \\
            AV & $10$ & $11.245$ & $23.391$ & $7.397$ & $2.080$ \\
            VR & $10$ & $5.028$ & $6.247$ & $1.975$ & $1.242$ \\
            \bottomrule
        \end{tabularx}
    }
\end{table}

\begin{table}[H]
    \centering
    \footnotesize
    \caption{Post Hoc Comparisons for Time Manipulating Across Modalities}
    \label{tab:results_dynamic_time_manipulating_post_hoc}
    {
        \begin{tabularx}{1.0\linewidth}{llrRRRR}
            \toprule
            $ $ & $ $ & Mean Difference & SE & df & t & p$_{holm}$ \\
            \cmidrule[0.4pt]{1-7}
            Reality & AR & $-137.447$ & $23.163$ & $9$ & $-5.934$ & $0.002$ \\
            $ $ & AR+ & $-31.489$ & $12.728$ & $9$ & $-2.474$ & $0.212$ \\
            & AV & $-5.585$ & $7.197$ & $9$ & $-0.776$ & $1.000$ \\
            & VR & $0.632$ & $3.350$ & $9$ & $0.189$ & $1.000$ \\
            AR & AR+ & $105.958$ & $34.893$ & $9$ & $3.037$ & $0.099$ \\
            $ $ & AV & $131.863$ & $25.660$ & $9$ & $5.139$ & $0.005$ \\
            & VR & $138.079$ & $20.893$ & $9$ & $6.609$ & $<$ .001 \\
            AR+ & AV & $25.905$ & $13.842$ & $9$ & $1.871$ & $0.376$ \\
            $ $ & VR & $32.122$ & $14.956$ & $9$ & $2.148$ & $0.301$ \\
            AV & VR & $6.217$ & $7.822$ & $9$ & $0.795$ & $1.000$ \\
            \bottomrule
        \end{tabularx}
    }
\end{table}


\subsubsection{Objects Grasped}

\begin{figure}[H]
    \centering
    \includegraphics[width=0.50\linewidth]{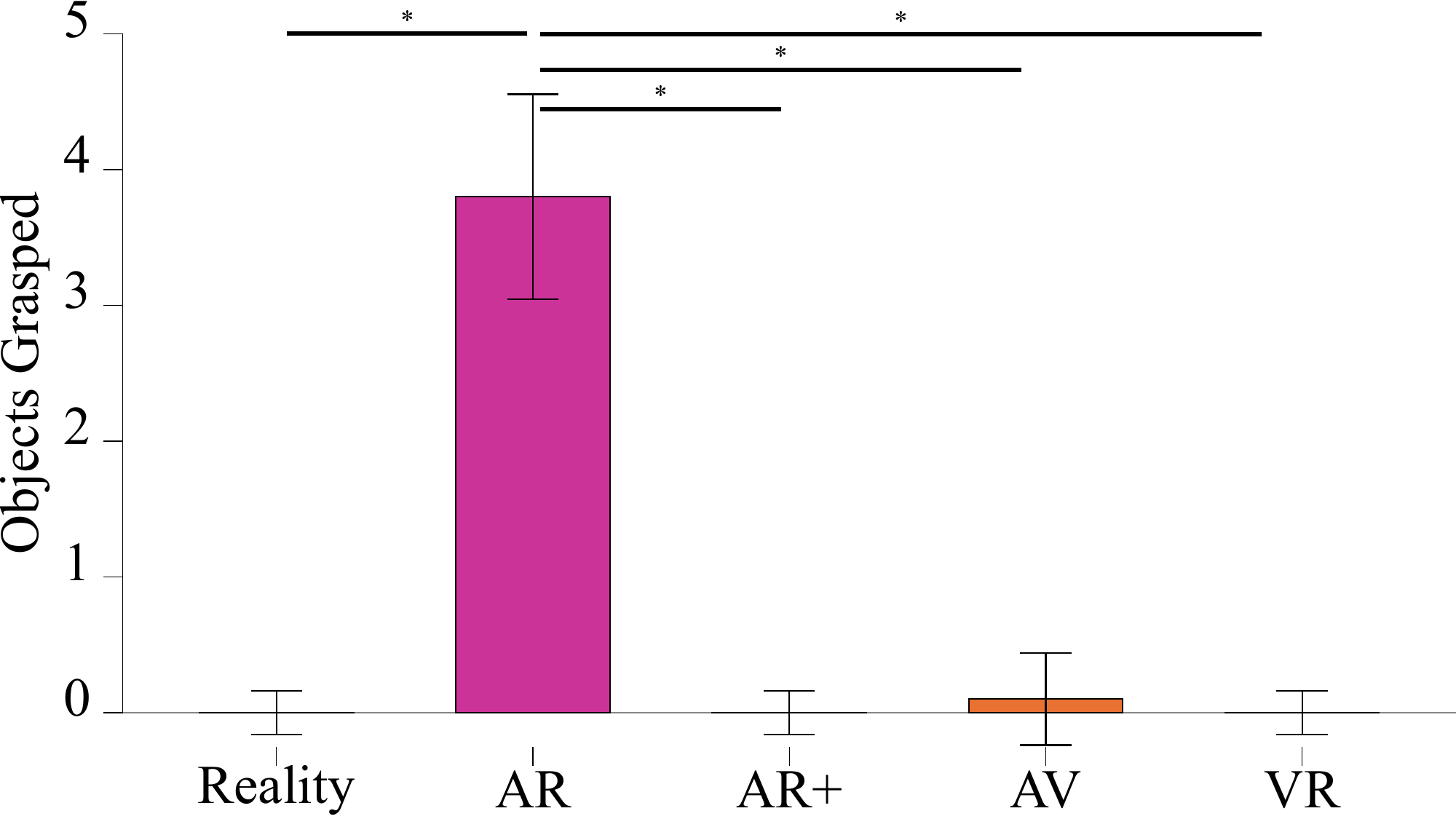}
    \caption{
        Mean Objects Grasped with 95\% confidence intervals across modalities.
        A statistically significant main effect was found, and pairwise differences are denoted by $*$.
    }
    \label{fig:results_dynamic_objects_grasped}
\end{figure}

A statistically significant main effect of modality on the number of objects grasped was found, $F(4, 36) = 95.776$, $p < .001$ (Table~\ref{tab:results_dynamic_objects_grasped_within_subjects}, Fig.~\ref{fig:results_dynamic_objects_grasped}).
Participants grasped a mean of $3.800$ objects in AR, compared to $0.100$ in AV and none in Reality, AR+, and VR (Table~\ref{tab:results_dynamic_object_grasped_descriptives}).
Holm-corrected post hoc comparisons confirmed that grasps in AR were significantly higher than in all other modalities (all $p_{holm} < .001$), with no other significant differences (Table~\ref{tab:results_dynamic_object_grasped_post_hoc}).

\begin{table}[H]
    \centering
    \footnotesize
    \caption{Within-Subjects Analysis of Variance for Objects Grasped Across Modalities}
    \label{tab:results_dynamic_objects_grasped_within_subjects}
    {
        \begin{tabularx}{1.0\linewidth}{lRRRRr}
            \toprule
            Cases & Sum of Squares & df & Mean Square & F & p \\
            \cmidrule[0.4pt]{1-6}
            Objects Grasped & $114.080$ & $4$ & $28.520$ & $95.776$ & $<$ .001 \\
            Residuals & $10.720$ & $36$ & $0.298$ & $ $ & $ $ \\
            \bottomrule
        \end{tabularx}
    }
\end{table}

\begin{table}[H]
    \centering
    \footnotesize
    \caption{Descriptive Statistics for Objects Grasped Across Modalities}
    \label{tab:results_dynamic_object_grasped_descriptives}
    {
        \begin{tabularx}{1.0\linewidth}{lRRRRr}
            \toprule
            Objects Grasped & N & Mean & SD & SE & Coefficient of variation \\
            \cmidrule[0.4pt]{1-6}
            Reality & $10$ & $0.000$ & $0.000$ & $0.000$ & NaN \\
            AR & $10$ & $3.800$ & $1.135$ & $0.359$ & $0.299$ \\
            AR+ & $10$ & $0.000$ & $0.000$ & $0.000$ & NaN \\
            AV & $10$ & $0.100$ & $0.316$ & $0.100$ & $3.162$ \\
            VR & $10$ & $0.000$ & $0.000$ & $0.000$ & NaN \\
            \bottomrule
        \end{tabularx}
    }
\end{table}

\begin{table}[H]
    \centering
    \footnotesize
    \caption{Post Hoc Comparisons for Objects Grasped Across Modalities}
    \label{tab:results_dynamic_object_grasped_post_hoc}
    {
        \begin{tabularx}{1.0\linewidth}{llrRRRR}
            \toprule
            $ $ & $ $ & Mean Difference & SE & df & t & p$_{holm}$ \\
            \cmidrule[0.4pt]{1-7}
            Reality & AR & $-3.800$ & $0.359$ & $9$ & $-10.585$ & $<$ .001 \\
            $ $ & AR+ & $0.000$ & $0.000$ & $9$ & NaN & NaN \\
            & AV & $-0.100$ & $0.100$ & $9$ & $-1.000$ & $1.000$ \\
            & VR & $0.000$ & $0.000$ & $9$ & NaN & NaN \\
            AR & AR+ & $3.800$ & $0.359$ & $9$ & $10.585$ & $<$ .001 \\
            $ $ & AV & $3.700$ & $0.423$ & $9$ & $8.748$ & $<$ .001 \\
            & VR & $3.800$ & $0.359$ & $9$ & $10.585$ & $<$ .001 \\
            AR+ & AV & $-0.100$ & $0.100$ & $9$ & $-1.000$ & $1.000$ \\
            $ $ & VR & $0.000$ & $0.000$ & $9$ & NaN & NaN \\
            AV & VR & $0.100$ & $0.100$ & $9$ & $1.000$ & $1.000$ \\
            \bottomrule
        \end{tabularx}
    }
\end{table}


\subsubsection{Objects Placed}

\begin{figure}[H]
    \centering
    \includegraphics[width=0.50\linewidth]{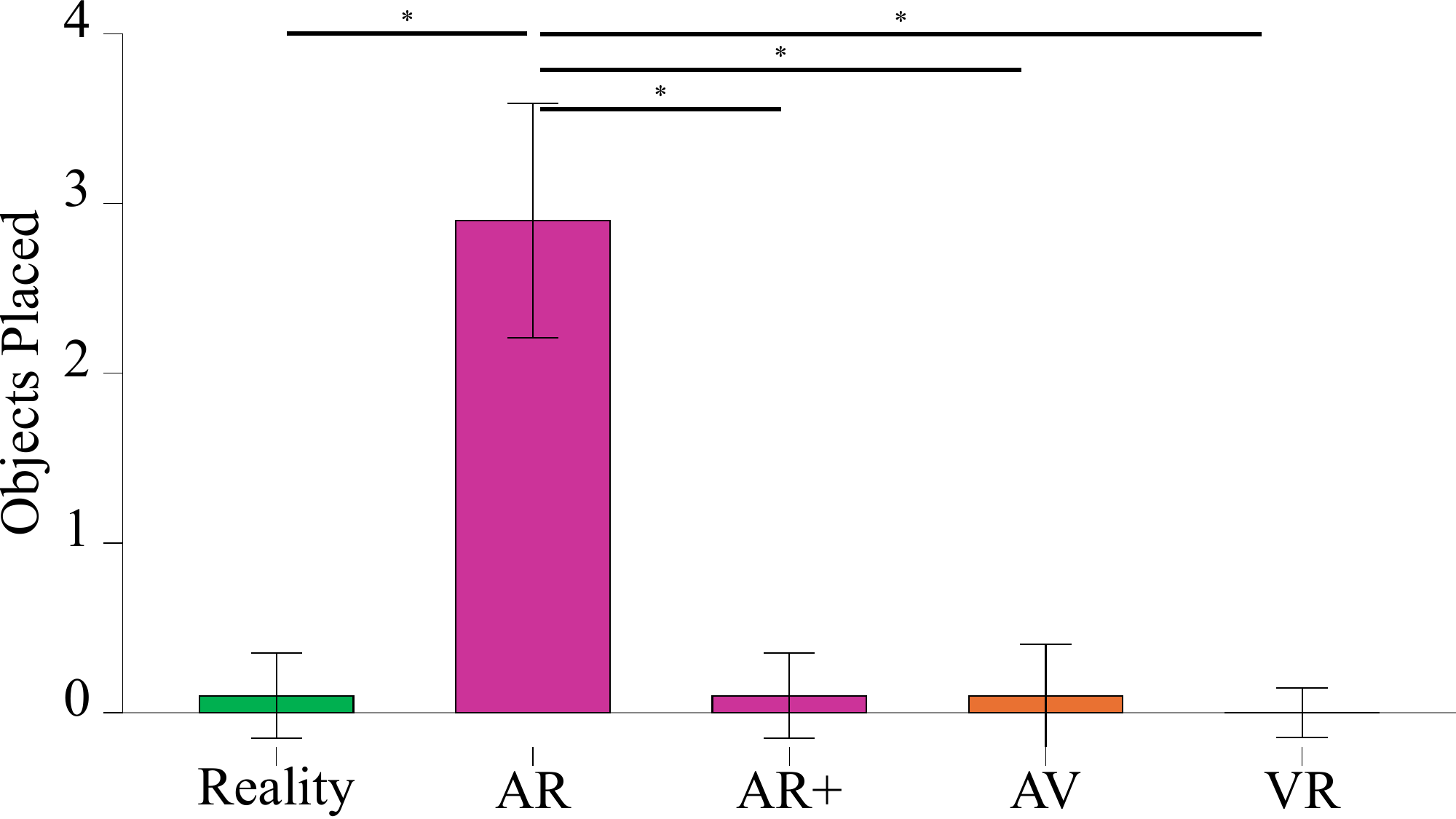}
    \caption{
        Mean Objects Placed with 95\% confidence intervals across modalities.
        A statistically significant main effect was found, and pairwise differences are denoted by $*$.
    }
    \label{fig:results_dynamic_objects_placed}
\end{figure}

A statistically significant main effect of modality on the number of objects placed was found, $F(4, 36) = 57.071$, $p < .001$ (Table~\ref{tab:results_dynamic_objects_placed_within_subjects}, Fig.~\ref{fig:results_dynamic_objects_placed}).
Participants placed the most objects in AR ($M = 2.900$), whereas Reality, AR+, AV, and VR showed markedly lower means between $M = 0.000$ and $M = 0.100$ (Table~\ref{tab:results_dynamic_objects_placed_descriptives}).
Holm-corrected post hoc comparisons (Table~\ref{tab:results_dynamic_objects_placed_post_hoc}) showed that placements in AR were significantly higher than in all other conditions (all $p_{holm} < .001$), with no significant differences among the remaining modalities.

\begin{table}[H]
    \centering
    \footnotesize
    \caption{Within-Subjects Analysis of Variance for Objects Placed Across Modalities}
    \label{tab:results_dynamic_objects_placed_within_subjects}
    {
        \begin{tabularx}{1.0\linewidth}{lRRRRR}
            \toprule
            Cases & Sum of Squares & df & Mean Square & F & p \\
            \cmidrule[0.4pt]{1-6}
            Objects Placed & $63.920$ & $4$ & $15.980$ & $57.071$ & $<$ .001 \\
            Residuals & $10.080$ & $36$ & $0.280$ & $ $ & $ $ \\
            \bottomrule
        \end{tabularx}
    }
\end{table}

\begin{table}[H]
    \centering
    \footnotesize
    \caption{Descriptive Statistics for Objects Placed Across Modalities}
    \label{tab:results_dynamic_objects_placed_descriptives}
    {
        \begin{tabularx}{1.0\linewidth}{lRRRRr}
            \toprule
            Objects Placed & N & Mean & SD & SE & Coefficient of variation \\
            \cmidrule[0.4pt]{1-6}
            Reality & $10$ & $0.100$ & $0.316$ & $0.100$ & $3.162$ \\
            AR & $10$ & $2.900$ & $0.994$ & $0.314$ & $0.343$ \\
            AR+ & $10$ & $0.100$ & $0.316$ & $0.100$ & $3.162$ \\
            AV & $10$ & $0.100$ & $0.316$ & $0.100$ & $3.162$ \\
            VR & $10$ & $0.000$ & $0.000$ & $0.000$ & NaN \\
            \bottomrule
        \end{tabularx}
    }
\end{table}

\begin{table}[H]
    \centering
    \footnotesize
    \caption{Post Hoc Comparisons for Objects Placed Across Modalities}
    \label{tab:results_dynamic_objects_placed_post_hoc}
    {
        \begin{tabularx}{1.0\linewidth}{llrRRRR}
            \toprule
            $ $ & $ $ & Mean Difference & SE & df & t & p$_{holm}$ \\
            \cmidrule[0.4pt]{1-7}
            Reality & AR & $-2.800$ & $0.359$ & $9$ & $-7.799$ & $<$ .001 \\
            $ $ & AR+ & $0.000$ & $0.000$ & $9$ & NaN & NaN \\
            & AV & $-2.776\times10^{-17}$ & $0.149$ & $9$ & $-1.862\times10^{-16}$ & $1.000$ \\
            & VR & $0.100$ & $0.100$ & $9$ & $1.000$ & $1.000$ \\
            AR & AR+ & $2.800$ & $0.359$ & $9$ & $7.799$ & $<$ .001 \\
            $ $ & AV & $2.800$ & $0.359$ & $9$ & $7.799$ & $<$ .001 \\
            & VR & $2.900$ & $0.314$ & $9$ & $9.222$ & $<$ .001 \\
            AR+ & AV & $-2.776\times10^{-17}$ & $0.149$ & $9$ & $-1.862\times10^{-16}$ & $1.000$ \\
            $ $ & VR & $0.100$ & $0.100$ & $9$ & $1.000$ & $1.000$ \\
            AV & VR & $0.100$ & $0.100$ & $9$ & $1.000$ & $1.000$ \\
            \bottomrule
        \end{tabularx}
    }
\end{table}


\subsubsection{Objects Placed on Incorrect Face}

\begin{figure}[H]
    \centering
    \includegraphics[width=0.50\linewidth]{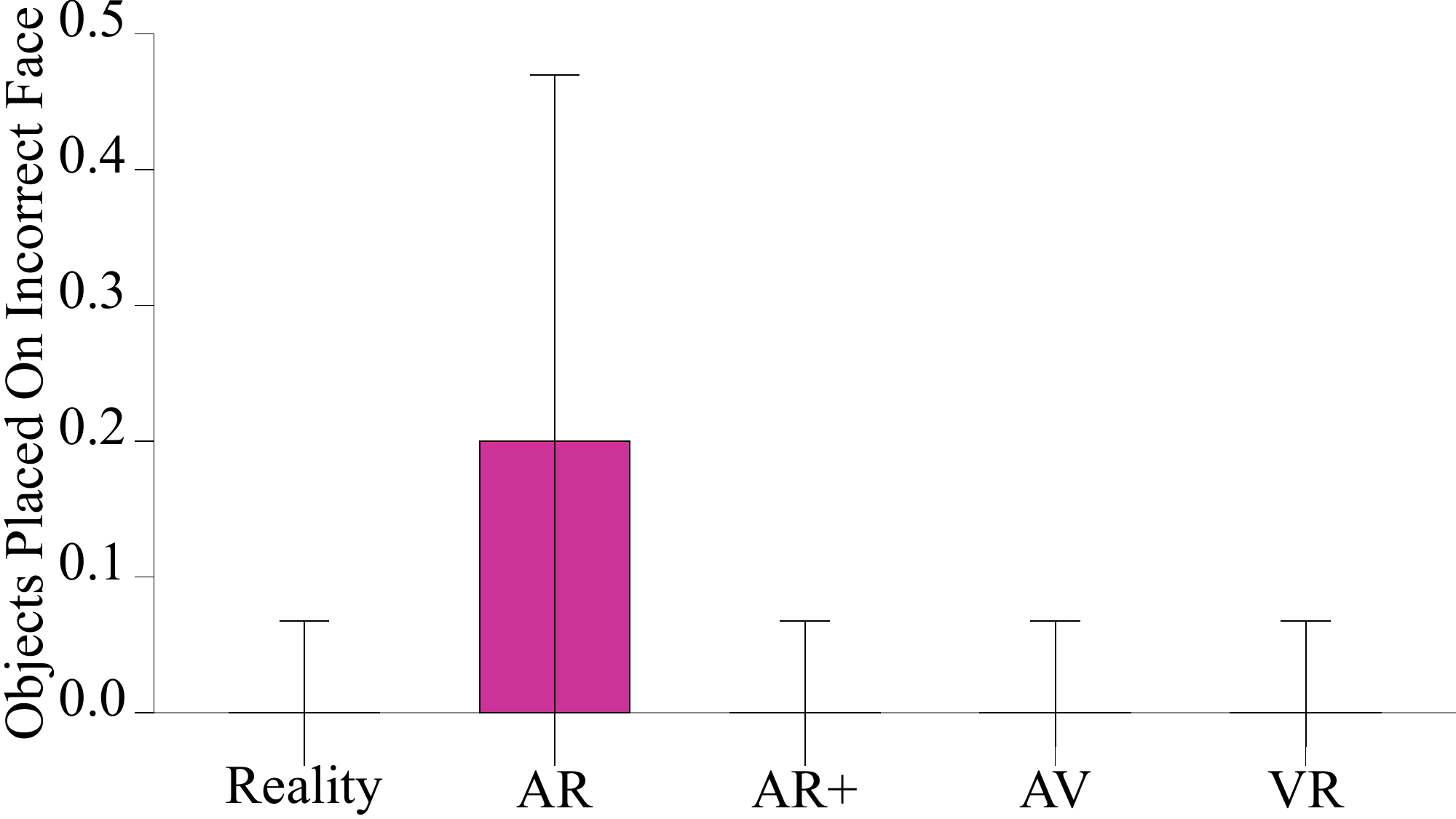}
    \caption{
        Mean Objects Placed on Incorrect Face with 95\% confidence intervals across modalities.
        A statistically significant main effect was found, and pairwise differences are denoted by $*$.
    }
    \label{fig:results_dynamic_incorrect_face}
\end{figure}

A marginally non-significant main effect of modality on the number of incorrectly placed objects was found, $F(4, 36) = 2.250$, $p = .083$ (Table~\ref{tab:results_dynamic_objects_placed_on_incorrect_face_within_subjects}, Fig.~\ref{fig:results_dynamic_incorrect_face}).
The mean was highest in AR $(M = 0.200)$, with no incorrect placements recorded in Reality, AR+, AV, or VR (Table~\ref{tab:results_dynamic_objects_placed_on_incorrect_face_descriptives}).

\begin{table}[H]
    \centering
    \footnotesize
    \caption{Within-Subjects Analysis of Variance for Objects Placed on Incorrect Face Across Modalities}
    \label{tab:results_dynamic_objects_placed_on_incorrect_face_within_subjects}
    {
        \begin{tabularx}{1.0\linewidth}{lrRRRR}
            \toprule
            Cases & Sum of Squares & df & Mean Square & F & p \\
            \cmidrule[0.4pt]{1-6}
            Objects Placed on Incorrect Face & $0.320$ & $4$ & $0.080$ & $2.250$ & $0.083$ \\
            Residuals & $1.280$ & $36$ & $0.036$ & $ $ & $ $ \\
            \bottomrule
        \end{tabularx}
    }
\end{table}

\begin{table}[H]
    \centering
    \footnotesize
    \caption{Descriptive Statistics for Objects Placed on Incorrect Face Across Modalities}
    \label{tab:results_dynamic_objects_placed_on_incorrect_face_descriptives}
    {
        \begin{tabularx}{1.0\linewidth}{lRRRRr}
            \toprule
            Objects Placed on Incorrect Face & N & Mean & SD & SE & Coefficient of variation \\
            \cmidrule[0.4pt]{1-6}
            Reality & $10$ & $0.000$ & $0.000$ & $0.000$ & NaN \\
            AR & $10$ & $0.200$ & $0.422$ & $0.133$ & $2.108$ \\
            AR+ & $10$ & $0.000$ & $0.000$ & $0.000$ & NaN \\
            AV & $10$ & $0.000$ & $0.000$ & $0.000$ & NaN \\
            VR & $10$ & $0.000$ & $0.000$ & $0.000$ & NaN \\
            \bottomrule
        \end{tabularx}
    }
\end{table}


\subsubsection{Objects Dropped}

\begin{figure}[H]
    \centering
    \includegraphics[width=0.50\linewidth]{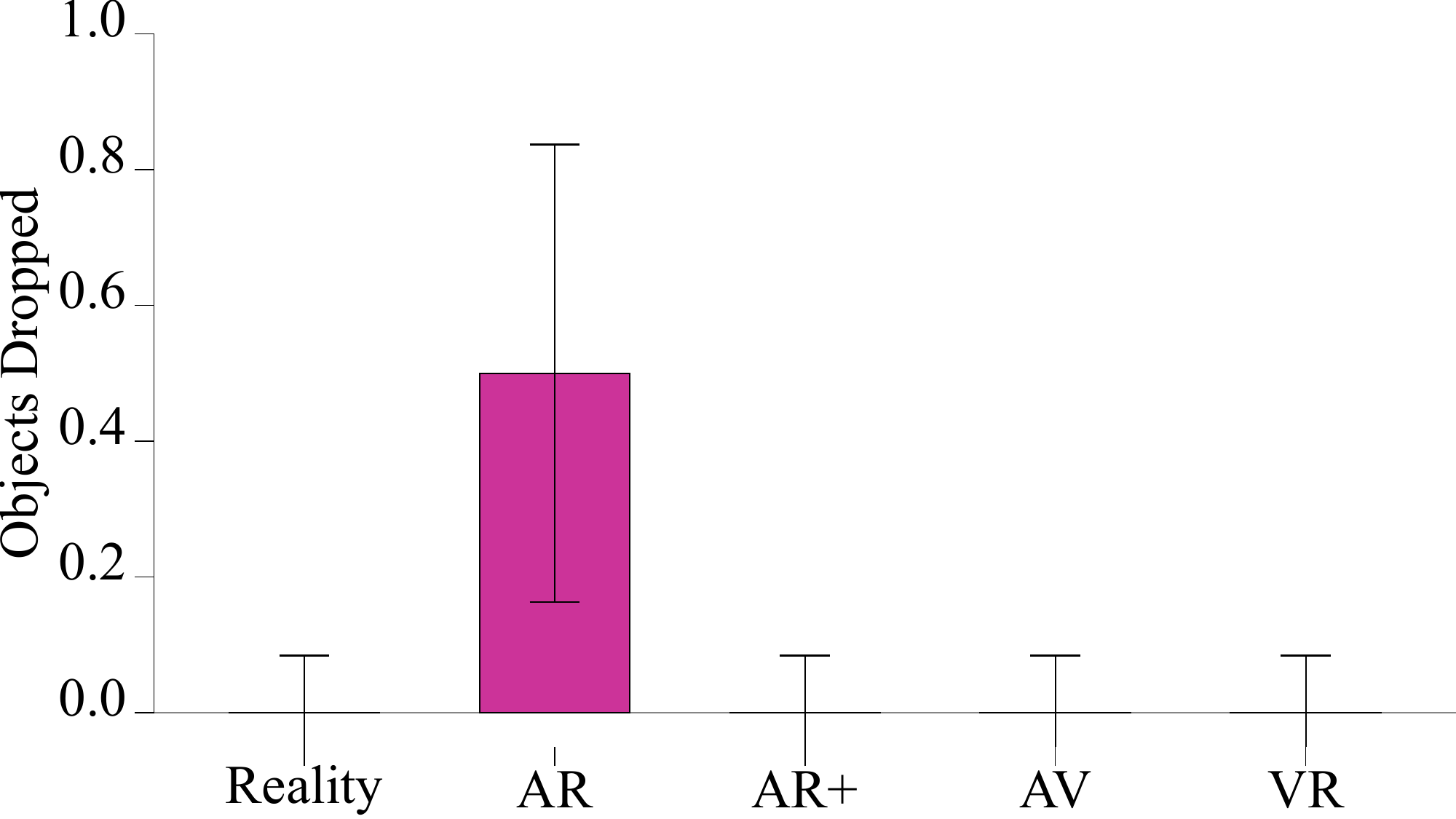}
    \caption{
        Mean Objects Dropped with 95\% confidence intervals across modalities.
        A statistically significant main effect was found, and pairwise differences are denoted by $*$.
    }
    \label{fig:results_dynamic_objects_dropped}
\end{figure}

A statistically significant main effect of modality on objects dropped was observed, $F(4, 36) = 9.000$, $p < .001$ (Table~\ref{tab:results_dynamic_objects_dropped_within_subjects}, Fig.~\ref{fig:results_dynamic_objects_dropped}).
Objects were dropped only in AR ($M = 0.500$) (Table~\ref{tab:results_dynamic_objects_dropped_descriptives}), though no pairwise comparison reached significance under Holm correction (all $p_{holm} = .060$) (Table~\ref{tab:results_dynamic_objects_dropped_post_hoc}).

\begin{table}[H]
    \centering
    \footnotesize
    \caption{Within-Subjects Analysis of Variance for Objects Dropped Across Modalities}
    \label{tab:results_dynamic_objects_dropped_within_subjects}
    {
        \begin{tabularx}{1.0\linewidth}{lRRRRR}
            \toprule
            Cases & Sum of Squares & df & Mean Square & F & p \\
            \cmidrule[0.4pt]{1-6}
            Objects Dropped & $2.000$ & $4$ & $0.500$ & $9.000$ & $<$ .001 \\
            Residuals & $2.000$ & $36$ & $0.056$ & $ $ & $ $ \\
            \bottomrule
        \end{tabularx}
    }
\end{table}

\begin{table}[H]
    \centering
    \footnotesize
    \caption{Descriptive Statistics for Objects Dropped Across Modalities}
    \label{tab:results_dynamic_objects_dropped_descriptives}
    {
        \begin{tabularx}{1.0\linewidth}{lRRRRr}
            \toprule
            Objects Dropped & N & Mean & SD & SE & Coefficient of variation \\
            \cmidrule[0.4pt]{1-6}
            Reality & $10$ & $0.000$ & $0.000$ & $0.000$ & NaN \\
            AR & $10$ & $0.500$ & $0.527$ & $0.167$ & $1.054$ \\
            AR+ & $10$ & $0.000$ & $0.000$ & $0.000$ & NaN \\
            AV & $10$ & $0.000$ & $0.000$ & $0.000$ & NaN \\
            VR & $10$ & $0.000$ & $0.000$ & $0.000$ & NaN \\
            \bottomrule
        \end{tabularx}
    }
\end{table}

\begin{table}[H]
    \centering
    \footnotesize
    \caption{Post Hoc Comparisons for Objects Dropped Across Modalities}
    \label{tab:results_dynamic_objects_dropped_post_hoc}
    {
        \begin{tabularx}{1.0\linewidth}{llrRRRR}
            \toprule
            $ $ & $ $ & Mean Difference & SE & df & t & p$_{holm}$ \\
            \cmidrule[0.4pt]{1-7}
            Reality & AR & $-0.500$ & $0.167$ & $9$ & $-3.000$ & $0.060$ \\
            $ $ & AR+ & $0.000$ & $0.000$ & $9$ & NaN & NaN \\
            & AV & $0.000$ & $0.000$ & $9$ & NaN & NaN \\
            & VR & $0.000$ & $0.000$ & $9$ & NaN & NaN \\
            AR & AR+ & $0.500$ & $0.167$ & $9$ & $3.000$ & $0.060$ \\
            $ $ & AV & $0.500$ & $0.167$ & $9$ & $3.000$ & $0.060$ \\
            & VR & $0.500$ & $0.167$ & $9$ & $3.000$ & $0.060$ \\
            AR+ & AV & $0.000$ & $0.000$ & $9$ & NaN & NaN \\
            $ $ & VR & $0.000$ & $0.000$ & $9$ & NaN & NaN \\
            AV & VR & $0.000$ & $0.000$ & $9$ & NaN & NaN \\
            \bottomrule
        \end{tabularx}
    }
\end{table}


\subsubsection{Collisions}

\begin{figure}[H]
    \centering
    \includegraphics[width=0.50\linewidth]{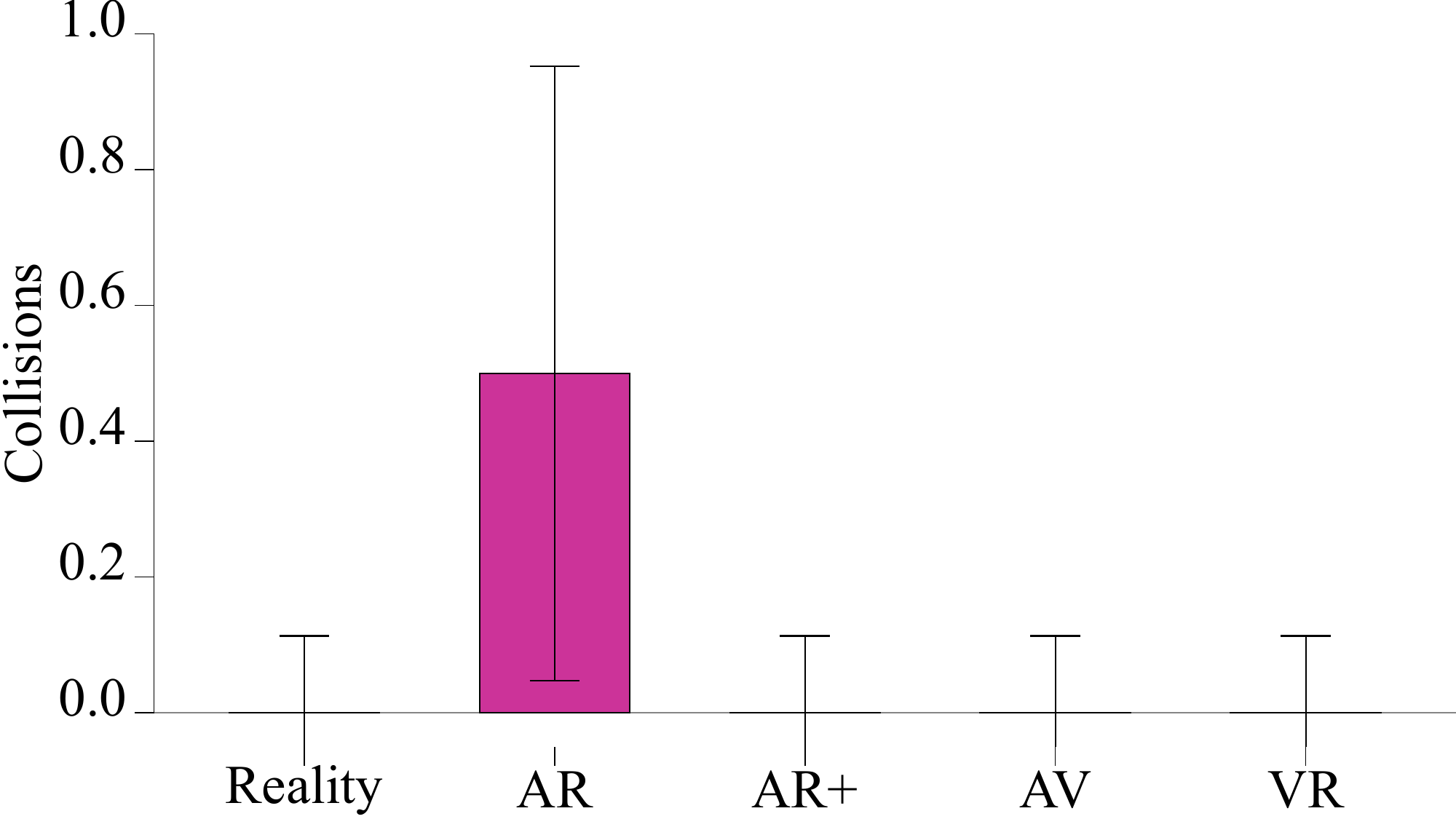}
    \caption{
        Mean Collisions with 95\% confidence intervals across modalities.
        A statistically significant main effect was found, and pairwise differences are denoted by $*$.
    }
    \label{fig:results_dynamic_collisions}
\end{figure}

A statistically significant main effect of modality on collision frequency was found, $F(4, 36) = 5.000$, $p = .003$ (Table~\ref{tab:results_dynamic_collisions_within_subjects}, Fig.~\ref{fig:results_dynamic_collisions}).
Only AR exhibited any collisions ($M = 0.500$) (Table~\ref{tab:results_dynamic_collisions_descriptives}), though no pairwise comparison reached significance under Holm correction (all $p_{holm} = .209$) (Table~\ref{tab:results_dynamic_collisions_post_hoc}).

\begin{table}[H]
    \centering
    \footnotesize
    \caption{Within-Subjects Analysis of Variance for Collisions Across Modalities}
    \label{tab:results_dynamic_collisions_within_subjects}
    {
        \begin{tabularx}{1.0\linewidth}{lRRRRR}
            \toprule
            Cases & Sum of Squares & df & Mean Square & F & p \\
            \cmidrule[0.4pt]{1-6}
            Collisions in RVC & $2.000$ & $4$ & $0.500$ & $5.000$ & $0.003$ \\
            Residuals & $3.600$ & $36$ & $0.100$ & $ $ & $ $ \\
            \bottomrule
        \end{tabularx}
    }
\end{table}

\begin{table}[H]
    \centering
    \footnotesize
    \caption{Descriptive Statistics for Collisions Across Modalities}
    \label{tab:results_dynamic_collisions_descriptives}
    {
        \begin{tabularx}{1.0\linewidth}{lRRRRr}
            \toprule
            Collisions in RVC & N & Mean & SD & SE & Coefficient of variation \\
            \cmidrule[0.4pt]{1-6}
            Reality & $10$ & $0.000$ & $0.000$ & $0.000$ & NaN \\
            AR & $10$ & $0.500$ & $0.707$ & $0.224$ & $1.414$ \\
            AR+ & $10$ & $0.000$ & $0.000$ & $0.000$ & NaN \\
            AV & $10$ & $0.000$ & $0.000$ & $0.000$ & NaN \\
            VR & $10$ & $0.000$ & $0.000$ & $0.000$ & NaN \\
            \bottomrule
        \end{tabularx}
    }
\end{table}

\begin{table}[H]
    \centering
    \footnotesize
    \caption{Post Hoc Comparisons for Collisions Across Modalities}
    \label{tab:results_dynamic_collisions_post_hoc}
    {
        \begin{tabularx}{1.0\linewidth}{llrRRRR}
            \toprule
            $ $ & $ $ & Mean Difference & SE & df & t & p$_{holm}$ \\
            \cmidrule[0.4pt]{1-7}
            Reality & AR & $-0.500$ & $0.224$ & $9$ & $-2.236$ & $0.209$ \\
            $ $ & AR+ & $0.000$ & $0.000$ & $9$ & NaN & NaN \\
            & AV & $0.000$ & $0.000$ & $9$ & NaN & NaN \\
            & VR & $0.000$ & $0.000$ & $9$ & NaN & NaN \\
            AR & AR+ & $0.500$ & $0.224$ & $9$ & $2.236$ & $0.209$ \\
            $ $ & AV & $0.500$ & $0.224$ & $9$ & $2.236$ & $0.209$ \\
            & VR & $0.500$ & $0.224$ & $9$ & $2.236$ & $0.209$ \\
            AR+ & AV & $0.000$ & $0.000$ & $9$ & NaN & NaN \\
            $ $ & VR & $0.000$ & $0.000$ & $9$ & NaN & NaN \\
            AV & VR & $0.000$ & $0.000$ & $9$ & NaN & NaN \\
            \bottomrule
        \end{tabularx}
    }
\end{table}


\section{Discussion}
\label{sec:discussion}

The first analysis of the experiment examines participants fixed in AR, AV, or VR, unable to change modality.
On Objects Grasped and Objects Placed, a significant main effect was observed ($p = .013$ and $p = .027$), yet the post hoc comparisons showed no difference between the fixed modalities, with the lowest $p$-values between AR and AV for Objects Placed ($p_{holm} = .121$) and AR and VR for Objects Grasped ($p_{holm} = .127$).
Qualitatively, however, the NASA-TLX scores differed: Mental Demand between AR vs. AV and AR vs. VR ($p_{holm} = .022$ and $p_{holm} = .003$), Effort between AR and AV ($p_{holm} = .011$), and Frustration between AR and VR ($p_{holm} = .004$).
Notably, no difference was found between AV and VR for any metric, attributable to both conditions relying on low-fidelity data: ``A lot of information (\textit{i.e.}, point cloud, image, robot position) makes cognition hard''.
Therefore, H1 ``The selected reality modality affects task results'' was partially confirmed: no pairwise quantitative difference reached significance, but the significant Mental Demand, Effort, and Frustration differences could indicate increased usability during prolonged operations.

The second analysis compares freely changing the level of reality along the RVC with a fixed reality.
For Objects Grasped, the post hoc comparison showed a significant difference between VR and RVC ($p_{holm} = .040$), where RVC yielded the highest number of grasps ($M = 3.600$) and VR the lowest ($M = 2.200$), qualitatively confirmed in the NASA-TLX comparisons between VR and RVC on Mental Demand, Effort, and Frustration ($p_{holm} < .001$, $p_{holm} = .004$, $p_{holm} < .001$).
For Objects Placed, despite the significant main effect, no pairwise difference emerged, attributable to participants dropping objects, disabling them from placing.
Furthermore, Mental Demand, Effort, and Frustration differed between RVC and AV or VR (all $p_{holm} \leq .007$), indicating higher qualitative performance and lower mental load when utilizing the full RVC.
Therefore, H2 ``Enabling the user to dynamically and freely select the level of immersion improves task results'' was confirmed.

To understand how participants used the system, the RVC condition data was analyzed.
Participants performed most Objects Grasped and Objects Placed in AR, and Time in Modality likewise differed between AR and all other modalities (all $p_{holm} < .001$).
Time Manipulating showed that participants favored AR, preferring to maximize high-fidelity data (video passthrough) while minimizing low-fidelity data, yet still using AR's non-redundant visual cues, such as the end effector clone: ``I found that AR was beneficial for the manipulation approach [...]''.
On the contrary, while navigating, participants utilized both AR and AR+, with no statistical difference between the two: ``Being able to switch between modes is a great addition, [...] being able to see the mapped area on AR+ mode [...] gave me the ability to give more precise navigation instructions'', indicating the usability of adaptive switching by task and context.
Objects Dropped and Collisions showed a main effect but no post hoc significance.
As most time was spent and most objects grasped in AR (``I prefer and use only AR mode''), drawing conclusions on these metrics in RVC is challenging.
H3 ``Users will utilize the RVC spectrum by changing the level of reality'' is thus partially confirmed, especially during navigation, where participants adjusted their level of reality to reduce mental load and gain insight into the robot's perception.
However, 60\% of participants preferred AR, attributable to some only or mostly using AR throughout the RVC condition, slightly but not significantly reducing their mental load.

Noteworthy is the lack of statistical difference for any metric between AV and VR in the fixed conditions, with comparable SUS scores, suggesting participants did not perceive the two modalities as distinct.
The key difference between AV and VR is the AR planes.
However, since the global cost map is calibrated to the physical world (Section~\ref{sec:world_frame_calibration}), a sense of physical presence is maintained even in VR, which AV could otherwise provide.


\section{Limitations and Future Work}
\label{sec:limitations_future_work}

The findings are based on ten participants, limiting statistical power and generalizability.
During manipulation, the robot lacks safety features and can collide with itself and its environment, requiring constant monitoring by participants and increasing their mental burden.
Additionally, inaccuracies in the hand-eye calibration and the robot's self-localization can lead to mismatches between the user's observed location of the robot and the robot's internal perceived position, particularly evident in AR and AV.
Such misalignment can introduce additional cognitive load and may affect the results.

Furthermore, the participant shared the robot's physical environment regardless of the modality, so even in VR, the robot's data remained calibrated with it, maintaining a perception of presence and potentially blurring the difference between AV and VR.
Future experiments should therefore place the operator out of sight of the robot.
Additionally, the system should be extended to a Multi-Robot System~(MRS) with distinct robot roles, where an operator controls robots both in-sight and out-of-sight, potentially affecting the preferred condition.
Finally, a higher level of semi-autonomous manipulation should be explored, removing the need to ``guide'' the end effector to its target pose based on the inverse kinematics~(IK) of the arm: ``[...] the most challenging part was to anticipate the IK of the robot arm joints by just moving the end effector''.


\section{Conclusion}
\label{sec:conclusion}

This study explored CAs, which extend human capabilities by letting operators act remotely through robotic embodiments in service domains~\cite{ishiguro_realisation_2021}.
As prior XR-HRI research commonly fixes a modality selected from the device's capabilities or precedent~\cite{suzuki_augmented_2022}, this study investigated whether the chosen XR modality impacts task outcomes and introduced an adaptive interface with real-time modality switching along the RVC~\cite{milgram_augmented_1995}.
In a simulated multi-room service environment, participants controlled a mobile manipulator through an XR HMD visualizing the robot's egocentric and exocentric data, either in a fixed XR modality (AR, AV, or VR) or free to switch dynamically along the RVC.

The results support the two contributions of Section~\ref{sec:introduction}.
Regarding the first contribution, H1 was partially confirmed: quantitative performance did not change significantly, but AR yielded significantly lower mental demand than AV and VR, lower effort than AV, and lower frustration than VR, demonstrating that the modality selected at design time affects HRI results.
Regarding the second contribution, H2 was confirmed and H3 partially confirmed: dynamically switching along the RVC yielded more objects grasped than VR and lower mental demand, effort, and frustration than AV and VR, with participants favoring AR during manipulation and AR or AR+ during navigation, demonstrating that operator-controlled modality changes improve HRI results.

Beyond controlled experiments, the developed system serves as a foundation for large-scale demonstrations~\cite{el_hafi_public_2025} and has been featured in media reports~\cite{nikkan_kogyo_shimbun_ritsumeikan_2024, nikkan_kogyo_shimbun_ritsumeikan_2023}, refining the platform-agnostic XR-HRI system, which is publicly available at \url{https://github.com/CarlTornberg/XR-HRI}.
By letting operators tailor how much virtual content complements the robot's physical embodiment, adaptive XR interfaces along the RVC form a human-centered building block toward the avatar-symbiotic society pursued by JST Moonshot R\&D Program Goal 1~\cite{ishiguro_realisation_2021}.


\section*{Disclosure Statement}

No potential conflict of interest was reported by the authors.


\section*{Funding}

This work was supported by the Japan Science and Technology Agency~(JST), Moonshot Research~\& Development Program, Grant Number JPMJMS2011, and by the Japan Society for the Promotion of Science~(JSPS), KAKENHI Grants-in-Aid for Scientific Research, Grant Number JP22K17981.









\bibliographystyle{tfnlm}
\bibliography{main}


\clearpage


\appendix

\section{Complementary Result Tables}
\label{sec:appendix}

\begin{table}[H]
    \centering
    \footnotesize
    \caption{Within-Subjects Analysis of Variance for the NASA-TLX Mental Demand Across Conditions}
    \label{tab:results_fixed_nasa_tlx_mental_within_subjects}
    {
        \begin{tabularx}{1.0\linewidth}{Lrrrrrr}
            \toprule
            Cases & Sphericity Correction & Sum of Squares & df & Mean Square & F & p \\
            \cmidrule[0.4pt]{1-7}
            Mental Demand & Greenhouse-Geisser & $516.875$ & $1.579$ & $327.355$ & $18.881$ & $<$ .001 \\
            Residuals & Greenhouse-Geisser & $246.375$ & $14.210$ & $17.338$ & $ $ & $ $ \\
            \bottomrule
        \end{tabularx}
    }
\end{table}

\begin{table}[H]
    \centering
    \footnotesize
    \caption{Descriptive Statistics for NASA-TLX Mental Demand Across Conditions}
    \label{tab:results_fixed_nasa_tlx_mental_descriptive}
    {
        \begin{tabularx}{1.0\linewidth}{lRRRRr}
            \toprule
            Mental Demand & N & Mean & SD & SE & Coefficient of variation \\
            \cmidrule[0.4pt]{1-6}
            AR & $10$ & $7.700$ & $4.990$ & $1.578$ & $0.648$ \\
            AV & $10$ & $14.100$ & $4.909$ & $1.552$ & $0.348$ \\
            VR & $10$ & $14.800$ & $2.936$ & $0.929$ & $0.198$ \\
            RVC & $10$ & $6.900$ & $4.067$ & $1.286$ & $0.589$ \\
            \bottomrule
        \end{tabularx}
    }
\end{table}

\begin{table}[H]
    \centering
    \footnotesize
    \caption{Post Hoc Comparisons for NASA-TLX Mental Demand Across Conditions}
    \label{tab:results_fixed_nasa_tlx_mental_post_hoc}
    {
        \begin{tabularx}{1.0\linewidth}{llrRRRR}
            \toprule
            $ $ & $ $ & Mean Difference & SE & df & t & p$_{holm}$ \\
            \cmidrule[0.4pt]{1-7}
            AR & AV & $-6.400$ & $1.863$ & $9$ & $-3.435$ & $0.022$ \\
            $ $ & VR & $-7.100$ & $1.362$ & $9$ & $-5.214$ & $0.003$ \\
            & RVC & $0.800$ & $0.512$ & $9$ & $1.562$ & $0.305$ \\
            AV & VR & $-0.700$ & $1.221$ & $9$ & $-0.573$ & $0.580$ \\
            $ $ & RVC & $7.200$ & $1.625$ & $9$ & $4.431$ & $0.007$ \\
            VR & RVC & $7.900$ & $1.110$ & $9$ & $7.117$ & $<$ .001 \\
            \bottomrule
        \end{tabularx}
    }
\end{table}

\begin{table}[H]
    \centering
    \footnotesize
    \caption{Within-Subjects Analysis of Variance for the NASA-TLX Physical Demand Across Conditions}
    \label{tab:results_fixed_nasa_tlx_physical_within_subjects}
    {
        \begin{tabularx}{1.0\linewidth}{lRRRRR}
            \toprule
            Cases & Sum of Squares & df & Mean Square & F & p \\
            \cmidrule[0.4pt]{1-6}
            Physical Demand & $80.900$ & $3$ & $26.967$ & $5.311$ & $0.005$ \\
            Residuals & $137.100$ & $27$ & $5.078$ & $ $ & $ $ \\
            \bottomrule
        \end{tabularx}
    }
\end{table}

\begin{table}[H]
    \centering
    \footnotesize
    \caption{Descriptive Statistics for NASA-TLX Physical Demand Across Conditions}
    \label{tab:results_fixed_nasa_tlx_physical_descriptive}
    {
        \begin{tabularx}{1.0\linewidth}{lRRRRr}
            \toprule
            Physical Demand & N & Mean & SD & SE & Coefficient of variation \\
            \cmidrule[0.4pt]{1-6}
            AR & $10$ & $3.800$ & $1.751$ & $0.554$ & $0.461$ \\
            AV & $10$ & $7.000$ & $3.621$ & $1.145$ & $0.517$ \\
            VR & $10$ & $6.900$ & $3.381$ & $1.069$ & $0.490$ \\
            RVC & $10$ & $4.500$ & $1.650$ & $0.522$ & $0.367$ \\
            \bottomrule
        \end{tabularx}
    }
\end{table}

\begin{table}[H]
    \centering
    \footnotesize
    \caption{Post Hoc Comparisons for NASA-TLX Physical Demand Across Conditions}
    \label{tab:results_fixed_nasa_tlx_physical_post_hoc}
    {
        \begin{tabularx}{1.0\linewidth}{llrRRRR}
            \toprule
            $ $ & $ $ & Mean Difference & SE & df & t & p$_{holm}$ \\
            \cmidrule[0.4pt]{1-7}
            AR & AV & $-3.200$ & $1.162$ & $9$ & $-2.753$ & $0.134$ \\
            $ $ & VR & $-3.100$ & $1.224$ & $9$ & $-2.532$ & $0.161$ \\
            & RVC & $-0.700$ & $0.517$ & $9$ & $-1.353$ & $0.418$ \\
            AV & VR & $0.100$ & $0.752$ & $9$ & $0.133$ & $0.897$ \\
            $ $ & RVC & $2.500$ & $1.035$ & $9$ & $2.414$ & $0.161$ \\
            VR & RVC & $2.400$ & $1.157$ & $9$ & $2.075$ & $0.203$ \\
            \bottomrule
        \end{tabularx}
    }
\end{table}

\begin{table}[H]
    \centering
    \footnotesize
    \caption{Within-Subjects Analysis of Variance for the NASA-TLX Temporal Demand Across Conditions}
    \label{tab:results_fixed_nasa_tlx_temporal_within_subjects}
    {
        \begin{tabularx}{1.0\linewidth}{lrRRRR}
            \toprule
            Cases & Sum of Squares & df & Mean Square & F & p \\
            \cmidrule[0.4pt]{1-6}
            Temporal Demand & $6.200$ & $3$ & $2.067$ & $0.351$ & $0.788$ \\
            Residuals & $158.800$ & $27$ & $5.881$ & $ $ & $ $ \\
            \bottomrule
        \end{tabularx}
    }
\end{table}

\begin{table}[H]
    \centering
    \footnotesize
    \caption{Descriptive Statistics for NASA-TLX Temporal Demand Across Conditions}
    \label{tab:results_fixed_nasa_tlx_temporal_descriptive}
    {
        \begin{tabularx}{1.0\linewidth}{lRRRRr}
            \toprule
            Temporal Demand & N & Mean & SD & SE & Coefficient of variation \\
            \cmidrule[0.4pt]{1-6}
            AR & $10$ & $8.400$ & $4.452$ & $1.408$ & $0.530$ \\
            AV & $10$ & $7.700$ & $2.869$ & $0.907$ & $0.373$ \\
            VR & $10$ & $7.300$ & $3.433$ & $1.086$ & $0.470$ \\
            RVC & $10$ & $7.800$ & $3.994$ & $1.263$ & $0.512$ \\
            \bottomrule
        \end{tabularx}
    }
\end{table}

\begin{table}[H]
    \centering
    \footnotesize
    \caption{Within-Subjects Analysis of Variance for the NASA-TLX Performance Across Conditions}
    \label{tab:results_fixed_nasa_tlx_performance_within_subjects}
    {
        \begin{tabularx}{1.0\linewidth}{lRRRRR}
            \toprule
            Cases & Sum of Squares & df & Mean Square & F & p \\
            \cmidrule[0.4pt]{1-6}
            Performance & $55.475$ & $3$ & $18.492$ & $0.892$ & $0.458$ \\
            Residuals & $559.775$ & $27$ & $20.732$ & $ $ & $ $ \\
            \bottomrule
        \end{tabularx}
    }
\end{table}

\begin{table}[H]
    \centering
    \footnotesize
    \caption{Descriptive Statistics for NASA-TLX Performance Across Conditions}
    \label{tab:results_fixed_nasa_tlx_performance_descriptive}
    {
        \begin{tabularx}{1.0\linewidth}{lRRRRr}
            \toprule
            Performance & N & Mean & SD & SE & Coefficient of variation \\
            \cmidrule[0.4pt]{1-6}
            AR & $10$ & $7.600$ & $4.427$ & $1.400$ & $0.583$ \\
            AV & $10$ & $8.700$ & $4.945$ & $1.564$ & $0.568$ \\
            VR & $10$ & $10.700$ & $5.229$ & $1.654$ & $0.489$ \\
            RVC & $10$ & $8.100$ & $6.315$ & $1.997$ & $0.780$ \\
            \bottomrule
        \end{tabularx}
    }
\end{table}

\begin{table}[H]
    \centering
    \footnotesize
    \caption{Within-Subjects Analysis of Variance for the NASA-TLX Effort Across Conditions}
    \label{tab:results_fixed_nasa_tlx_effort_within_subjects}
    {
        \begin{tabularx}{1.0\linewidth}{lRRRRR}
            \toprule
            Cases & Sum of Squares & df & Mean Square & F & p \\
            \cmidrule[0.4pt]{1-6}
            Effort & $313.675$ & $3$ & $104.558$ & $11.833$ & $<$ .001 \\
            Residuals & $238.575$ & $27$ & $8.836$ & $ $ & $ $ \\
            \bottomrule
        \end{tabularx}
    }
\end{table}

\begin{table}[H]
    \centering
    \footnotesize
    \caption{Descriptive Statistics for NASA-TLX Effort Across Conditions}
    \label{tab:results_fixed_nasa_tlx_effort_descriptive}
    {
        \begin{tabularx}{1.0\linewidth}{lRRRRr}
            \toprule
            Effort & N & Mean & SD & SE & Coefficient of variation \\
            \cmidrule[0.4pt]{1-6}
            AR & $10$ & $7.100$ & $4.254$ & $1.345$ & $0.599$ \\
            AV & $10$ & $11.800$ & $4.662$ & $1.474$ & $0.395$ \\
            VR & $10$ & $11.500$ & $4.089$ & $1.293$ & $0.356$ \\
            RVC & $10$ & $5.300$ & $3.713$ & $1.174$ & $0.701$ \\
            \bottomrule
        \end{tabularx}
    }
\end{table}

\begin{table}[H]
    \centering
    \footnotesize
    \caption{Post Hoc Comparisons for NASA-TLX Effort Across Conditions}
    \label{tab:results_fixed_nasa_tlx_effort_post_hoc}
    {
        \begin{tabularx}{1.0\linewidth}{llrRRRR}
            \toprule
            $ $ & $ $ & Mean Difference & SE & df & t & p$_{holm}$ \\
            \cmidrule[0.4pt]{1-7}
            AR & AV & $-4.700$ & $1.155$ & $9$ & $-4.069$ & $0.011$ \\
            $ $ & VR & $-4.400$ & $1.600$ & $9$ & $-2.750$ & $0.067$ \\
            & RVC & $1.800$ & $1.041$ & $9$ & $1.728$ & $0.236$ \\
            AV & VR & $0.300$ & $1.680$ & $9$ & $0.179$ & $0.862$ \\
            $ $ & RVC & $6.500$ & $1.088$ & $9$ & $5.975$ & $0.001$ \\
            VR & RVC & $6.200$ & $1.272$ & $9$ & $4.875$ & $0.004$ \\
            \bottomrule
        \end{tabularx}
    }
\end{table}

\begin{table}[H]
    \centering
    \footnotesize
    \caption{Within-Subjects Analysis of Variance for the NASA-TLX Frustration Across Conditions}
    \label{tab:results_fixed_nasa_tlx_frustration_within_subjects}
    {
        \begin{tabularx}{1.0\linewidth}{Lrrrrrr}
            \toprule
            Cases & Sphericity Correction & Sum of Squares & df & Mean Square & F & p \\
            \cmidrule[0.4pt]{1-7}
            Frustration & Greenhouse-Geisser & $648.200$ & $1.741$ & $372.407$ & $12.702$ & $<$ .001 \\
            Residuals & Greenhouse-Geisser & $459.300$ & $15.665$ & $29.320$ & $ $ & $ $ \\
            \bottomrule
        \end{tabularx}
    }
\end{table}

\begin{table}[H]
    \centering
    \footnotesize
    \caption{Descriptive Statistics for NASA-TLX Frustration Across Conditions}
    \label{tab:results_fixed_nasa_tlx_frustration_descriptive}
    {
        \begin{tabularx}{1.0\linewidth}{lRRRRr}
            \toprule
            Frustration & N & Mean & SD & SE & Coefficient of variation \\
            \cmidrule[0.4pt]{1-6}
            AR & $10$ & $5.900$ & $5.109$ & $1.616$ & $0.866$ \\
            AV & $10$ & $9.900$ & $3.872$ & $1.224$ & $0.391$ \\
            VR & $10$ & $14.000$ & $4.422$ & $1.398$ & $0.316$ \\
            RVC & $10$ & $3.400$ & $2.066$ & $0.653$ & $0.608$ \\
            \bottomrule
        \end{tabularx}
    }
\end{table}

\begin{table}[H]
    \centering
    \footnotesize
    \caption{Post Hoc Comparisons for NASA-TLX Frustration Across Conditions}
    \label{tab:results_fixed_nasa_tlx_frustration_post_hoc}
    {
        \begin{tabularx}{1.0\linewidth}{llrRRRR}
            \toprule
            $ $ & $ $ & Mean Difference & SE & df & t & p$_{holm}$ \\
            \cmidrule[0.4pt]{1-7}
            AR & AV & $-4.000$ & $2.671$ & $9$ & $-1.498$ & $0.337$ \\
            $ $ & VR & $-8.100$ & $1.670$ & $9$ & $-4.851$ & $0.004$ \\
            & RVC & $2.500$ & $1.845$ & $9$ & $1.355$ & $0.337$ \\
            AV & VR & $-4.100$ & $1.683$ & $9$ & $-2.436$ & $0.113$ \\
            $ $ & RVC & $6.500$ & $1.310$ & $9$ & $4.961$ & $0.004$ \\
            VR & RVC & $10.600$ & $1.593$ & $9$ & $6.654$ & $<$ .001 \\
            \bottomrule
        \end{tabularx}
    }
\end{table}


\end{document}